\documentclass[journal]{IEEEtran}
\usepackage[bookmarks=true, colorlinks, linkcolor=red, anchorcolor=blue, citecolor=green]{hyperref}
\usepackage{times}
\usepackage{amsmath}
\usepackage{amssymb}
\usepackage{commath}
\usepackage{tabularx}
\usepackage{booktabs}
\usepackage{float}
\usepackage{algorithm}
\usepackage{algpseudocode}
\usepackage{amsmath}
\usepackage{graphicx}
\usepackage{subfigure}
\usepackage{color}
\usepackage{CJK}
\usepackage{cite}
\usepackage{multirow}

\begin{document}

\title{Comparative evaluation of 2D feature correspondence selection algorithms}
\author{Chen~Zhao,~Jiaqi~Yang,~Yang~Xiao,~and Zhiguo~Cao
\thanks{C. Zhao, Jiaqi Yang, Y. Xiao, and Zhiguo Cao was with School of Artificial Intelligence and Automation, Huazhong University of Science and Technology, Wuhan, 430074, China e-mail: (hust\_zhao@hust.edu.cn, jqyang@hust.edu.cn, Yang\_Xiao@hust.edu.cn, zgcao@hust.edu.cn) (Corresponding author: Zhiguo Cao.)}
}

\markboth{Journal of \LaTeX\ Class Files,~Vol.~14, No.~8, August~2015}%
{Shell \MakeLowercase{\textit{et al.}}: Bare Demo of IEEEtran.cls for IEEE Journals}

\maketitle

\begin{abstract}
Correspondence selection aiming at seeking correct feature correspondences from raw feature matches is pivotal for a number of feature-matching-based tasks. Various 2D (image) correspondence selection algorithms have been presented with decades of progress. Unfortunately, the lack of an in-depth evaluation makes it difficult for developers to choose a proper algorithm given a specific application. This paper fills this gap by evaluating eight 2D correspondence selection algorithms ranging from classical methods to the most recent ones on four standard datasets. The diversity of experimental datasets brings various nuisances including zoom, rotation, blur, viewpoint change, JPEG compression, light change, different rendering styles and multi-structures for comprehensive test. To further create different distributions of initial matches, a set of combinations of detector and descriptor is also taken into consideration. We measure the quality of a correspondence selection algorithm from four perspectives, i.e., precision, recall, F-measure and efficiency. According to evaluation results, the current advantages and limitations of all considered algorithms are aggregately summarized which could be treated as a ``user guide'' for the following developers.
\end{abstract}


\section{Introduction}\label{sec:intr}
\begin{figure}[htb]  
	\centering
	\subfigure[Initial feature matches]
    {%
	\begin{minipage}[t]{0.47\linewidth}
		\includegraphics[width=1\linewidth]{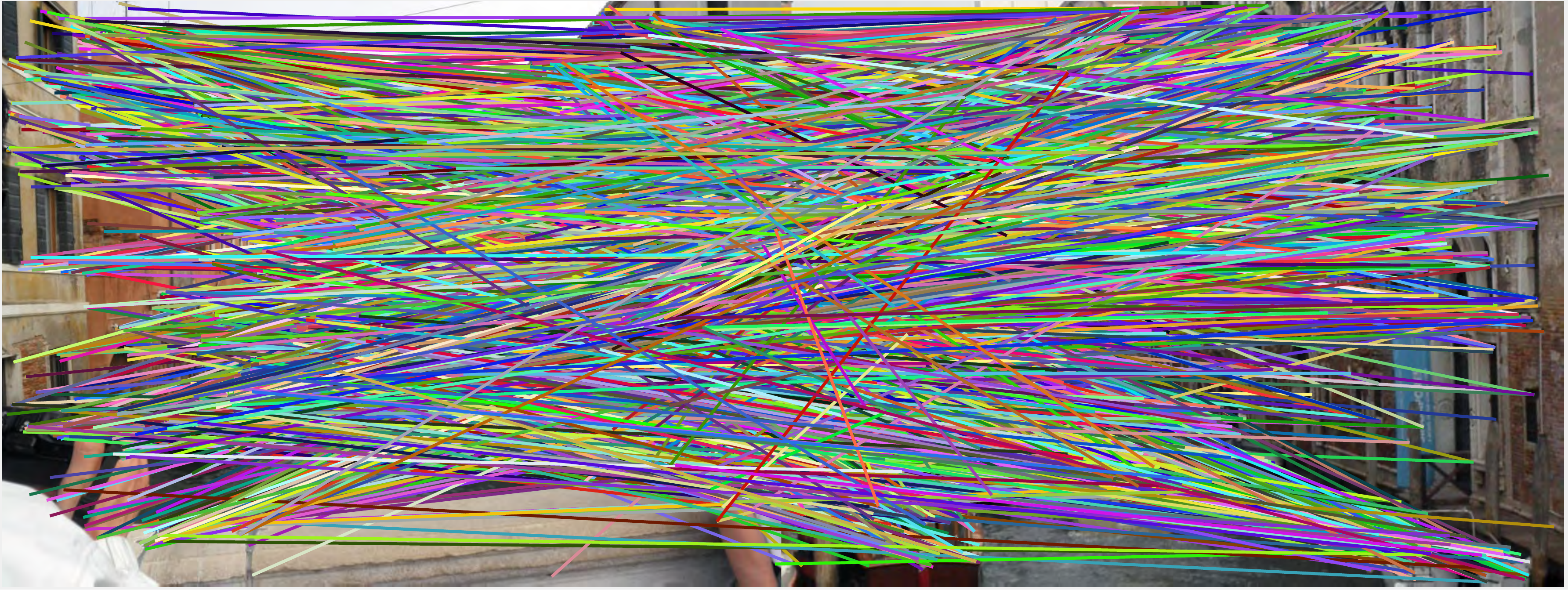}
	\end{minipage}
	}
	\subfigure[Selected matching]
    {%
	\begin{minipage}[t]{0.47\linewidth}
		\includegraphics[width=1\linewidth]{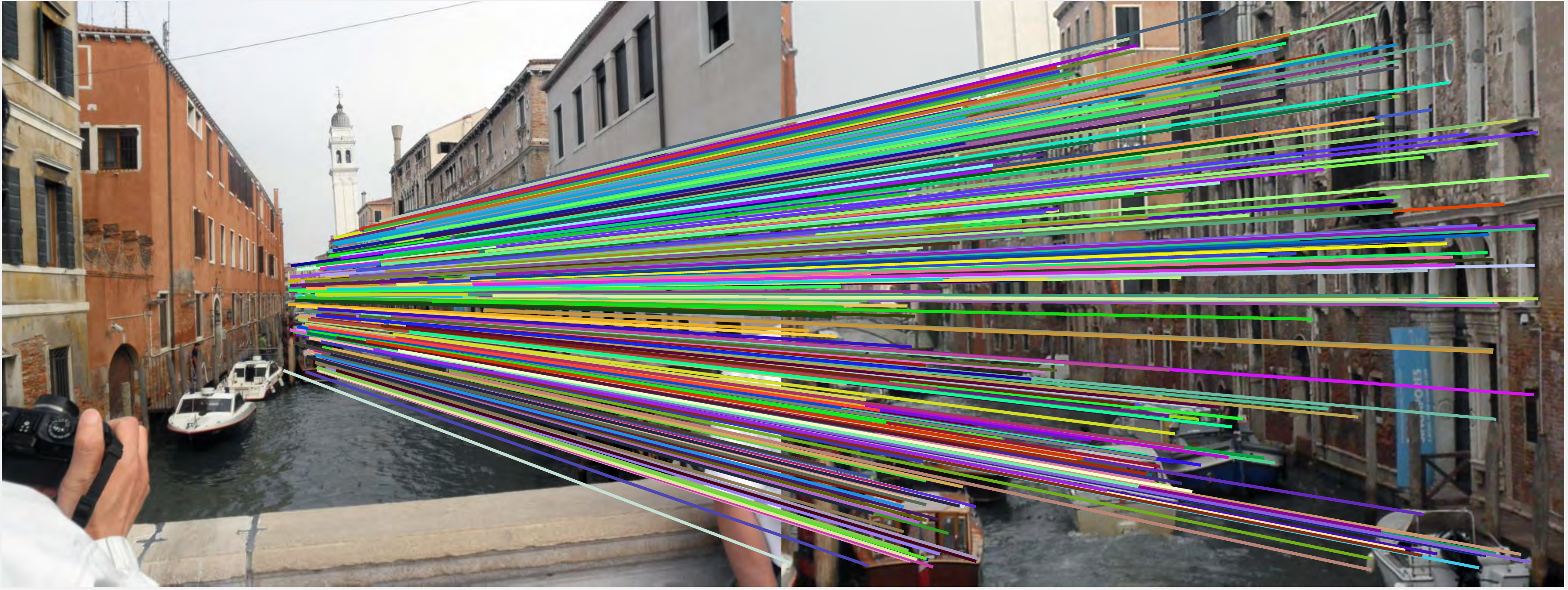}
	\end{minipage}
	}
	\caption{An exemplar illustration of the 2D feature correspondence selection, where the colorized lines represent correspondences between two images. (a) Initial feature matches generated by brute-force descriptor matching. (b) Feature matches after correspondence selection.}
	\label{fig:inspr}
\end{figure}  

Feature correspondence selection is a fundamental and critical task in computer vision and robotics. It is the basis for a wide range of applications, such as structure-from-motion~\cite{snavely2008modeling}, simultaneous localization and mapping~\cite{benhimane2004real}, tracking~\cite{hare2012efficient}, image stitching~\cite{brown2007automatic}, and object recognition~\cite{lowe1999object}, to name just a few. 

The main purpose of correspondence selection is retrieving as many as correct correspondences (also known as \textit{inliers}) from the initial feature matches. Usually, this task is under the background of feature matching. The general process of feature matching starts by detecting representative points, namely keypoints, for two images to be matched. Then, local descriptors such as SIFT~\cite{lowe2004distinctive} and ORB~\cite{rublee2011orb} are employed to perform feature description for those keypoints. To build the connection between two images, keypoints with similar feature descriptors are matched, generating a set of raw feature matches. However, the initial feature matches often suffer from severe wrong matches (as shown in Fig.~\ref{fig:inspr}(a)) due to the limited distinctiveness of feature descriptors or/and external interferences such as noise and occlusion. This problem makes correspondence selection a necessity for accurate feature matching. Fig.~\ref{fig:inspr}(b) shows that those matches after correspondence selection are far more consistent than the initial feature matches. This consensus allows massive high-level vision tasks. For instance, homography, affine and essential matrices can be estimated from those consistent correspondences, thus allowing us to compute the transformation between two images and warp them into a unified coordinate system~\cite{brown2007automatic}. Other applications also involve camera parameter estimation~\cite{snavely2008modeling} and object tracking~\cite{hare2012efficient}. Nonetheless, the correspondence selection problem is difficult in real applications due to several factors, e.g., zoom, rotation, blur, viewpoint change, JPEG compression, light change, different rendering styles, multi-structures, and etc. Different scenarios will also lead to different distributions of feature matches which are linearly non-separable.

To address these problems, many approaches that have been presented during the past two decades can be divided into two categories~\cite{Ma2014Robust}: parametric and non-parametric methods. (i) For parametric methods, they seek consistent correspondences grounded on parametric geometric models. Typical methods include the random sample consensus (RANSAC)~\cite{fischler1981random}, the progressive sample consensus (PROSAC)~\cite{Chum2005Matching}, the universal framework for random sample consensus (USAC)~\cite{Raguram2013USAC}, and etc. (ii) For non-parametric methods, they are independent from parametric model assumptions. Some of them search correspondence inliers via either feature similarity constraint or geometric constraint, such as the nearest neighbor similarity ratio (NNSR)~\cite{lowe2004distinctive}, spectral technique (ST)~\cite{Leordeanu2005A}, game-theoretic matching (GTM)~\cite{albarelli2012imposing}, graph-based affine invariant matching (GAIM)~\cite{Collins2014An} and locality preserving matching (LPM)~\cite{Ma2017Locality}. There are also constraint-independent non-parametric methods such as identifying point correspondences by correspondence function (ICF)~\cite{Li2010Rejecting}, vector field consensus (VFC)~\cite{Ma2014Robust}, grid-based motion statistics (GMS)~\cite{bian2017gms} and coherence based decision boundaries (CODE)~\cite{Lin2017CODE}. With the wealth of existing correspondence selection methods, however, it is on the one hand difficult for developers to choose the most proper method given a specific application and on the other hand confusing for researchers to compare these methods under different conditions. This problem is mainly due to the fact that most methods were tested under a specific application scenario and compared with a limited number of baselines.

Some performance evaluations in the field of image feature matching also exist. For instance, Mikolajczyk et al.~\cite{mikolajczyk2005performance} and Heinly et al.~\cite{Heinly2012Comparative} evaluated the performance of several 2D feature descriptors. Aans et al.~\cite{Aan2012Interesting} investigated the performance of 2D feature detectors. Moreels et al.~\cite{moreels2007evaluation} performed an aggregated evaluation of both 2D detectors and descriptors. In addition to feature detectors and descriptors, Raguram et al.~\cite{Raguram2008A} tested the performance of a set of random sample consensus methods including the popular RANSAC and its variants. However, all these evaluations are either not in line with 2D correspondence selection or not comprehensive enough for an in-depth comparison. First, the critical step in correspondence selection is finding correspondence consensus, while feature detection and description aim at building high-quality initial feature correspondences (such quality is difficult to be guaranteed without correspondence selection~\cite{lowe2004distinctive}). Second, the performance of non-parametric approaches and some recent algorithms remains unclear (only parametric methods were tested in~\cite{Raguram2008A}). 

In these regards, we present the first comprehensive evaluation, to the best of our knowledge, for 2D correspondence selection from different perspectives in a uniform experimental framework. The considered methods in our evaluation range from classical algorithms to the most recent ones, typically covering both parametric and non-parametric approaches. To be specific, RANSAC~\cite{fischler1981random} and USAC~\cite{Raguram2013USAC} are selected from the parametric family, as RANSAC is arguably the most popular parametric approach and USAC is a well-known modified version of RANSAC. As for non-parametric methods, we choose NNSR~\cite{lowe2004distinctive} as a representative of those methods based on descriptor similarity constraints. ST~\cite{Leordeanu2005A}, GTM~\cite{albarelli2012imposing} and LPM~\cite{Ma2017Locality} are selected as they all rely on the geometric consensus. VFC~\cite{Ma2014Robust} and the recent GMS~\cite{bian2017gms} are taken into consideration since they eliminate outliers from the perspective of  statistical measures. In order to compare those methods from different perspectives, we choose four standard datasets, i.e., VGG~\cite{Mikolajczyk2005A}, Heinly~\cite{Heinly2012Comparative}, Symbench~\cite{Snavely2012Image}, AdelaideRMF~\cite{Wong2011Dynamic}, as experimental platforms under the motivation to test those correspondence selection methods' overall performance when faced with a variety of nuisances rather than in their favoring circumstances. For instance, geometric constraints may turn to be vulnerable under rigid/non-rigid transformations such as zoom and rotation; feature similarity constraints are suspicious when the image undergoes blur and light changes; parametric models (the homography matrix) can hardly cope with scenes with parallax (all above conclusions have been verified in Sect.~\ref{sec:exper}). The considered datasets well cover these concerns. To be specific, VGG is a hybrid dataset containing challenges including zoom, rotation, blur, JPEG compression, light and viewpoint change. Heinly contains pure zoom and rotation. Symbench involves scenes with light changes and varying rendering styles. AdelaideRMF possesses viewpoint change and multi-structures, resulting in parallax. The behavior of each method is quantitatively measured using precision, recall and F-measure~\cite{bian2017gms,lin2014bilateral,Ma2017Locality}. In addition, the performance under preselected correspondences (with higher inlier ratios) and different detector-descriptor combinations are also accessed to test their flexibility with respect to the inlier ratio and correspondence distribution changes. Finally, the efficiency with respect to different scales of initial feature matches are examined. According to the experimental outcomes, we make an aggregated summary of the current advantages and limitations of our evaluated methods as well as their suitable applications.

In a nutshell, the contributions of this paper are threefold:
\begin{itemize}
	\item A review and the core computation steps of eight state-of-the-art 2D correspondence selection algorithms are presented.
	\item We comprehensively evaluate and compare the performance, the robustness to a variety of perturbations and the efficiency of each algorithm on four standard datasets consisting of hundreds of images with zoom, rotation, blur, viewpoint change, JPEG compression, light change, different rendering styles and multi-structures. 
	\item Instructive summarizations including merits, demerits and suitable applications of the tested methods are given that can be served as a ``user guide''  for the developers.
\end{itemize}

The remainder of this paper is organized as follows. Sect.~\ref{sec:rela} gives a review of 2D correspondence selection algorithms and relevant evaluations. Sect.~\ref{sec:method} presents the core computation steps of eight state-of-the-art approaches. Sect.~\ref{sec:exper} describes the experimental setup including datasets, criteria and implementation details of the evaluated methods. Qualitative and quantitative experimental results are shown in Sect.~\ref{sec:res}. Summary and discussion are presented in Sect.~\ref{sec:sum}. Conclusions are finally drawn in Sect.~\ref{sec:con}.
\section{Related work}\label{sec:rela}
This section briefly reviews the prior works of 2D correspondence selection including both parametric and non-parametric categories. Relevant evaluations in the field of feature matching are also discussed.
\subsection{Correspondence selection methods}
For parametric methods, the most well-known algorithm is arguably RANSAC presented by Fischler et al.~\cite{fischler1981random}. RANSAC iteratively explores the space of model parameters by  randomly sampling and estimates the most reliable model based on the maximum number of inliers. Then, outliers can be removed using the generated model. Several variants of RANSAC such as MLESAC~\cite{Torr2000MLESAC}, LO-RANSAC~\cite{Chum2003Locally}, PROSAC~\cite{Chum2005Matching} and USAC~\cite{Raguram2013USAC} were proposed in the following decades. MLESAC employs the maximum likelihood estimation rather than the inlier count to check the solutions. LO-RANSAC inserts an optimization process where the generated model is refined by the subset of inliers. A weighted sampling step is adopted instead of random sampling in PROSAC. This method sorts the raw correspondences by matching quality and generates hypotheses from the most promising correspondences. USAC extends the standard hypothesize-and-verify structure in RANSAC and presents a universal framework that integrates advantages of previous parametric methods. In addition, some other approaches relying on local parametric structures have also been developed, such as agglomerative correspondence clustering (ACC)~\cite{cho2009feature}, multi-structures robust fitting (Multi-GS)~\cite{chin2010accelerated}, Hough voting and inverted Hough voting (HVIV)~\cite{Chen2013Robust}. ACC uses Hessian-affine detector~\cite{Mikolajczyk2004Hessian}, which is invariant to affine transformations, to estimate the local homography matrix as constraints. The initial correspondences are then clustered based on the constraints, and the clusters with inliers are supposed to be larger than the ones constituted by outliers. Multi-GS generates a series of tentative hypotheses by random sampling and considers that two correspondences from the same local structure are inliers if they share a common list of hypotheses. HVIV employs the BPLR detector~\cite{Kim2011Boundary} to cluster correspondences and estimates the homographic transformation for each correspondence as well. The most plausible correspondence in each cluster is then selected using normalized kernel density estimation.

For non-parametric methods, their theoretical foundations are not always the same. A widely-used strategy is exploiting the consistency information of local geometric structures or appearance (feature similarity). Specifically, Lowe et al.~\cite{lowe2004distinctive} proposed a nearest neighbor similarity ratio (NNSR) method that assigns a penalty equaling to the ratio of the closest to the second-closest feature distance to each correspondence and treats those correspondences with low ratios as inliers. Leordeanu et al.~\cite{Leordeanu2005A} presented spectral technique (ST), where an affinity matrix is built using pairwise geometric constraints to remove mismatches in conflict with the most credible correspondences. Albarelli et al.~\cite{albarelli2012imposing} casted the selection of
correspondences in a game theoretic framework, known as game-theoretic matching (GTM), where a natural selection process allows corresponding points that satisfy a mutual distance constraint to thrive. Cho et al.~\cite{Cho2010Reweighted} presented reweighted random walk algorithm (RRWM) for graph matching. An associated graph between two sets of candidate correspondences is drawn at first, and reliable nodes indicating the consistent correspondences in this graph are then selected by the reweighted random walk algorithm. Ma et al.~\cite{Ma2017Locality} proposed locality preserving matching (LPM) to improve inlier selection by maintaining the local neighborhood structures of those potential true matches.
Some non-parametric approaches that formulate the correspondence selection problem as a statistics problem have also been used, e.g., vector field consensus (VFC)~\cite{Ma2014Robust} and grid-based motion statistics (GMS)~\cite{bian2017gms}. VFC supposes that the noise around inliers and outliers falls in different distributions. This approach estimates the probability of inliers by the maximum likelihood estimation for parameters in the mixture probabilistic model. Additionally, GMS rejects false matches by counting the quantity of matches in small neighborhoods and achieves real-time performance with an efficient grid-based score estimator.
\subsection{Other evaluations}
In the feature matching field, some evaluations of 2D/3D local descriptors and detectors have been performed. For instance, Mikolajczyk et al.~\cite{mikolajczyk2005performance} evaluated the performance of 2D feature descriptors under transformations of rotation, zoom, viewpoint change, blur, JPEG compression, light change and keypoint localization errors. Moreels et al.~\cite{moreels2007evaluation} conducted an evaluation of several groups of 2D feature detectors and descriptors on images captured from the same 3D object with different viewpoints and lighting conditions. Heinly et al.~\cite{Heinly2012Comparative} performed an evaluation of several 2D binary descriptors, aiming at testing their descriptiveness under different feature detectors on several scenes with illumination change, viewpoint change, pure camera rotation and pure scale change. Aans et al.~\cite{Aan2012Interesting} investigated the performance of  several 2D feature detectors on a particular dataset wherein each scene was depicted from 119 camera positions with a range of light directions. In 3D domain, Tombari et al.~\cite{tombari2013performance} compared two categories (i.e., fixed-scale and adaptive-scale) of 3D feature detectors in terms of distinctiveness, repeatability and efficiency under the nuisances of viewpoint changes, clutter, occlusions and noise. Guo et al.~\cite{guo2016comprehensive} tested the descriptiveness, robustness, compactness and efficiency of ten local geometric descriptors on eight datasets with radius variations, varying mesh resolution, Gaussian noise and etc. More relevant to our work is the evaluation performed by Raguram et al.~\cite{Raguram2008A}, where RANSAC and a set of its variants were examined under different ratios of inliers. This paper, compared with~\cite{Raguram2008A}, considers both parametric and non-parametric methods as well as a variety of nuisances for more comprehensive evaluation.
\section{Considered methods}\label{sec:method}
Eight 2D correspondence selection algorithms including two parametric ones, i.e., RANSAC~\cite{fischler1981random} and USAC~\cite {Raguram2013USAC}, and six non-parametric ones, i.e.,  NNSR~\cite{lowe2004distinctive},  ST~\cite{Leordeanu2005A}, GTM~\cite{albarelli2012imposing},  VFC~\cite{Ma2014Robust}, GMS~\cite{bian2017gms}, LPM~\cite{Ma2017Locality}, are considered in our evaluation. Before introducing their theories, we give some general notations for better readability.

Given two images $(I,I^{'})$ to be matched, keypoints and local feature descriptors are computed for them as $(\mathcal{K}, \mathcal{K}^{'})$ and $(\mathcal{F}, \mathcal{F}^{'})$, respectively. This procedure can be accomplished using off-the-shelf detectors and descriptors, e.g., SIFT~\cite{lowe2004distinctive}. To generate initial feature matches $\cal C$, keypoints are matched with each other based on feature similarity, i.e., a correspondence (match) in $\cal C$ is defined as $c=\{\mathbf{x},\mathbf{x}^{'},\arg \mathop {\max }\limits_{{\bf f}^{'}} \; {s_{\mathcal{F}(\mathbf{f},\mathbf{f}^{'})}} \}$ with  $\mathbf{x}\in{\mathcal{K}}$, $\mathbf{x}^{'}\in{\mathcal{K}^{'}}$, $\mathbf{f}\in{\mathcal{F}}$, $\mathbf{f}^{'}\in{\mathcal{F}^{'}}$ and $s_{\mathcal{F}}$ being the feature similarity score. The objective of correspondence selection is digging out the maximum consensus (inlier) subset ${\mathcal{C}_{inlier}}\subseteq \mathcal{C}$. Core principles and computation steps of evaluated algorithms are given as follows.

\textbf{Nearest Neighbor Similarity Ratio~\cite{lowe2004distinctive}.}  NNSR directly utilizes descriptor similarities to remove less distinctive matches. Specifically, the term equaling to the ratio of the closest to the second-closest feature distance to each correspondence
is used as a penalty. Therefore, a correspondence is judged as inlier if
\begin{equation}\label{eq:LRF1}
{\frac{\parallel \mathbf{f}-{\mathbf{f}_{1}^{'}}{{\parallel }_{2}}}{\parallel \mathbf{f}-{\mathbf{f}_{2}^{'}}{{\parallel }_{2}}}\leq t_{nnsr}},
\end{equation}
where $t_{nnsr}\in [0,1]$, ${{\| {\cdot} \|}_{2}}$ hereinafter denotes the $L_2$ norm (this distance metric is suggested in~\cite{lowe2004distinctive}), $\mathbf{f}_{1}^{'}$ and $\mathbf{f}_{2}^{'}$ represent the most and the second most similar feature descriptors of $\mathbf{f}$, respectively. Values of threshold $t_{nnsr}$ and other mentioned thresholds in the following are presented in Table~\ref{tab:para}.

\textbf{Random Sample Consensus~\cite{fischler1981random}.}  RANSAC follows a hypothesize-and-verify framework by repeating procedures of random sampling and checking to maximize the object function. For 2D correspondence selection, the desired parametric model is usually a plane homography matrix or a fundamental matrix. 
Taking the homography matrix as an example, it first randomly samples several correspondences (at least 4) from $\mathcal{C}$ and generates the model hypothesis ${\mathbf{H}}_i$ for those samples at the $i$th iteration. Then, the hypothesis ${\mathbf{H}}_i$ is verified via the following object function
\begin{equation}\label{eq:LRF2}
{O_i=\sum\limits_{ {c} \in \cal C }{h_i({c})}},
\end{equation}
where $h(\cdot)$ is a binary function defined as
\begin{equation}\label{eq:LRF3}
h_i({c})=\left\{ \begin{array}{*{35}{l}}1,\text{ if }{{{\| {\mathbf{x}^{'}}-{\rho} \left({\mathbf{H}_i}\left[ {\begin{array}{*{20}{c}}{\bf x}\\1\end{array}} \right]\right)\|_2}}}\le {{t}_{ransac}} \\
0,\text{ otherwise} \\
\end{array}\right.,
\end{equation}
with $\rho([a_1\;a_2\;a_3]^{T})=[a_1/a_3\;a_2/a_3]^{T}$ and $t_{ransac}$ being a threshold that determines the accuracy of a judged inlier. Above steps are repeated  $n_{ransac}$ times and the model with the maximum object function is selected as the final model ${\mathbf{H}}^{\star}$. Correspondences agreeing with ${\mathbf{H}}^{\star}$ (producing 1 values using Eq.~\ref{eq:LRF3}) are identified as inliers.

\textbf{Spectral Technique~\cite{Leordeanu2005A}.}  ST locates the most reliable element by matrix decomposition. It assumes that the connection among correct matches is much tighter than the one among mismatches. Based on this assumption, ST first builds an adjacency matrix $\mathbf{A}$ as
\begin{equation}\label{eq:LRF5}
{{{a}_{ij}}=\min \left( \frac{{{\| {\mathbf{x}_{i}}-{\mathbf{x}_{j}} \|}_{2}}}{{{\| \mathbf{x}_{i}^{'}-\mathbf{x}_{j}^{'} \|}_{2}}},\text{ }\frac{{{\| \mathbf{x}_{i}^{'}-\mathbf{x}_{j}^{'} \|}_{2}}}{{{\| {\mathbf{x}_{i}}-{\mathbf{x}_{j}} \|}_{2}}} \right)},
\end{equation}
where $a_{ij}\in{\mathbf{A}}$ is the affinity between $c_i$ and $c_j$. 
Second, the principle eigenvector $\mathbf{v}_{st}$ of $\mathbf{A}$ is computed using the singular value decomposition algorithm. Third, the maximum element in $\mathbf{v}_{st}$ is selected as ${v}_{i}$ indicating ${c}_{i}$ being the most reliable correspondence. Fourth, set $v_i$ to zero and remove other components of $\mathcal{C}$ that are in conflict with ${c}_{i}$, i.e.,
\begin{equation}\label{eq:LRF6}
{{{a}_{ij}}\le t_{st}},
\end{equation}
where $t_{st}$ is a predefined threshold. By repeating the third and fourth steps until $\mathcal{C}$ is empty or ${v}_{i}=0$, the correspondences related to all elements selected from $\mathbf{v}_{st}$ are determined as inliers.

\textbf{Game Theory Matching~\cite{albarelli2012imposing}.}  GTM concentrates on extracting correspondences being consistent to the majority of $\mathcal{C}$. Specifically, this strategy interprets the filtering process as a game-theoretic framework where players attempt to obtain high payoffs. At the beginning of this game, every two players extracted from a large population choose a pair of correspondences (served as strategies in this context) from $\mathcal{C}$. Then they will receive a payoff linearly correlated to the coherence between these correspondences. The player who gets high payoffs will receive higher supports. In general, as the game going on, players will prefer to select more reliable correspondences to pursue higher pay-offs.

Given a pair of correspondences $(c_i,c_j)$, the payoff function is defined as
\begin{equation}\label{eq:LRF7}
{{\Pi }_{ij}}={{e}^{-{{\lambda }_{gtm}}\max (\left| {{T}_{i}}({\mathbf{x}_{i}})-{{T}_{j}}({\mathbf{x}_{i}}) \right|,\left| {{T}_{i}}({\mathbf{x}_{j}})-{{T}_{j}}({\mathbf{x}_{j}}) \right|)}},
\end{equation}
where ${\lambda }_{gtm}$ is a selectivity parameter, $\left|{\cdot}\right|$ represents the $L_1$ norm and $T_{i}(\mathbf{x})$ is the similarity transformation estimated by (similarly for $T_{j}(\mathbf{x})$)
\begin{equation}\label{eq:LRFt}
{{T}_{i}}(\mathbf{x})=\rho \left( {\mathbf{H}_{c_i}}\left[ \begin{matrix}
\mathbf{x}  \\
1  \\
\end{matrix} \right] \right),
\end{equation}
where $\mathbf{H}_{c_i}$ is the homographic transformation of $c_i$. Note that this algorithm particularly requires the local affine transformation cue to compute the pay-off function.
Next, the payoff matrix $\mathbf{P}_{gtm}$ with the element in the $i$th row and $j$th column that is defined as
\begin{equation}\label{eq:LRF8}
{{{p}_{ij}}=\left\{ \begin{array}{*{35}{l}}{{\Pi}_{ij}},\text{ if }i\ne j  \\
0,\text{ otherwise}  \\
\end{array}\right.},
\end{equation}
can be generated. The population vector $\mathbf{q}$ is updated by the evolutionary stable states algorithm (ESS's)~\cite{weibull1997evolutionary} as 
\begin{equation}\label{eq:LRF9}
{{{q}_{i}}(k+1)={{q}_{i}}(k)\frac{{{(\mathbf{P}_{gtm}\mathbf{q}(k)})_{i}}}{\mathbf{q}{{(k)^{T}}}\mathbf{P}_{gtm}\mathbf{q}(k)}},
\end{equation}
where $q_{i}$ represents the element in the $i$th row of $\mathbf{q}$ and $k$ is the iteration number. After $n_{gtm}$ iterations, a correspondence $c_i$ is identified as inlier if its corresponding $q_{i}$ is higher than a threshold $t_{gtm}$.

\textbf{Universal RANSAC~\cite{Raguram2013USAC}.}  USAC integrates a universal framework for RANSAC, where each original step is optimized by referring to the advantages of previous parametric approaches such as PROSAC~\cite{Chum2005Matching}, SPRT test~\cite{matas2005randomized} and LO-RANSAC~\cite{Chum2003Locally}. Further, this algorithm inserts degeneracy and local optimization processes after generating the minimal-sample model.

During the sampling step, USAC uses a weighted sampling algorithm named PROSAC~\cite{Chum2005Matching}, where the initial correspondences are reordered at first based on the descending sort order of brute-force matching scores and correspondences with higher scores are preserved. At the checking stage of the model (homography matrix or fundamental matrix), a correspondence is judged as inlier by Eq.~\ref{eq:LRF3} with the threshold $t_{\mathbf{H}}$ or by the equation
\begin{equation}\label{eq:LRF3'}
{\frac{{{\left( {\mathbf{y}^{'T}}{\mathbf{F}_{i}}\mathbf{y} \right)}^{2}}}{\left( {\mathbf{F}_{i}}\mathbf{y} \right)_{1}^{2}+\left( {\mathbf{F}_{i}}\mathbf{y} \right)_{2}^{2}+\left( \mathbf{F}_{i}^{T}{\mathbf{y}^{'}} \right)_{1}^{2}+\left( \mathbf{F}_{i}^{T}{\mathbf{y}^{'}} \right)_{2}^{2}}}\le {{t}_\mathbf{F}},
\end{equation}
where $\mathbf{F}_{i}$ is the $i$th hypothetic fundamental matrix, $t_\mathbf{F}$ is the threshold and ${\bf y} = {[\begin{array}{*{20}{c}}
	{\bf x}&1
	\end{array}]^T}$ (similarly for $\mathbf{y}^{'}$).
After generating the minimal-sample model, USAC verifies whether the model is interesting by the SPRT test~\cite{matas2005randomized}. The likelihood ratio can be computed after evaluating $n$ correspondences as
\begin{equation}\label{eq:LRF10}
{{{\xi }_{n}}=\prod\limits_{i=1}^{n}{\frac{p({{r}_{i}}|{\mathbf{H}_{b}})}{p({{r}_{i}}|{\mathbf{H}_{g}})}}},
\end{equation}
where $\mathbf{H}_{g}$ and $\mathbf{H}_{b}$ respectively represent a ``good'' model and a ``bad'' model, $r_{i}$ is equal to 1 if $c_i$ is consistent with the generated model and 0 otherwise, $p(1|\mathbf{H}_g)$ is approximated by the inlier ratio and $p(r_i|\mathbf{H}_b)$ follows a Bernoulli distribution. If the $\xi_n$ is higher than an adaptive threshold, the model will be discarded. When fitting the fundamental matrix by epipolar geometry constraint, USAC utilizes DEGENSAC~\cite{Chum2005Two} for degeneracy. It assumes that the generated model is often incorrect in the context of images containing a dominant scene plane. Accordingly, DEGENSAC employs a homographic transformation to reject the generated fundamental model if there are five or more sampled correspondences lying on the same plane. Eventually, USAC adds a local optimization (LO-RANSAC~\cite{Chum2003Locally}) to refine the minimal-sample model. It re-samples correspondences only from the set of selected inliers and refines the previous model by the sampling subset. This whole process is repeated until achieving confidence in solution or iterations reach the upper bound $n_{usac}$.

\textbf{Vector Field Consensus~\cite{Ma2014Robust}.}  VFC interpolates a vector field where the posteriori probability of a correct correspondence is estimated by the Bayes rule.

For a correspondence $c_i$, the transformation to a motion field is expressed as $(\mathbf{x}_{i},\text{ }\mathbf{x}_{i}^{'})\to(\mathbf{u}_{i},\text{ }\mathbf{v}_{i})$, where $\mathbf{u}_{i}=\mathbf{x}_{i}$ and $\text{ }\mathbf{v}_{i}=\mathbf{x}_{i}^{'}-\mathbf{x}_{i}$. In this motion field, VFC holds the assumption that the noise around inliers indicated by $z_{i}=1$ follows the Gaussian distribution and the noise around outliers indicated by $z_{i}=0$ follows the uniform distribution. Thus, the probability is a mixture model given by
\begin{equation}\label{eq:LRF11}
{p(\mathcal{U}|\mathcal{V},\mathbf{\theta} )=\prod\limits_{i=1}^{N}{(\frac{\gamma}{{{(2\pi {{\sigma }^{2}})}^{D/2}}}{{e}^{-\frac{{{\| {\mathbf{v}_{i}}-\mathbf{f}_{vfc}({\mathbf{u}_{i}}) \|}_{2}}}{2{{\sigma }^{2}}}}}+\frac{1-\gamma }{a})}},
\end{equation}
where $\mathcal{\theta}=\left\{{\mathbf{f}_{vfc},\sigma^{2},\gamma}\right\}$ is a set of unknown parameters, $\mathbf{f}_{vfc}$ is the vector field expected to be recovered, $\gamma$ is the mixing coefficient of the mixture probability model, i.e, $p(z_{i}=1)=\gamma$, $\mathcal{U}$ and $\mathcal{V}$ respectively are sets of $\mathbf{u}$ and $\mathbf{v}$, $\sigma$ is the uniform standard deviation of Gaussian distribution, $\frac{1}{a}$ is the probability density of the uniform distribution and $D$ is the dimension of the output space. VFC employs the EM~\cite{Dempster1977Maximum} algorithm to deal with the maximum likelihood estimation with latent variables. At E-step, the diagonal element of a diagonal matrix $\mathbf{P}$, i.e.,  $p_{i}=p(z_{i}=1|\mathbf{u}_{i},\mathbf{v}_{i},\theta)$, can be computed by the Bayes rule
\begin{equation}\label{eq:LRF12}
{{p_i} = \frac{{\gamma {e^{ - \frac{{{{\| {\mathbf{v}_{i} - \mathbf{f}_{vfc}(\mathbf{u}_{i})} \|}_2}}}{{2{\sigma ^2}}}}}}}{{\gamma {e^{ - \frac{{{{\| {\mathbf{v}_{i} - \mathbf{f}_{vfc}(\mathbf{u}_{i})} \|}_2}}}{{2{\sigma ^2}}}}} + (1 - \gamma )\frac{{{{(2\pi {\sigma ^2})}^{D/2}}}}{a}}}}.
\end{equation}
At M-step, a coefficient matrix $\mathbf{C}$ is created first by
\begin{equation}\label{eq:LRF13}
{(\mathbf{K}_{Gauss}+\lambda_{vfc}{\sigma^{2}}\mathbf{P}^{-1})\mathbf{C}=\mathcal{V}},
\end{equation}
where $\mathbf{K}_{Gauss}$ is a matrix consisting of the Gaussian kernel $k(\mathbf{u}_{i},\mathbf{u}_{j}) = {e^{ - \beta {{\| {\mathbf{u}_{i} - \mathbf{u}_{j}} \|}_2}}}$ and $\lambda_{vfc}$ is a regularization constant. Second, the vector field $\mathbf{f}_{vfc}$ is estimated by
\begin{equation}\label{eq:LRF14}
{\mathbf{f}_{vfc}(\mathbf{u}) = \sum\limits_{i = 1}^{N} {k}(\mathbf{u},\mathbf{u}_{i})\mathbf{c}_{i}},
\end{equation}
where $\mathbf{c}_i\in{\mathbf{C}}$. Third, values of $\sigma^{2}$ and $\gamma$ are updated by 
\begin{equation}\label{eq:LRF15}
{{\sigma ^2} = \frac{{{{(\mathcal{V} - \mathcal{F}_{vfc})}^T}\mathbf{P}(\mathcal{V} - \mathcal{F}_{vfc})}}{{D \cdot \text{tr}(\mathbf{P})}}},
\end{equation}
and
\begin{equation}\label{eq:LRF16}
{\gamma  = \text{tr}(\mathbf{P})/N},
\end{equation} 
where $\mathcal{F}_{vfc}={({\mathbf{f}_{vfc}(\mathbf{u}_{1})}^T,...{\mathbf{f}_{vfc}(\mathbf{u}_{N})}^T)}^T$.
The E-step and M-step are repeated until parameters are converged. Finally, the inlier set is generated as
\begin{equation}\label{eq:LRF17}
{\mathcal{C}_{inlier}=\left\{{c}_i:p_i>t_{vfc}\right\}},
\end{equation} 
where $t_{vfc}$ is a predefined threshold.

\textbf{Grid-based Motion Statistics~\cite{bian2017gms}.}  GMS proves that besides feature descriptiveness,  feature number also contributes to the quality of correspondences. It supposes that the quantity of correspondences in a small neighborhood around a true match is larger than that around a false match under the smooth motion. In over-large neighborhoods, regions are divided into multiple small region pairs where distributions of correspondence number are approximated by Binomial distributions. Given a correspondence $c_i$, the joint statistical distribution is modeled as
\begin{equation}\label{eq:LRF18}
{{{S}_{i}}\sim{\ }\left\{ \begin{array}{*{35}{l}}
	B(Kn,{{p}_{t}}),\text{ if }{{{c}}_{i}}\text{ is inlier}  \\
	B(Kn,{{p}_{f}}),\text{ otherwise}  \\
	\end{array} \right.},
\end{equation}
where $S_{i}$ is the total number of correspondences in a region pair ($a$, $b$) around ${c}_i$, $K$ is the quantity of small region pairs, $p_t$ is the probability that the nearest neighbor of each keypoint in $a$ is located in $b$ under the condition that $a$ and $b$ view the same location, and $p_f$ is the probability provided that $a$ and $b$ view the different locations. $p_t$ and $p_f$ can be estimated by
\begin{equation}\label{eq:LRF19}
{p_t=\delta+(1-\delta)\zeta{m}/M},
\end{equation}
and
\begin{equation}\label{eq:LRF20}
{p_f=\zeta(1-\delta)(m/M)},
\end{equation}
where $\delta$ is the probability of a correspondence being correct, $m$ is the amount of keypoints in region $b$, $M$ is the size of $\mathcal{K}^{'}$ in $I^{'}$, and $\zeta$ is a factor added to balance deviations caused by repeated structures. A quantitative score is next designed to evaluate the distinction between two distributions as
\begin{equation}\label{eq:LRF21}
{{P}=\frac{{{m}_{t}}-{{m}_{f}}}{{{s}_{t}}+{{s}_{f}}}},
\end{equation}
where $m$ is the mean value and $s$ is the standard deviation. This equation can be simplified as
\begin{equation}\label{eq:LRF22}
{{P}\propto \sqrt{Kn}},
\end{equation}
where the distinction is positive correlated to the number of correspondences.

In addition, to incorporate this approach into a real-time system, a fast gird-based score estimator is developed as follows. First, $I$ and $I^{'}$ are divided into $20\times20$ non-overlapping cells. Second, for each cell in $I$, the cell containing the maximum amount of correspondences is grouped in $I^{'}$. Third, in cell-pair $(i,j)$ as well as its small neighborhoods (eight cell-pairs), $S_{ij}$ is estimated as
\begin{equation}\label{eq:LRF23}
{{{S}_{ij}}=\sum\limits_{k=1}^{K=9}{\left| {{\chi }_{{{i}^{k}}{{j}^{k}}}} \right|}},
\end{equation}
where $\left|{\chi}\right|$ is the amount of correspondences in the cell-pair $(i^{k},j^{k})$. All correspondences in $(i,j)$ are judged as inliers if $S_{ij}>t_{gms}$, where $t_{gms}$ is a threshold approximated by $\alpha \sqrt{n_i}$ with $\alpha$ being a given parameter and $n_i$ being the average (of the nine cell-pairs) amount of correspondences.

\textbf{Locality Preserving Matching~\cite{Ma2017Locality}.}  This algorithm removes mismatches by digging out the local geometric structure consensus. With the hypothesis that the local structure around a correspondence may not change freely, a cost function is defined as
\begin{equation}\label{eq:LRF24}
\begin{aligned}
L({\mathcal{K}_{inlier}},\lambda_{lpm} )=\sum\limits_{i\in {\mathcal{K}_{inlier}}}{\left( \sum\limits_{j|{\mathbf{x}_{j}}\in {\mathcal{K}_{{\mathbf{x}_{i}}}}}{{{\left( d\left( {\mathbf{x}_{i}},{\mathbf{x}_{j}} \right)-d\left( \mathbf{x}_{i}^{'},\mathbf{x}_{j}^{'} \right) \right)}^{2}}} \right.}\\
+\left. \sum\limits_{j|{\mathbf{x}_{j}}^{'}\in {\mathcal{K}^{'}_{\mathbf{x}_{i}^{'}}}}{{{\left( d\left( {\mathbf{x}_{i}},{\mathbf{x}_{j}} \right)-d\left( \mathbf{x}_{i}^{'},\mathbf{x}_{j}^{'} \right) \right)}^{2}}} \right)+\lambda_{lpm} \left(N-\left| {\mathcal{K}_{inlier}} \right| \right),
\end{aligned}
\end{equation}
where $\lambda_{lpm}$ is a regularization parameter, $d$ is the Euclidean distance between two keypoints, $\mathcal{K}_{\mathbf{x}_i}$ and $\mathcal{K}_{\mathbf{x}_{i}^{'}}^{'}$ respectively are sets of the $k$ nearest neighbors of $\mathbf{x}_i$ and $\mathbf{x}_{i}^{'}$, $N$ is the size of $\mathcal{K}$, and $\mathcal{K}_{inlier}$ is an inlier subset of $\mathcal{K}$.
Under non-rigid transformations such as deformation, the absolute distance in Eq.~\ref{eq:LRF24} may not be preserved well. To address this issue, LPM converts the cost function to
\begin{equation}\label{eq:LRF25}
\begin{aligned}
L(\mathcal{W},\lambda_{lpm})=\sum\limits_{i=1}^{N}{{{w}_{i}}\left(\sum\limits_{j|{\mathbf{x}_{j}}\in {\mathcal{K}_{{\mathbf{x}_{i}}}}}{d\left( \mathbf{x}_{i}^{'},\mathbf{x}_{j}^{'} \right)} \right.}\\
+\left.\sum\limits_{j|\mathbf{x}_{j}^{'}\in {\mathcal{K}^{'}_{\mathbf{x}_{i}^{'}}}}{d\left( {\mathbf{x}_{i}},{\mathbf{x}_{j}} \right)} \right)+\lambda_{lpm} \left( N-\sum\limits_{i=1}^{N}{{{w}_{i}}} \right),
\end{aligned}
\end{equation}
where $\mathcal{W}$ is a set of indicators where $w_i=1$ indicates the inlier and $w_i=0$ otherwise. This equation can be further reorganized by merging the related items of $w_i$ as
\begin{equation}\label{eq:LRF26}
L(\mathcal{W},\lambda_{lpm} )=\sum\limits_{i=1}^{N}{{{w}_{i}}\left( {{l}_{i}}-\lambda_{lpm}  \right)}+\lambda_{lpm} N,
\end{equation}
where 
\begin{equation}\label{eq:LRF27}
l_i=\sum\limits_{j|{\mathbf{x}_{j}}\in {\mathcal{K}_{{\mathbf{x}_{i}}}}}{d\left( \mathbf{x}_{i}^{'},\mathbf{x}_{j}^{'} \right)}+\sum\limits_{j|\mathbf{x}_{j}^{'}\in {\mathcal{K}^{'}_{\mathbf{x}_{i}^{'}}}}{d\left( {\mathbf{x}_{i}},{\mathbf{x}_{j}} \right)}
\end{equation}
is a constraint item measuring the local geometric structure changes. With the objective of minimizing the cost function, a correspondence with the cost, i.e., ${l_i}>\lambda_{lpm}$, is negative. For this purpose, the correct correspondence set is determined by
\begin{equation}\label{eq:LRF28}
{{w}_{i}}=\left\{ \begin{array}{*{35}{l}}
1,\text{ if }{{l}_{i}}\le \lambda_{lpm}   \\
0,\text{ otherwise}  \\
\end{array}\right.,i=1,\text{ }...\text{ },N.
\end{equation}

\section{Experimental setup}\label{sec:exper}
The experimental setup is introduced detailedly in this section. First, we list implementations and parameter settings of the evaluated methods. Second, characteristics of four datasets, the experimental criteria and the deployment are formulated.
\subsection{Implementations}
In our experiments, Hessian-affine detector~\cite{Mikolajczyk2004Hessian} and SIFT descriptor~\cite{lowe2004distinctive} (a popular detector-descriptor combination~\cite{moreels2007evaluation}) are employed in default for image keypoint detection and description. Notice that another reason for using the Hessian-affine detector is that the evaluated GTM method requires local affine information, while we also consider different detector-descriptor combinations in Sect.~\ref{sub:sub3}. The initial correspondence set $\mathcal{C}$ is generated by brute-force matching, i.e., greedy comparison of two feature sets. Parameters and implementations for each algorithm are listed in Table~\ref{tab:para}.
\begin{table}[t]\scriptsize
	\renewcommand{\arraystretch}{1}
	\caption{Parameter settings and implementations of evaluated algorithms, where \textit{pix} represents the pixel unit.}
	\label{tab:para}
	\centering
	\begin{tabular}	{lccccc}
		\hline
		No. & Algorithm & Implementation & Parameters & Setting\\
		\hline
		1 & NNSR~\cite{lowe2004distinctive} & OPENCV & $t_{nnsr}$ & Adaptive~\cite{otsu1979threshold} \\
		\hline
		2 & RANSAC~\cite{fischler1981random} & OPENCV & $t_{ransac}$ & 10\textit{pix}  \\
		  &        &                 & $n_{ransac}$ & 2000 \\ 
		\hline
		3 & ST~\cite{Leordeanu2005A} & MATLAB  & $t_{st}$ & 0.3 \\
		\hline
		4 & GTM~\cite{albarelli2012imposing} & OPENCV & $t_{gtm}$ & Adaptive~\cite{otsu1979threshold} \\
		  &     &                         & $n_{gtm}$ & 100 \\
		  &     &                         & $\lambda{gtm}$ & 0.0001 \\
		\hline
		5 & USAC~\cite {Raguram2013USAC} & OPENCV & $n_{usac}$ & 850000  \\
		  &      &                & $t_{\mathbf{H}}$ & 10\textit{pix}   \\
		  &      &                & $t_{\mathbf{F}}$ & 1.5\textit{pix}   \\   
		\hline
		6 & VFC~\cite{Ma2014Robust} & OPENCV & $\beta $ & 0.1 \\
		  &     &                 & $\lambda_{vfc} $ & 3  \\
		  &     &                 & $t_{vfc} $ & 0.75  \\
		  &     &                 & $\gamma $ & 0.9  \\		
		\hline
		7 & GMS~\cite{bian2017gms} & OPENCV & $\alpha $ & 4  \\
		\hline
		8 & LPM~\cite{Ma2017Locality} & MATLAB	 & $\lambda_{lpm}$ & 6   \\
		  &     &                          & $k$ & 4  \\	  	
		\hline	  		  
	\end{tabular} 
\end{table}
Notably, for NNSR and GTM we set $t_{nnsr}$ and $t_{gtm}$ adaptively using the OTSU~\cite{otsu1979threshold} algorithm to reduce thresholding errors as proper thresholds may vary in different scenarios even in different images.

All those methods are implemented in OPENCV or MATLAB with a PC equipped with a 3.2GHz processor and 8GB memory.
\subsection{Datasets}
\begin{figure}[t] 
	\centering
	\subfigure[VGG]
	{%
		\includegraphics[width=0.23\linewidth]{figure/vgg1.pdf} 
		\includegraphics[width=0.23\linewidth]{figure/vgg2.pdf} 
		\includegraphics[width=0.23\linewidth]{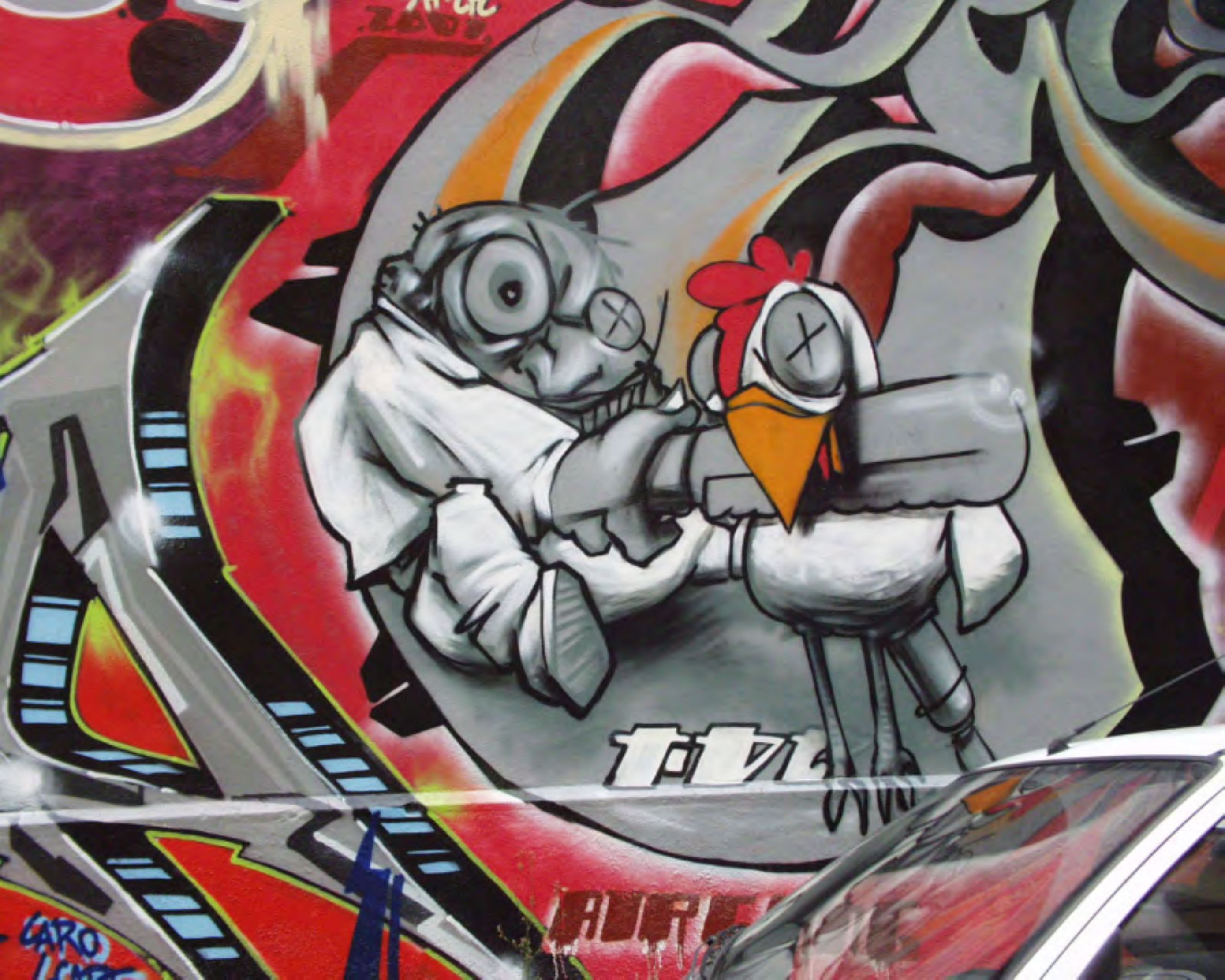} 
		\includegraphics[width=0.23\linewidth]{figure/vgg4.pdf}
	}
	\subfigure[Symbench]
	{%
		\includegraphics[width=0.23\linewidth]{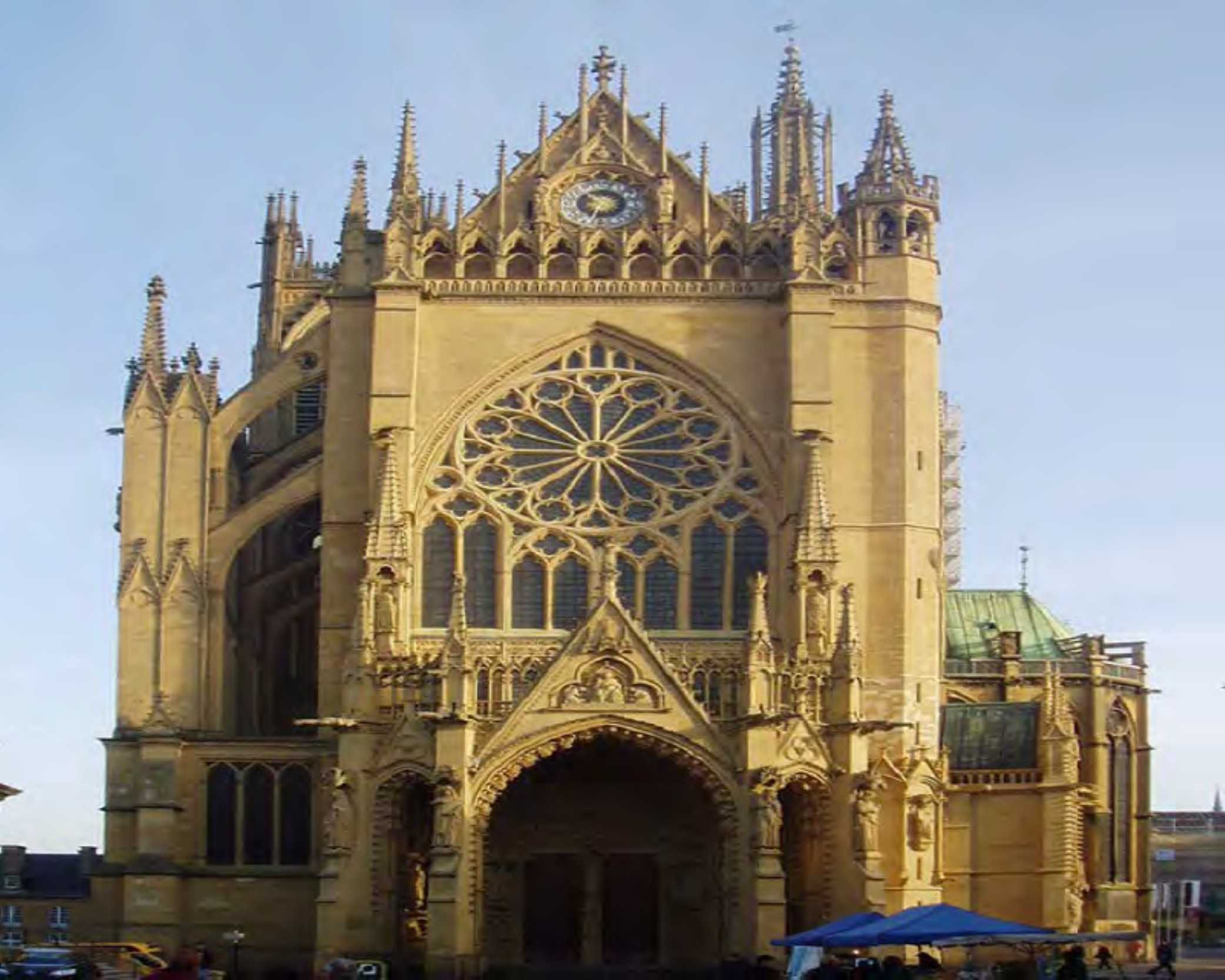} 
		\includegraphics[width=0.23\linewidth]{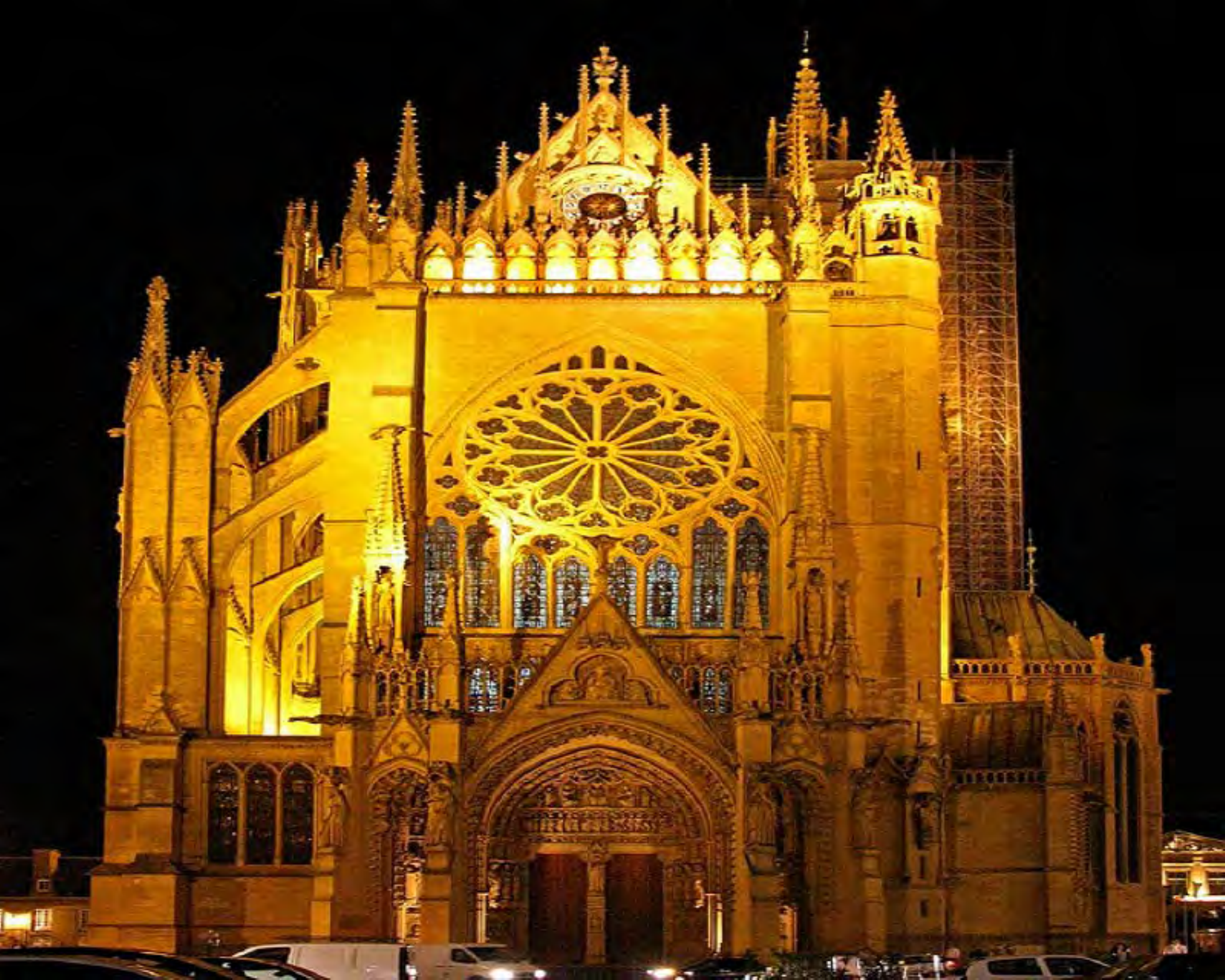} 
		\includegraphics[width=0.23\linewidth]{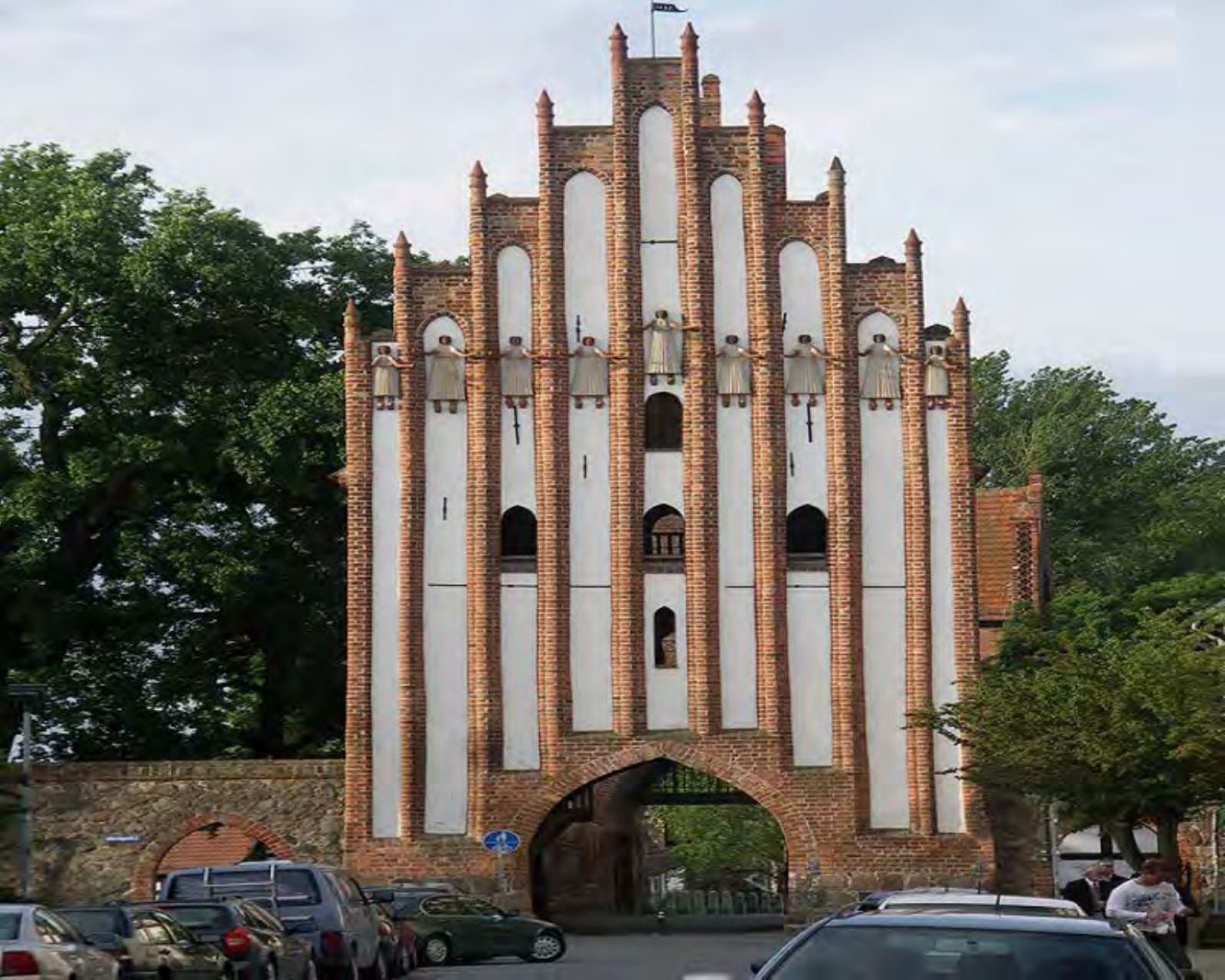} 
		\includegraphics[width=0.23\linewidth]{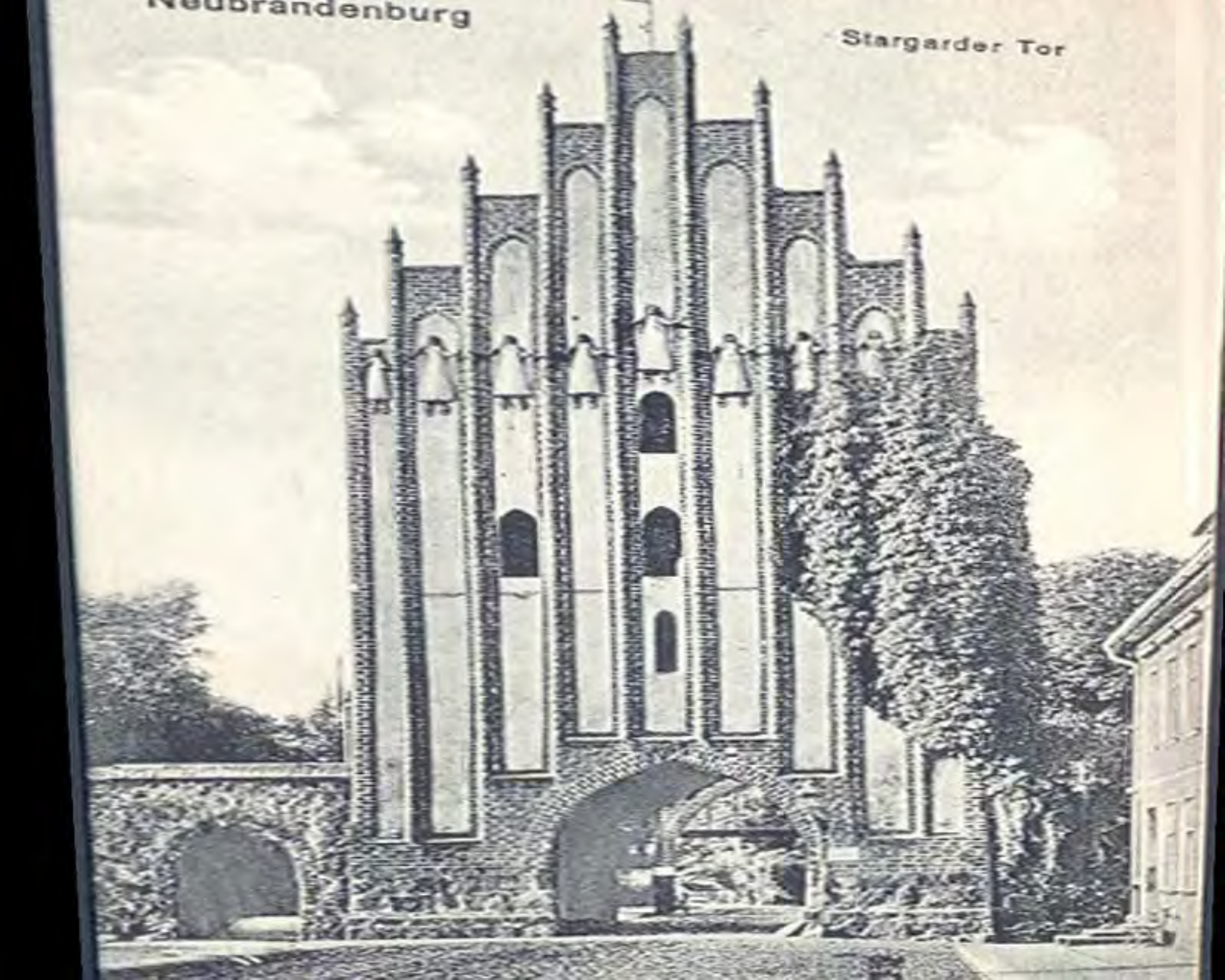} 
	}
	\subfigure[Heinly]
	{%
		\includegraphics[width=0.23\linewidth]{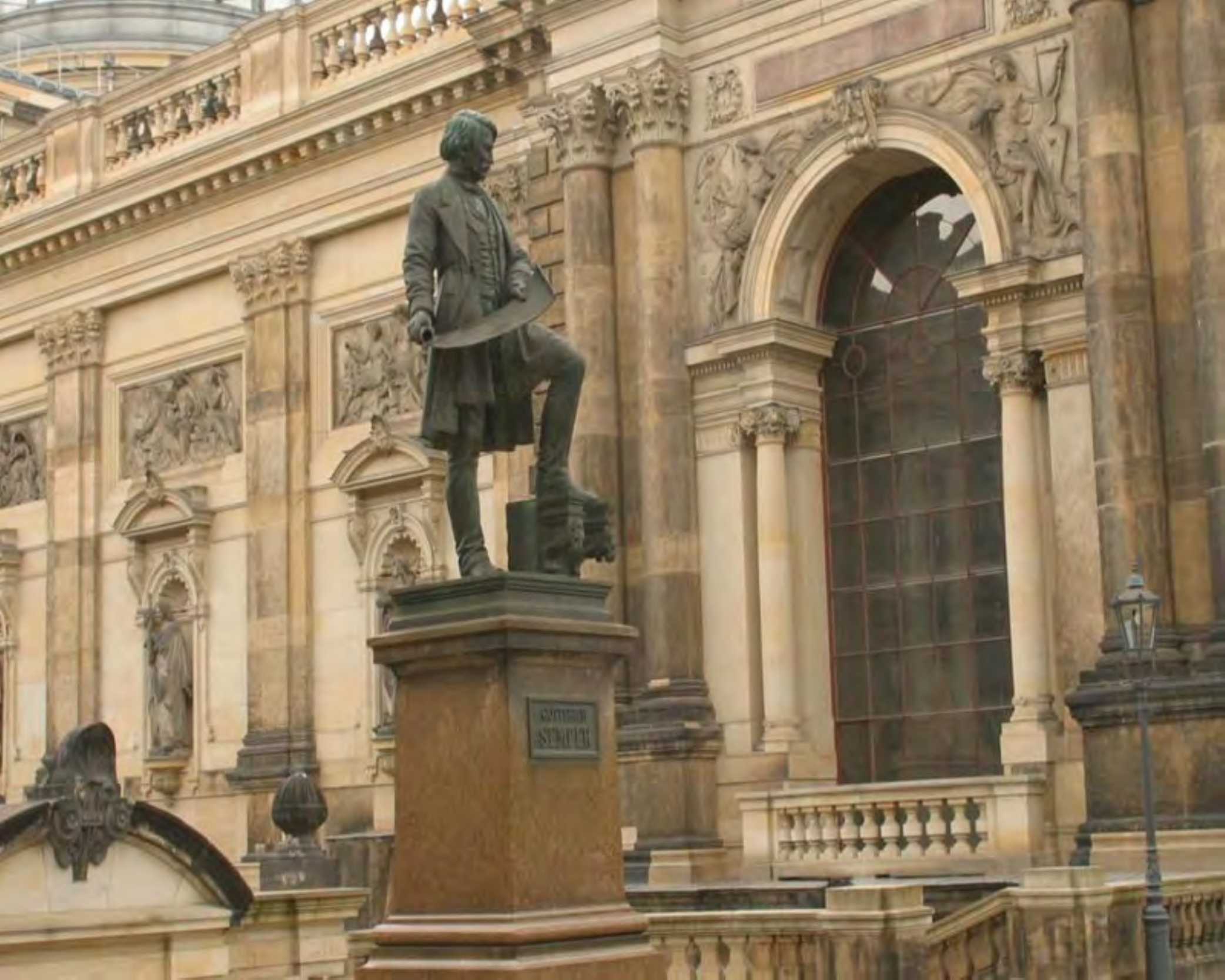} 
		\includegraphics[width=0.23\linewidth]{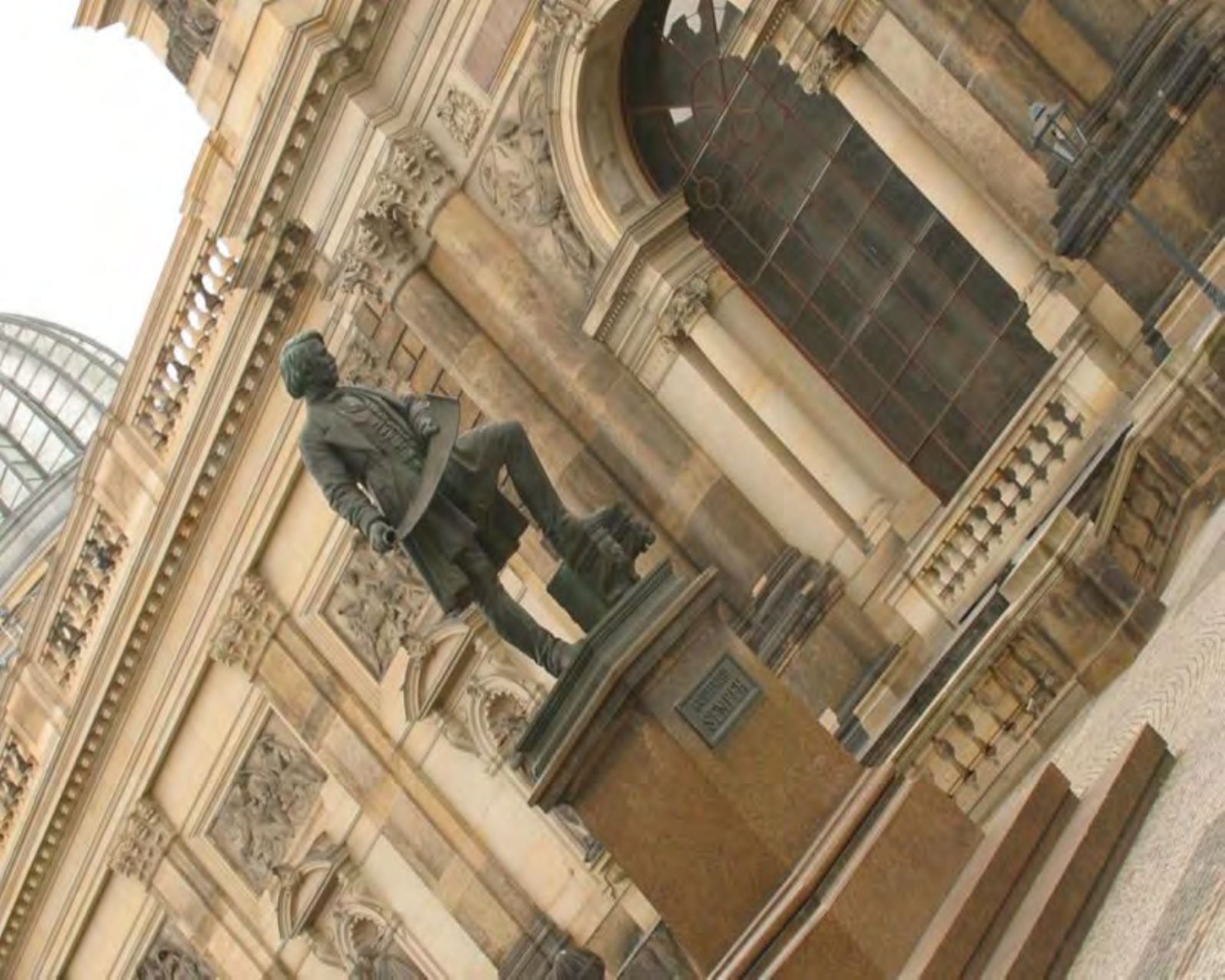} 
		\includegraphics[width=0.23\linewidth]{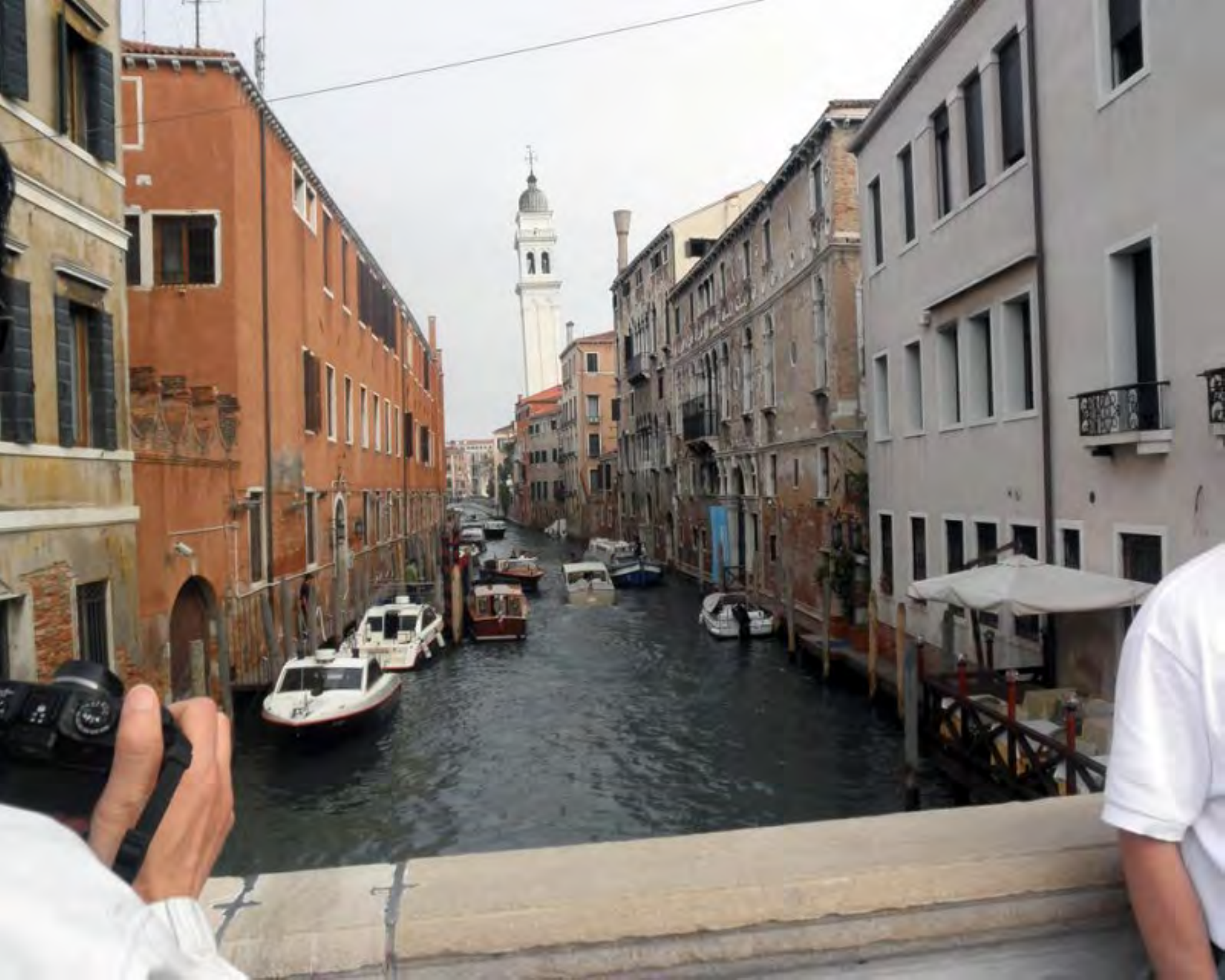} 
		\includegraphics[width=0.23\linewidth]{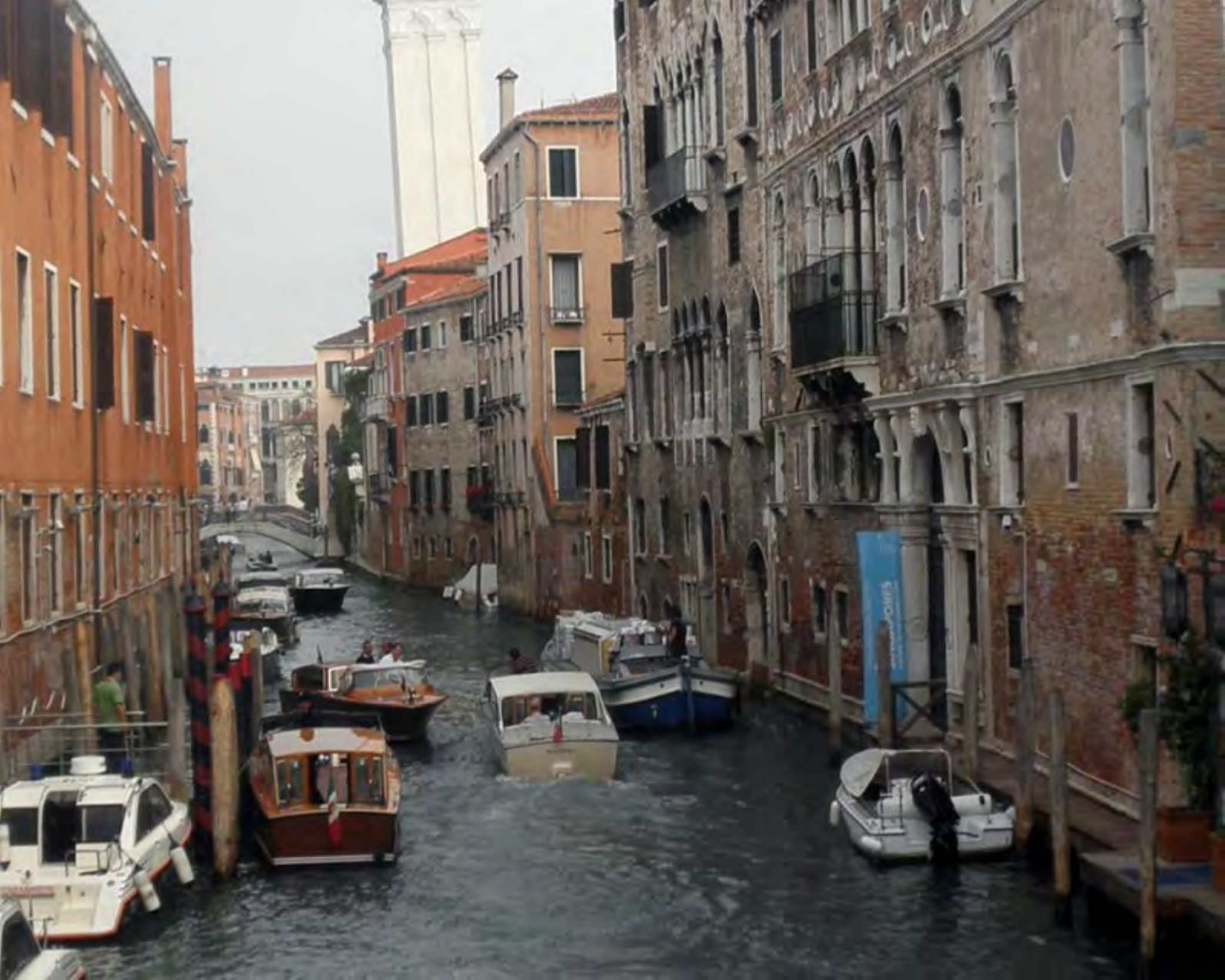}
	}
	\subfigure[AdelaideRMF]
	{%
		\includegraphics[width=0.235\linewidth]{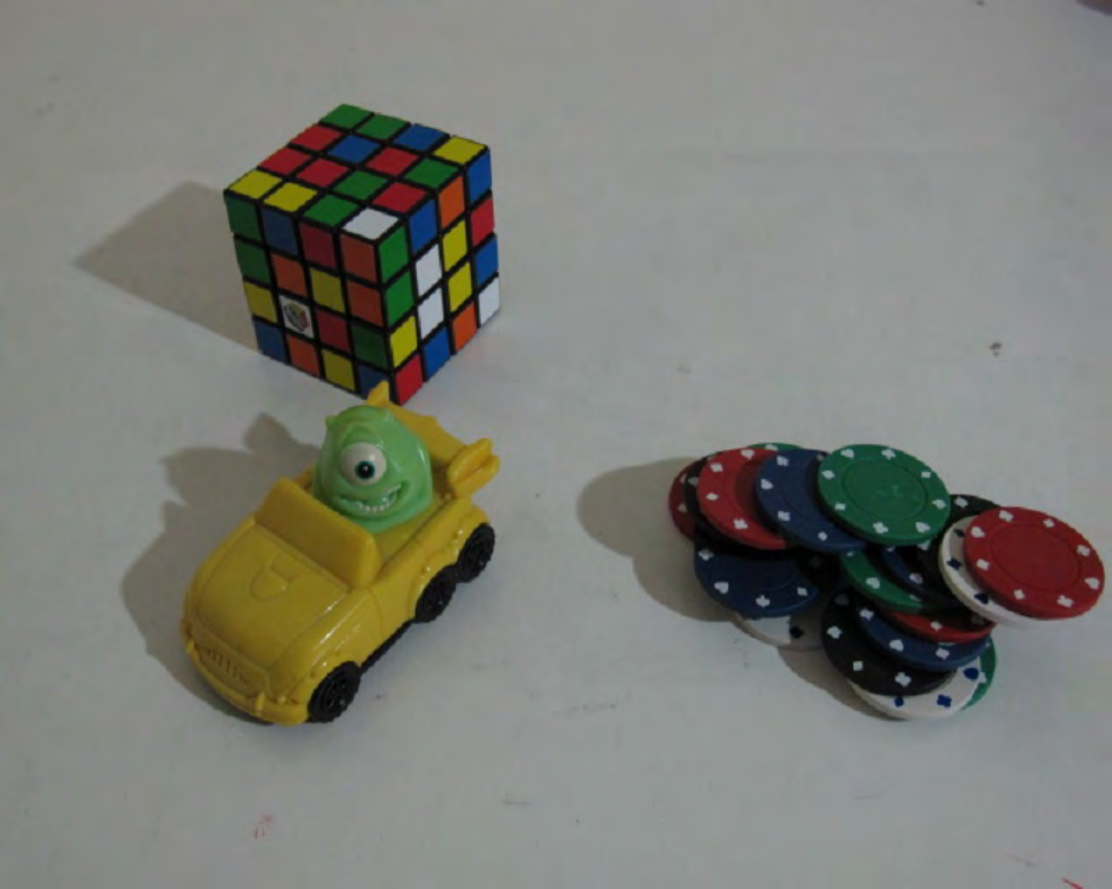} 
		\includegraphics[width=0.235\linewidth]{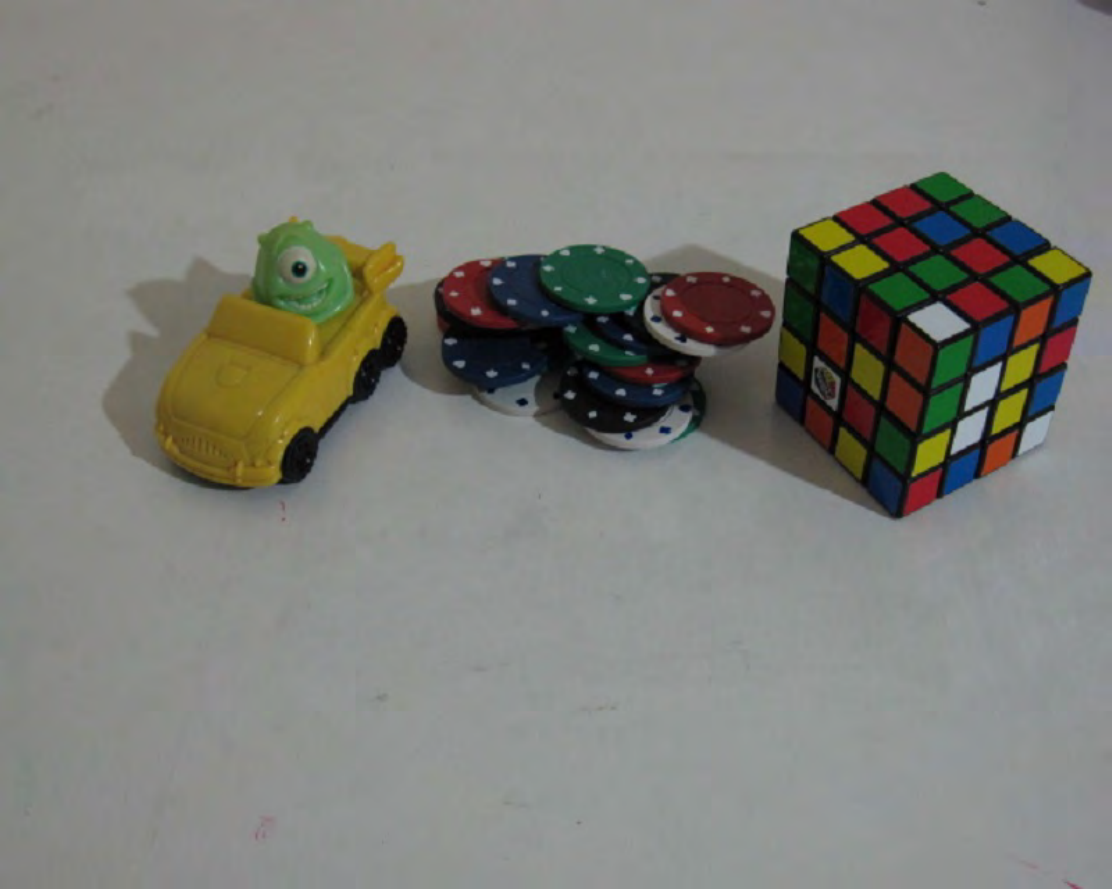} 
		\includegraphics[width=0.235\linewidth]{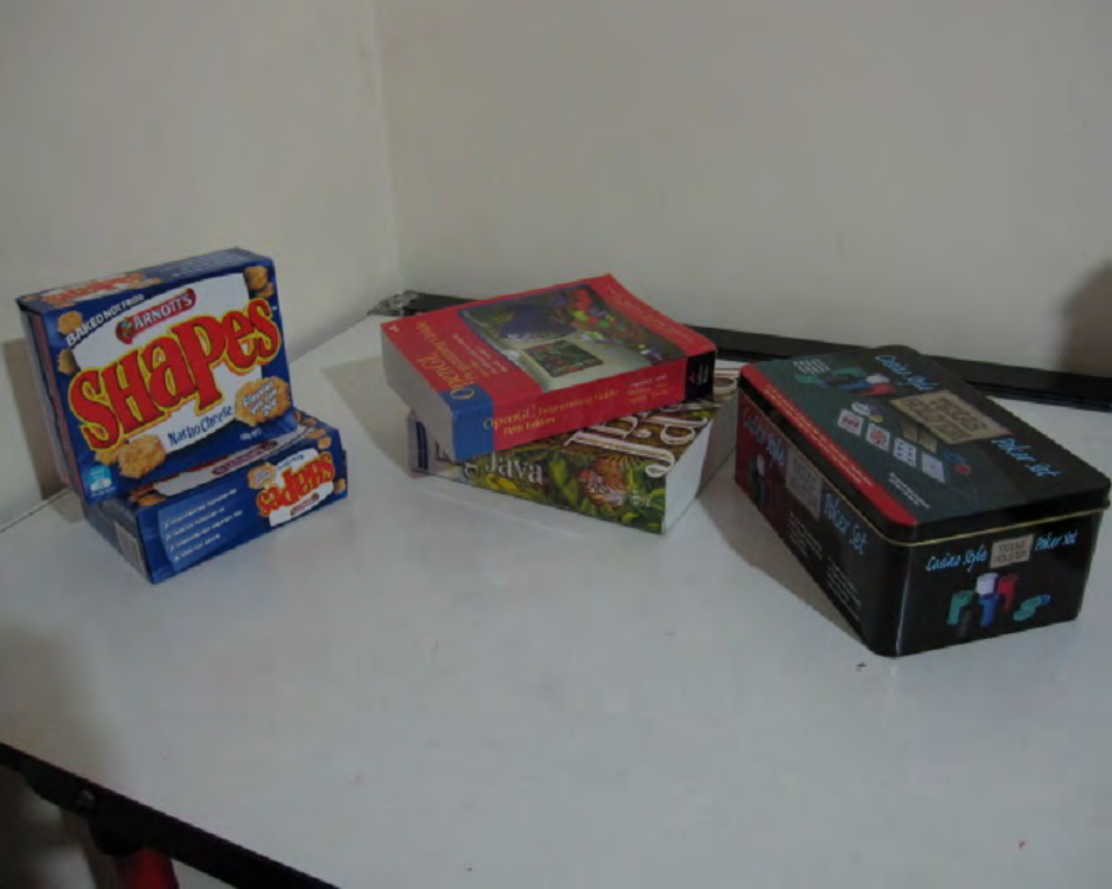} 
		\includegraphics[width=0.235\linewidth]{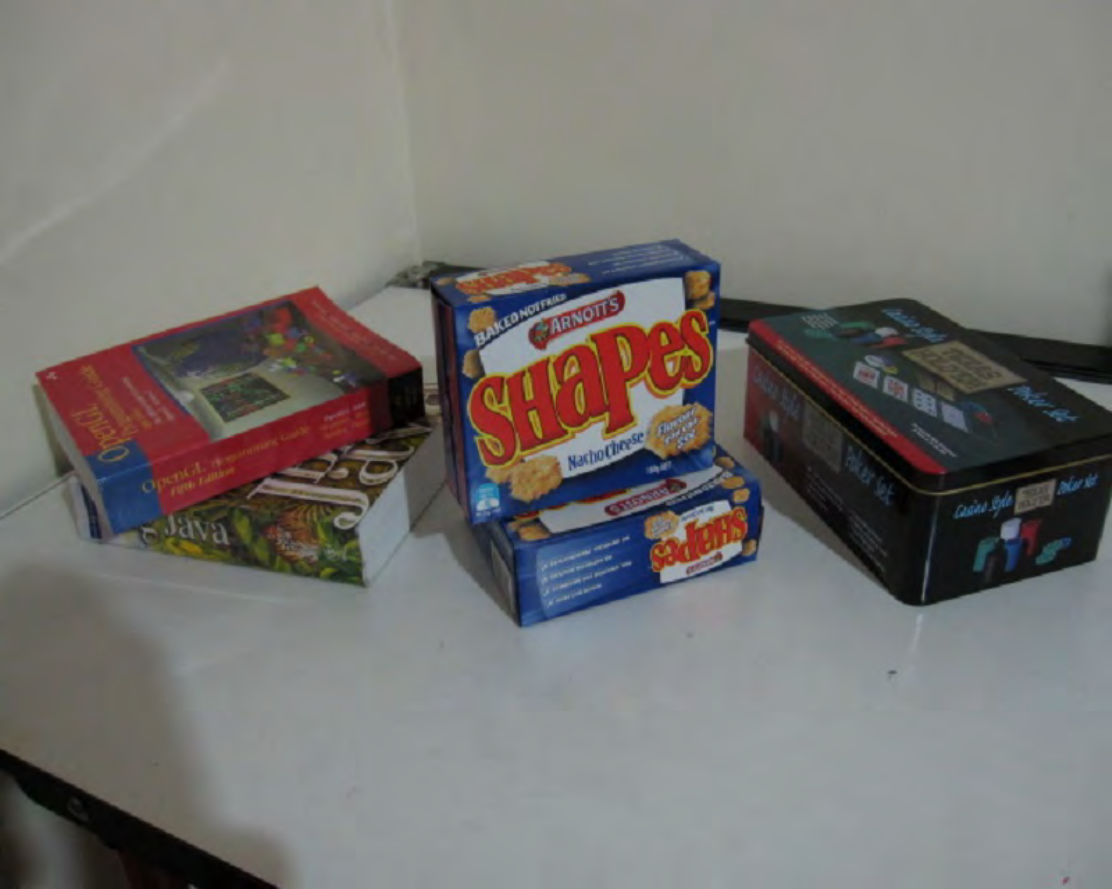} 
	}	 
	\caption{Sampled image pairs from four experimental datasets. (a) VGG~\cite{Mikolajczyk2005A}, (b) Symbench~\cite{Snavely2012Image}, (c) Heinly~\cite{Heinly2012Comparative}, and (d) AdelaideRMF~\cite{Wong2011Dynamic}.}
	\label{fig:dataset}
\end{figure}
We perform our experiments on four datasets, i.e., VGG~\cite{Mikolajczyk2005A}, Symbench~\cite{Snavely2012Image}, Heinly~\cite{Heinly2012Comparative}, and AdelaideRMF~\cite{Wong2011Dynamic}. Exemplar images from these datasets and a brief summarization of their inherited nuisances are shown in Fig.~\ref{fig:dataset} and Table~\ref{tab:dataset}, respectively. 

\begin{table}[t]\scriptsize
	\renewcommand{\arraystretch}{1}
	\caption{Properties of the experimental datasets.}
	\label{tab:dataset}
	\centering
	\begin{tabular}	{lccc}
	\hline
	Dataset & Challenges & Matching pairs 	\\
	\hline
	VGG~\cite{Mikolajczyk2005A} & Zoom, rotation, blur, viewpoint change, & 40	\\
	& light change and JPEG compression & \\
	
	Symbench~\cite{Snavely2012Image} &  Light change, & 46 \\
	& different rendering styles & \\

	Heinly~\cite{Heinly2012Comparative} & Zoom and rotation & 29 \\
	AdelaideRMF~\cite{Wong2011Dynamic} & Multi-structures,  & 38 \\
	& viewpoint change & \\
	\hline
	\end{tabular} 
\end{table}

\textbf{The VGG dataset~\cite{Mikolajczyk2005A}.}  VGG is a hybrid dataset involving eight scenes. Each scene consists of six images with the first image being the reference one with respect to the others. Challenges including blur, viewpoint change, zoom, rotation, light change, and JPEG compression exist in this dataset. The ground-truth is the homography matrix $\mathbf{H}$, indicating that the transformation between two images on each scene satisfies the plane homographic constraint.

\textbf{The Symbench dataset~\cite{Snavely2012Image}.}  The Symbench dataset is composed of 46 image pairs. Each pair includes the same object with light change or different rendering styles. The homographic transformation $\mathbf{H}$ of each image pair is given as the ground-truth.

\textbf{The Heinly dataset~\cite{Heinly2012Comparative}.}  The Heinly dataset comprises images with dense or sparse viewpoint change, illumination, pure large-scale zoom or rotation. Considering that nuisances of viewpoint change and illumination have been covered in the other three datasets, we choose a subset of Heinly containing 29 pairs of image shot on 4 scenes with the specific challenges, i.e., pure zoom or rotation, to perform a more targeted test. The ground-truth is provided as the homographic transformation.

\textbf{The AdelaideRMF dataset~\cite{Wong2011Dynamic}.}  AdelaideRMF includes 38 pairs of image with viewpoint change and multi-structures. The keypoint coordinates of initial correspondences are provided and the ground-truth correspondences are manually labeled in this dataset. 

Motivations of employing these datasets can be summarized as: (i) The eight scenes in the VGG dataset cover a peculiar wide range of interferences such as the rigid/non-rigid transformation and image quality variation. Both the generality to diverse conditions and the robustness to a specific nuisance can be assessed on this dataset. (ii) The focus of Symbench is the image quality variation caused by light change and different rendering styles that give rise to potential errors of feature detection and description. The performance in the context of image quality variation can be specifically evaluated. (iii) The subset of Heinly is selected with the aim of testing the performance under the condition of a geometrical structure deformation (pure zoom or rotation). (iv) AdelaideRMF aims at evaluating the performance of those correspondence selection algorithms where plane homographic constraint fails and multiple consistent correspondence sets are involved due to multi-structures. All above peculiarities make the evaluation benchmarks complementary to each other and allow us to find prominent algorithms under a specific nuisance.
\subsection{Criteria}  
The performance of evaluated algorithms is measured via precision, recall and F-measure as in~\cite{lin2014bilateral,bian2017gms,Ma2017Locality}. First, we denote the selected correspondence set, the ground-truth correspondence set and the correct subset in the selected correspondence set as $\mathcal{C}_{inlier}$, $\mathcal{C}_{inlier}^{GT}$ and $\mathcal{C}_{inlier}^{correct}$, respectively. Then, the precision, recall and F-measure are respectively defined  as
\begin{equation}\label{eq:LRF30}
\text{Precision}=\frac{\left| \mathcal{C}_{inlier}^{correct}\right|}{\left| {\mathcal{C}_{inlier}} \right|},
\end{equation}
\begin{equation}\label{eq:LRF31}
\text{Recall}=\frac{\left| \mathcal{C}_{inlier}^{correct} \right|}{\left| {\mathcal{C}_{inlier}^{GT}} \right|},
\end{equation}
and
\begin{equation}\label{eq:LRF32}
\text{F-measure}=\frac{2\text{Precision}\times\text{Recall}}{\text{Precision}+\text{Recall}},
\end{equation}
where $\left|{\cdot}\right|$ denotes the cardinality of a set. A correspondence $c=\{\mathbf{x},\mathbf{x}^{'}\}$ belongs to $\mathcal{C}_{inlier}^{GT}$ if
\begin{equation}\label{eq:LRF29}
{{{\| {\mathbf{x}_i^{'}}-{\rho} \left({\mathbf{H}_{gt}}\left[ {\begin{array}{*{20}{c}}{\mathbf {x}_{i}}\\1\end{array}} \right]\right)\|_2}}}\le t_{gt},
\end{equation}
where $\mathbf{H}_{gt}$ is the ground-truth homography matrix and $t_{gt}$ is a threshold set to $10$\textit{pix} (\textit{pix} being the unit of pixel) that controls the upper bound of the accuracy of a true inlier in our experiments.

Similarly, a correct correspondence in $\mathcal{C}_{inlier}$  is defined as
\begin{equation}\label{eq:LRF33}
{{{\| {\mathbf{x}_i^{'}}-{\rho} \left({\mathbf{H}_{gt}}\left[ {\begin{array}{*{20}{c}}{\mathbf {x}_{i}}\\1\end{array}} \right]\right)\|_2}}}\le \tau
\end{equation}
with $\tau$ being the matching tolerance. We vary $\tau$ from 1\textit{pix} to $t_{gt}$ with an interval of 1\textit{pix}, thus generating a curve~\cite{lin2014bilateral,bian2017gms}.
\subsection{Experimental deployment}\label{subsec:exp_deploy}
Our experiments are deployed as follows. In Sect.~\ref{sub:sub1}, the overall performance of the evaluated algorithms in different scenarios, i.e., the four experimental datasets, is tested. In Sect.~\ref{sub:sub2}, the performance with preselected correspondences by NNSR, i.e., commonly employed to improve the inlier ratio of initial matches~\cite{Ma2014Robust,Raguram2008A,yang2016fast,yang2017multi}, is tested on the four datasets. In Sect.~\ref{sub:sub3}, different detector-descriptor combinations are considered to examine the performance variation of correspondence selection algorithms. Notice that different combinations of detector and descriptor are desired in different application contexts~\cite{Mikolajczyk2004Hessian,moreels2007evaluation} and will result in different distributions and inlier ratios. In Sect.~\ref{sub:sub4}, the robustness to different nuisances, i.e., blur, viewpoint change, zoom, rotation, light change, and JPEG compression, is independently examined on the VGG dataset. In Sect.~\ref{sub:sub5}, we address concerns about the efficiency in those algorithms by examining their overall time cost on different datasets paired with the speed comparison under different scales of initial matches. Finally, some representative visual results of the evaluated algorithms are shown in Sect.~\ref{sub:sub6}.
\section{Results}\label{sec:res}
Following the experimental arrangement in Sect.~\ref{subsec:exp_deploy}, this section presents the corresponding results together with necessary discussions and explanations.
\subsection{Performance on the different datasets}\label{sub:sub1}
\begin{figure}[t] 
	\centering
	\subfigure[VGG]
	{%
		\includegraphics[width=0.32\linewidth]{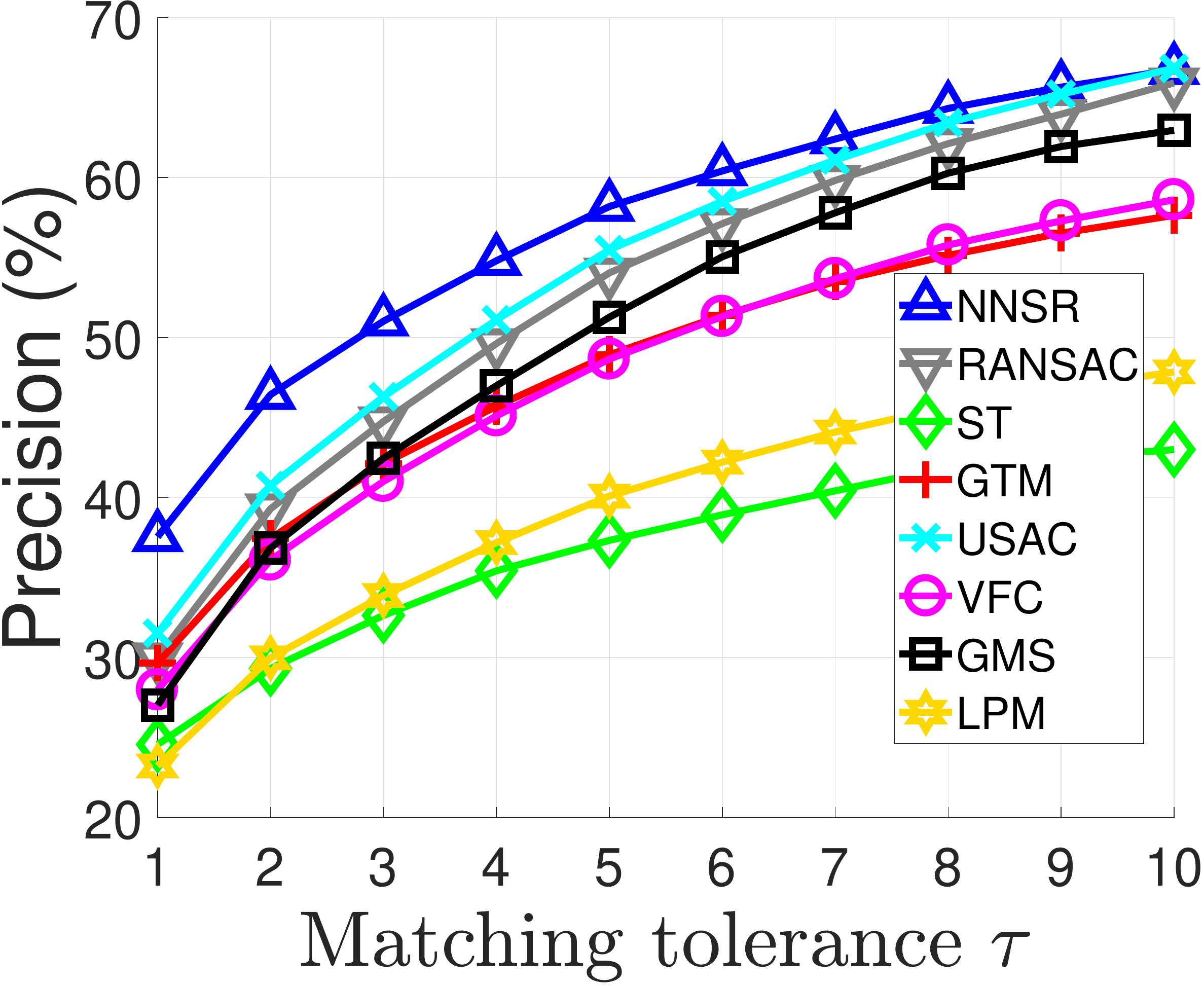} 
		\includegraphics[width=0.32\linewidth]{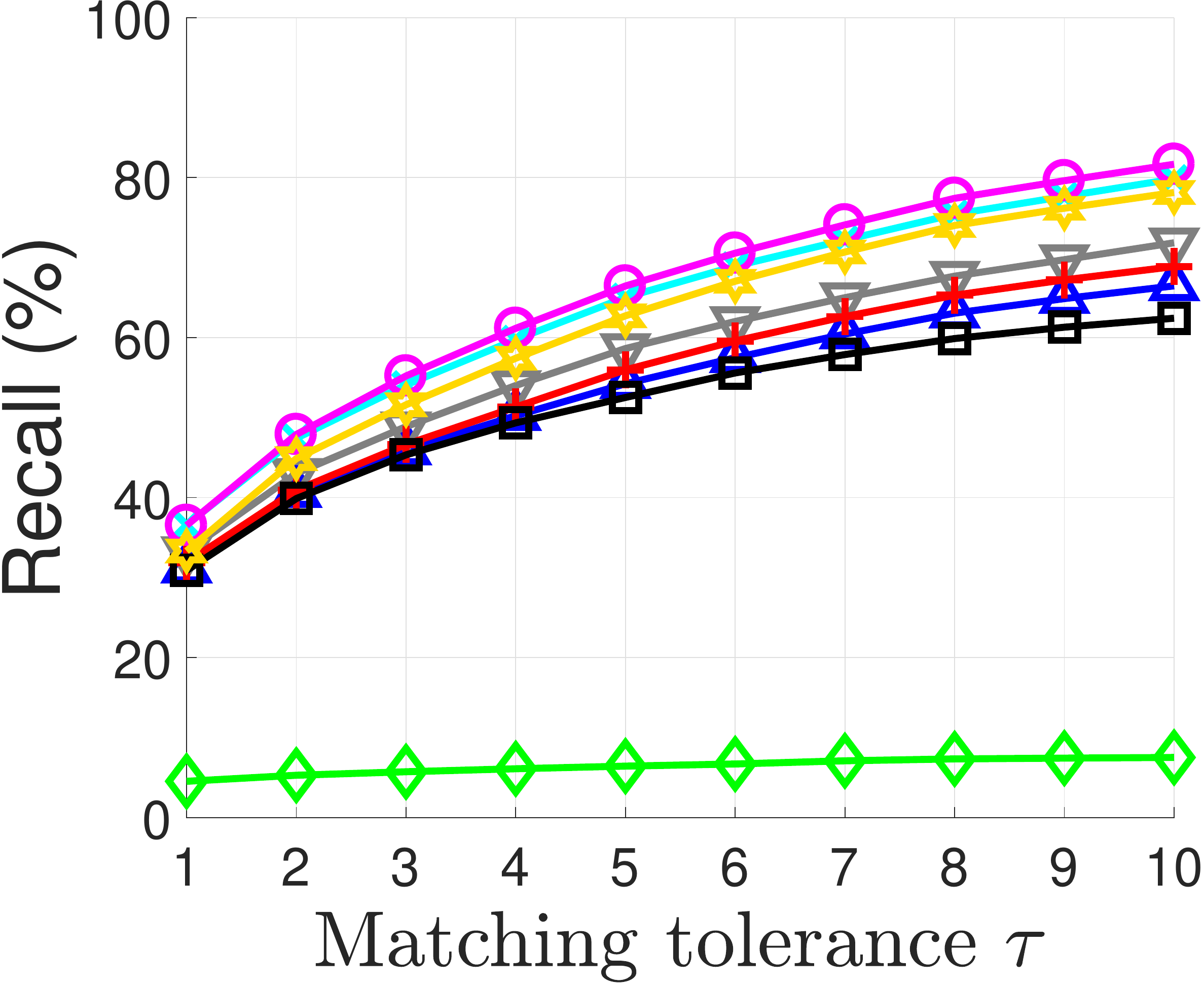}
		\includegraphics[width=0.32\linewidth]{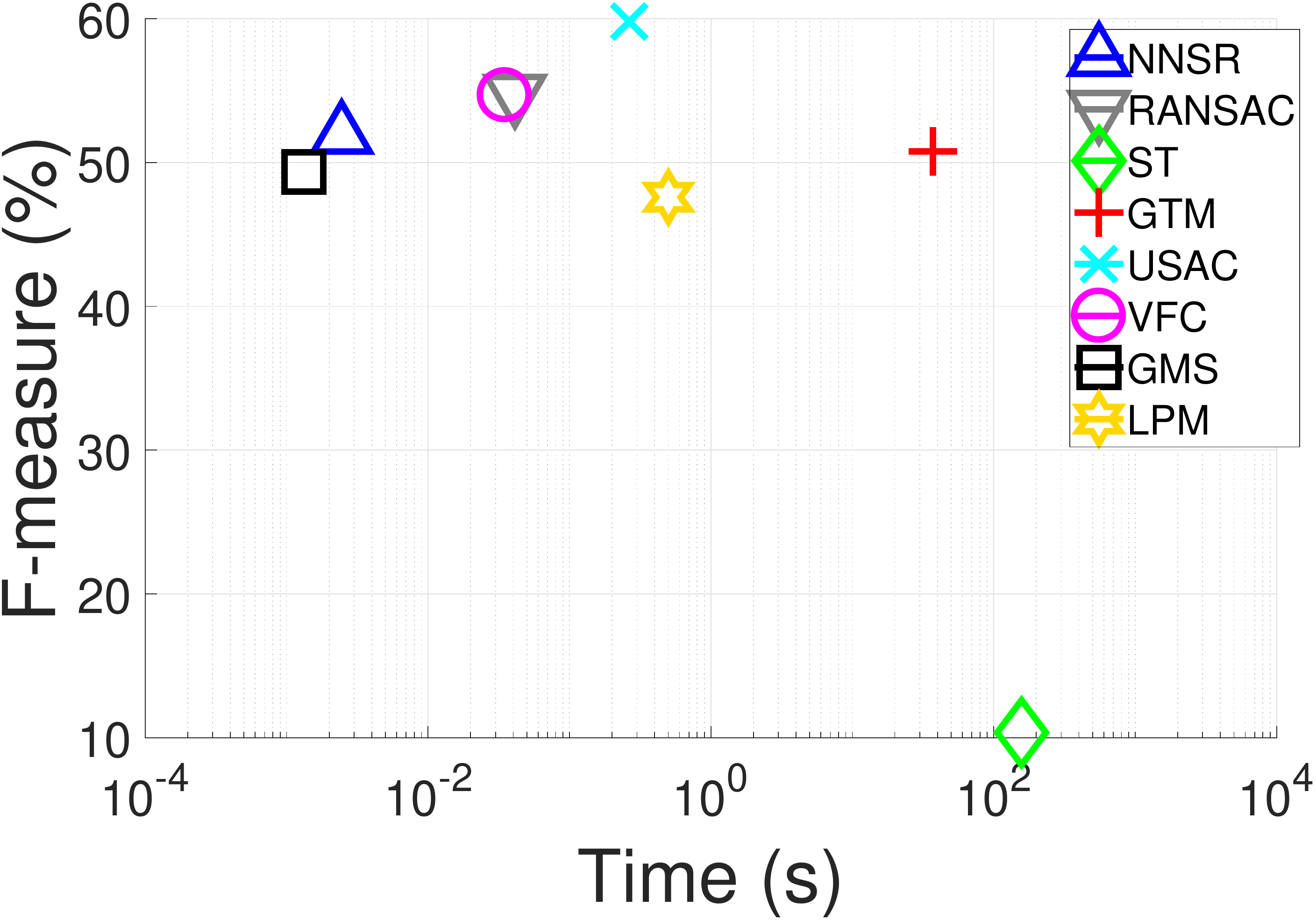} 
	}
    \subfigure[Symbench]
	{%
		\includegraphics[width=0.32\linewidth]{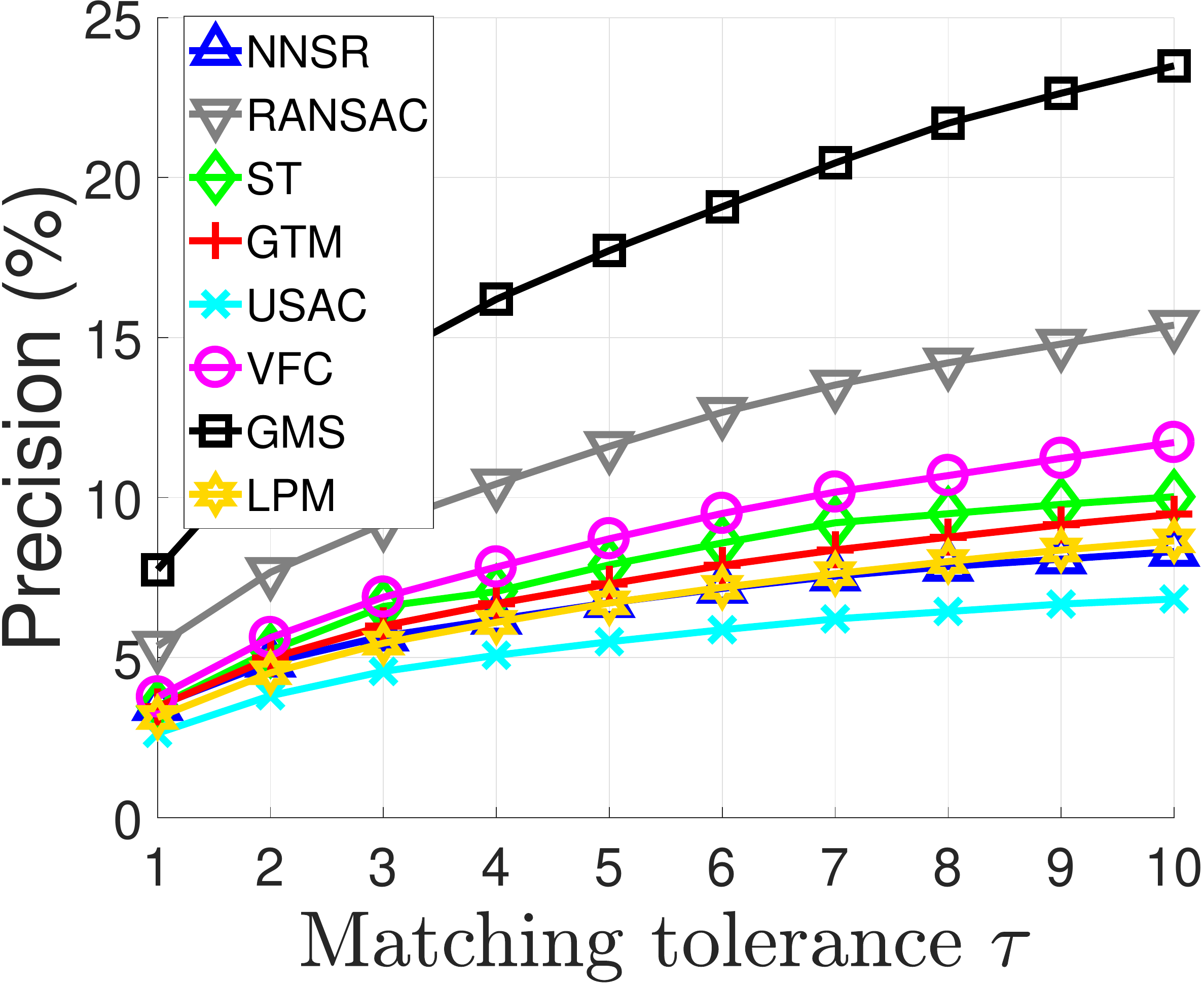} 
		\includegraphics[width=0.32\linewidth]{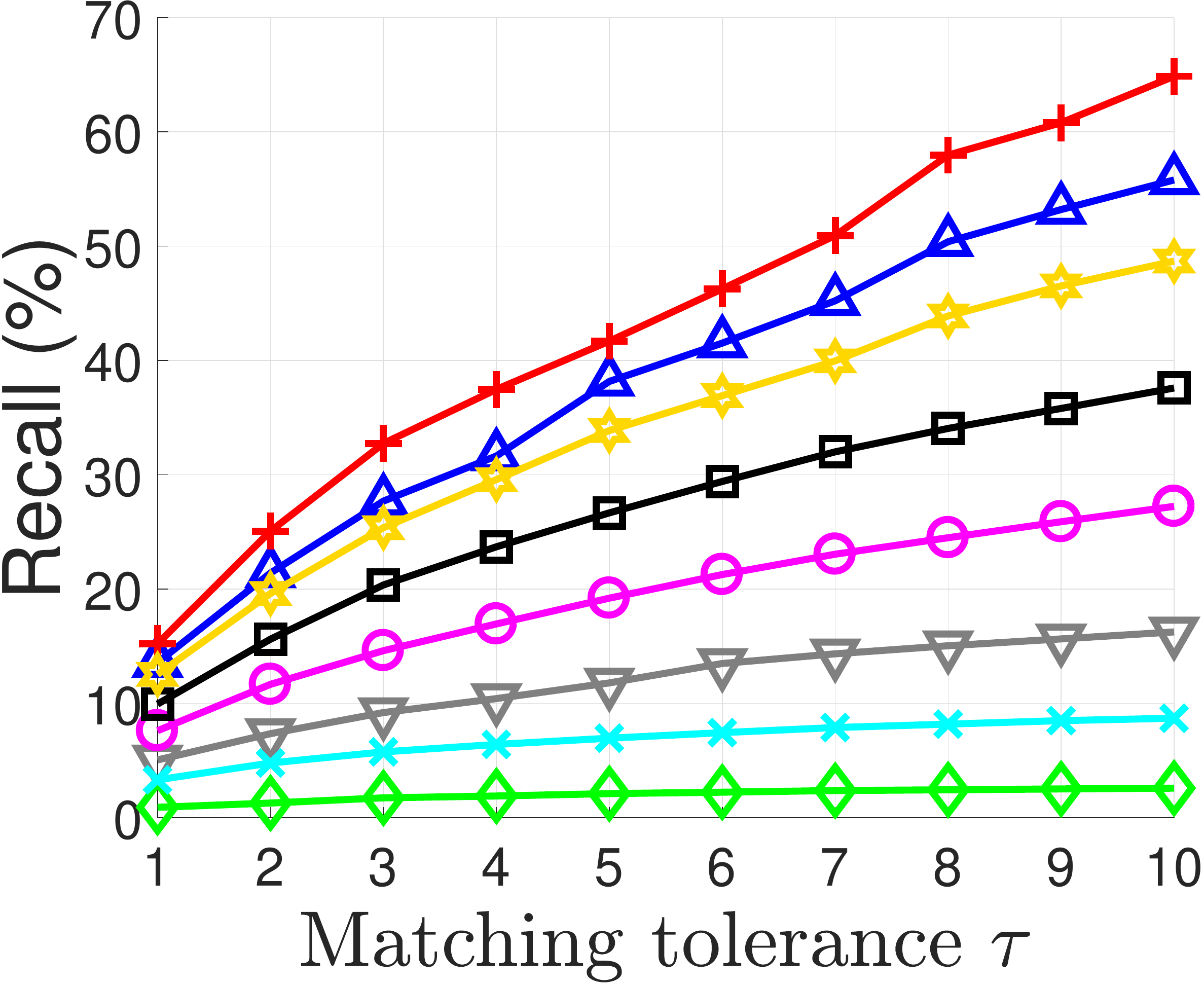} 
		\includegraphics[width=0.32\linewidth]{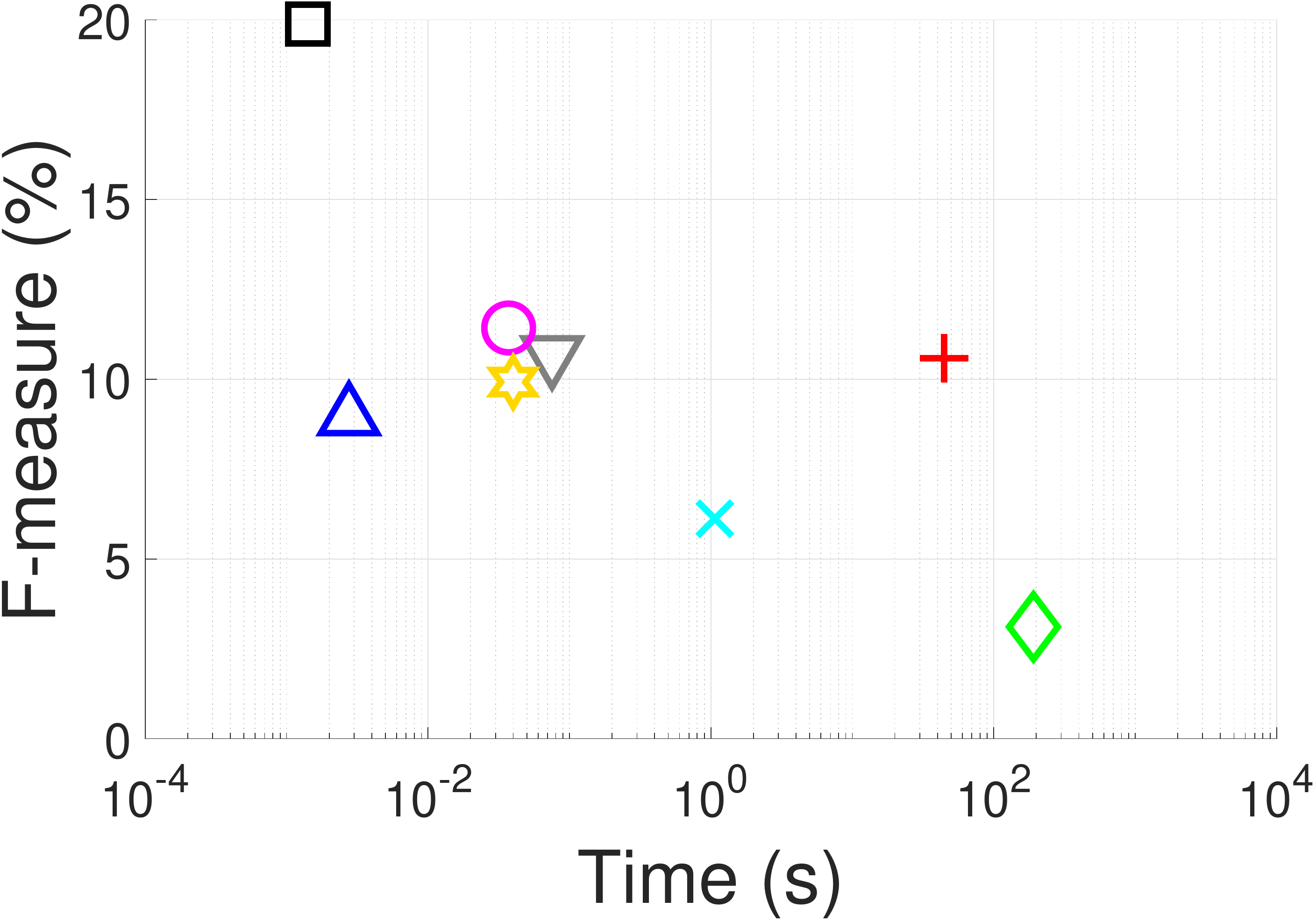}
	}
    \subfigure[Heinly]
	{%
		\includegraphics[width=0.32\linewidth]{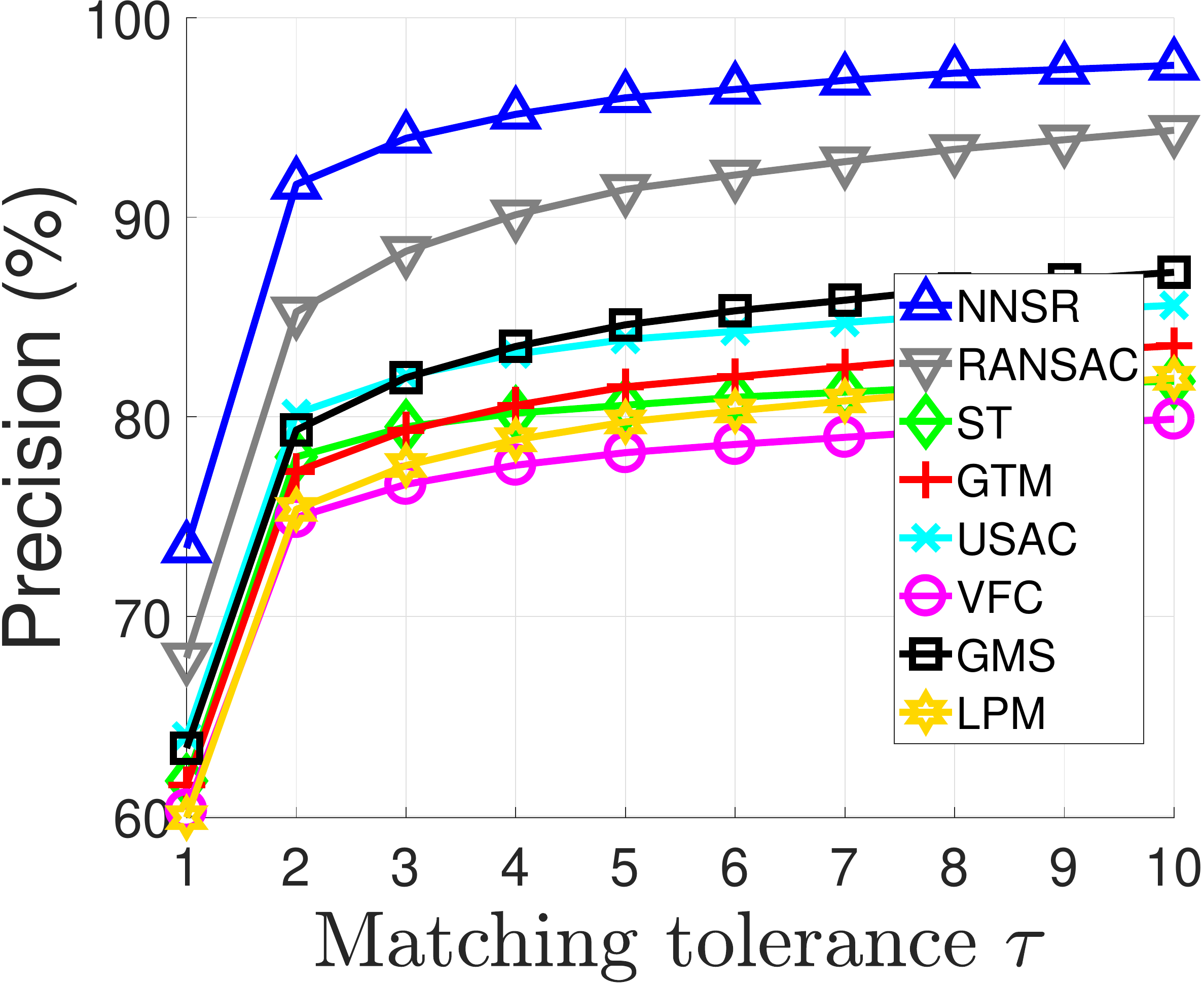} 
		\includegraphics[width=0.32\linewidth]{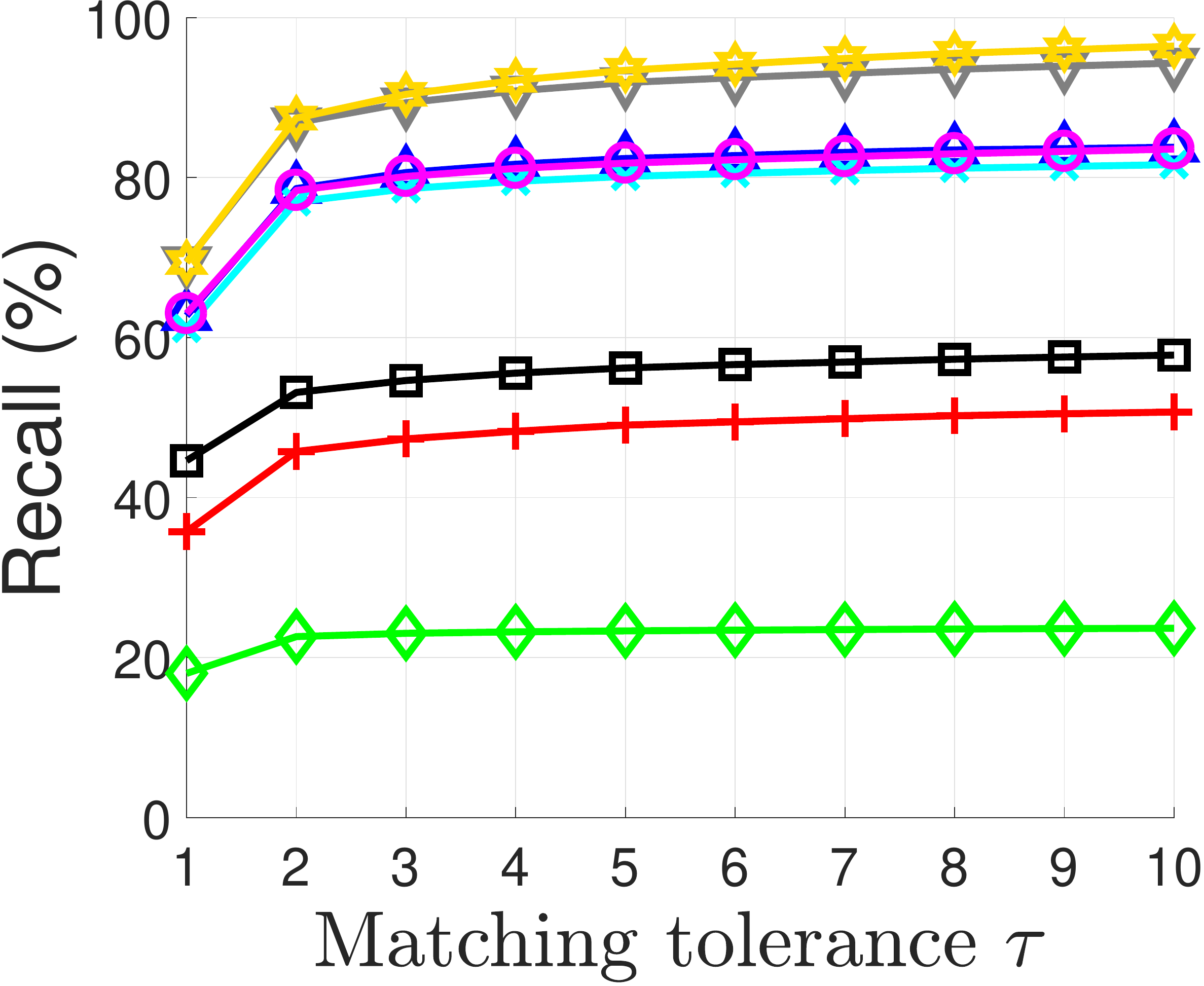} 
		\includegraphics[width=0.32\linewidth]{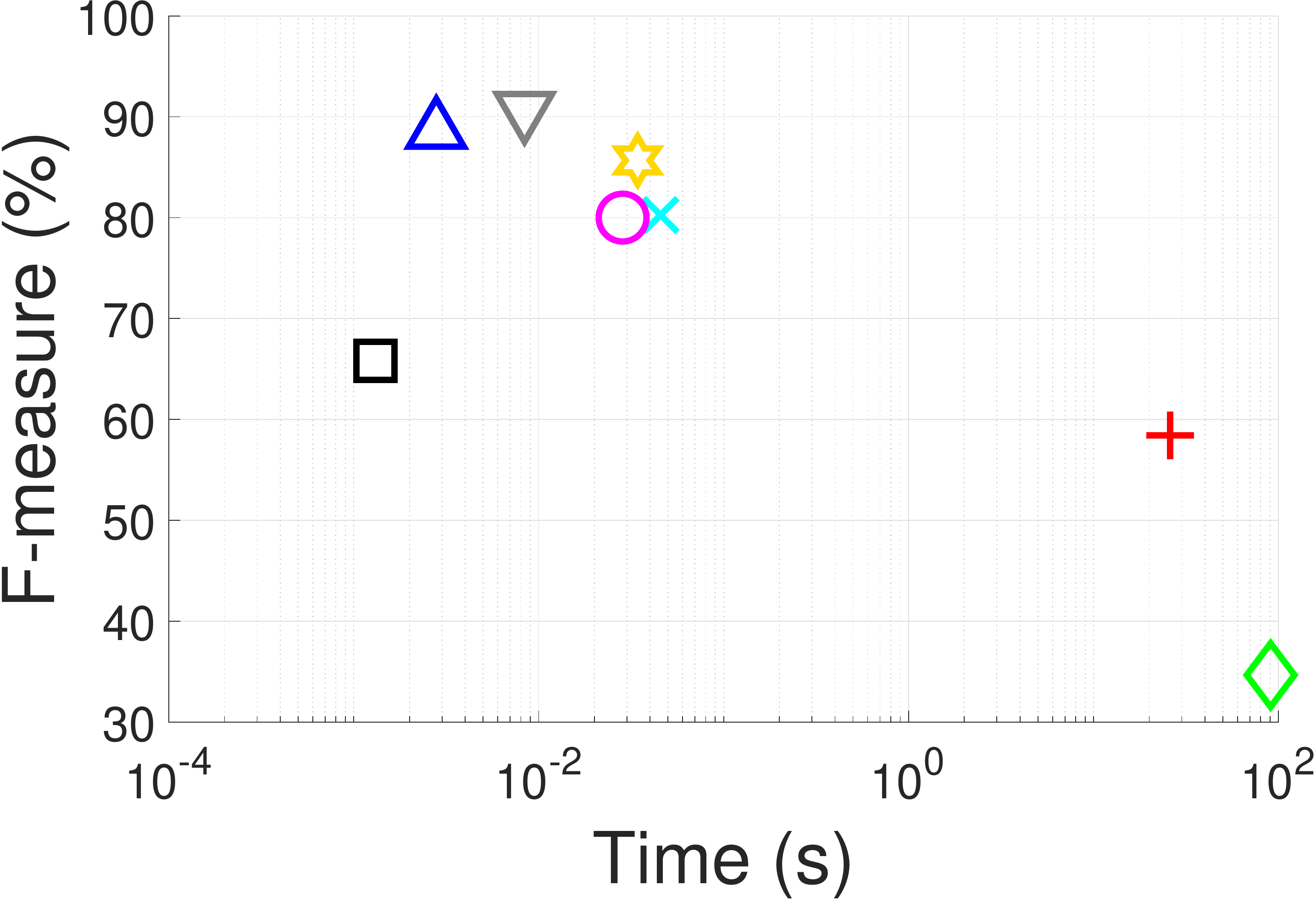}
	}
	\subfigure[AdelaideRMF]
	{
    	\scalebox{0.8}[0.8]
    	{%
    		\begin{tabular}	{lccccccc}	
    			\hline
    		  	& RANSAC & ST & USAC & VFC & GMS & LPM 	\\
    			\hline
    		Precision (\%) & 93.84 & 77.32 & 96.28 & 97.59 & \textbf{98.54} & 81.21 \\		
    		Recall (\%) & 68.77 & 21.31 & 79.13 & 87.36 & 79.22 & \textbf{99.62} & 
    		\\	
    		F-measure (\%) & 77.03 & 32.18 & 84.89 & \textbf{91.63} & 87.25 & 88.91
    		\\
    		\hline	
    		\end{tabular}\label{tab:ade}
    		
    	}
	}
	\caption{Performance of the evaluated algorithms on four datasets, i.e.,  (a) VGG, (b) Symbench, (c) Heinly, and (d) AdelaideRMF, in terms of precision, recall and F-measure under different matching tolerance $\tau$. The maximum values of precision, recall and F-measure are shown in bold face on the AdelaideRMF dataset.}
	\label{fig:per1}
\end{figure}
\begin{figure*}[t] 
	\centering
	\subfigure[VGG]
	{%
		\includegraphics[width=0.9\linewidth]{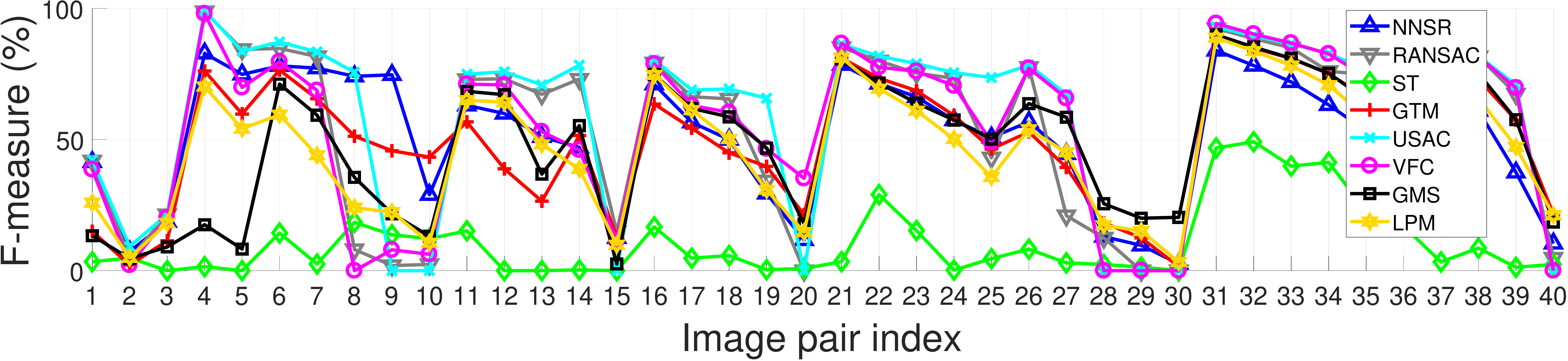}  	
	}
	\subfigure[Symbench]
	{%
		\includegraphics[width=0.9\linewidth]{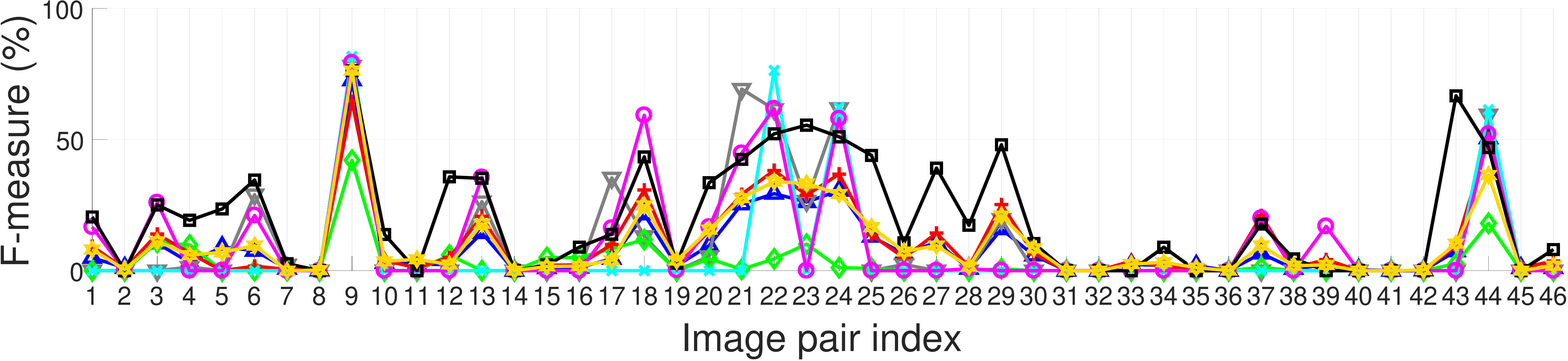}
	}
	\subfigure[Heinly]
	{%
		\includegraphics[width=0.9\linewidth]{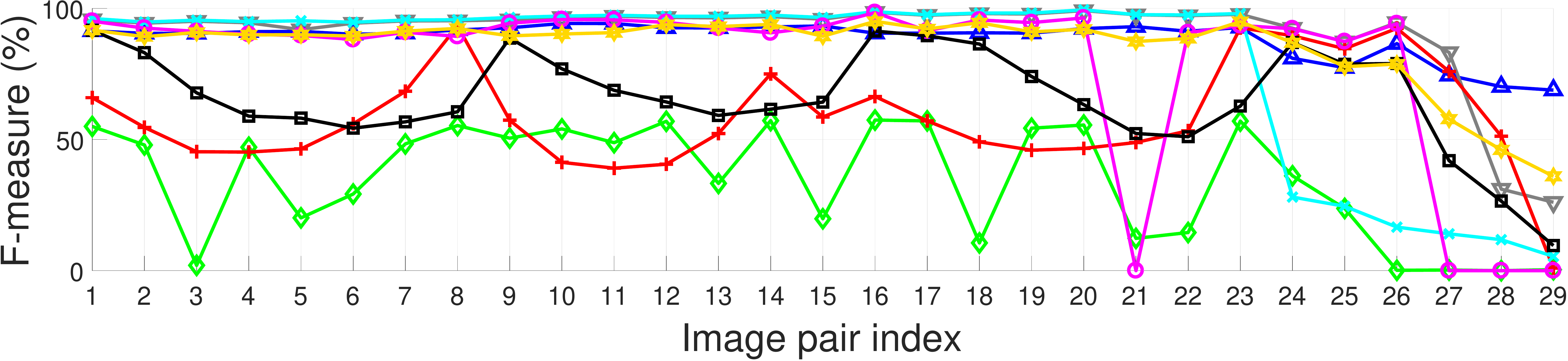}  
	}
	\subfigure[AdelaideRMF]
	{%
		\includegraphics[width=0.9\linewidth]{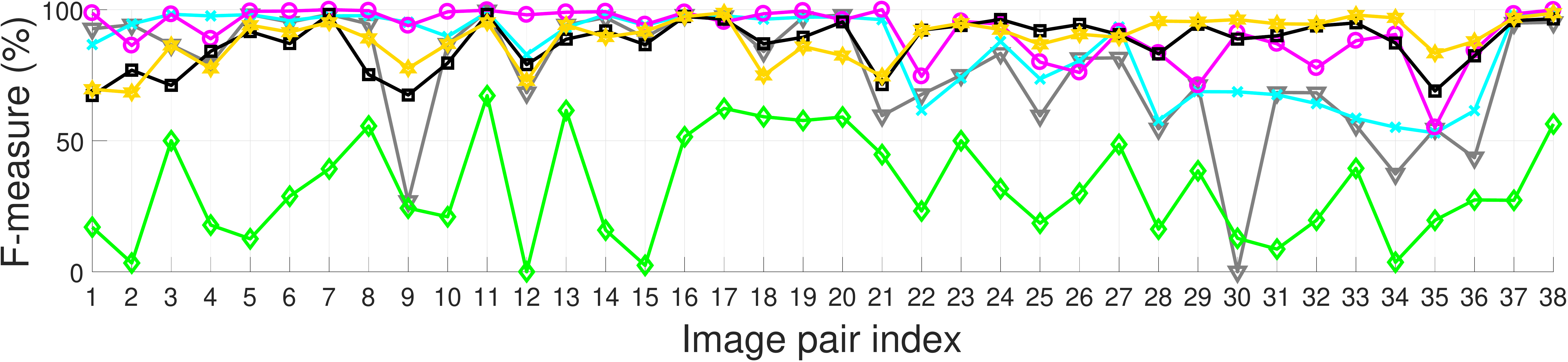}  
	}
	\caption{F-measure performance tested on each image pair on the (a) VGG, (b) Symbench, (c) Heinly, and (d) AdelaideRMF datasets for all evaluated algorithms. Here, the matching tolerance $\tau$ is set as $5$.}
	\label{fig:per2}
\end{figure*}
In the following, we show the precision, recall and F-measure performance of our evaluated algorithms on different datasets, i.e., under different scenarios. In particular, the overall precision, recall and F-measure curves are shown in Fig.~\ref{fig:per1} for aggregately view and the F-measure scores for each image pair on the four datasets are shown in Fig.~\ref{fig:per2} to give a more detailed view. We mainly discuss the performance based on Fig.~\ref{fig:per1}.
\subsubsection{Performance on the VGG dataset}
Fig.~\ref{fig:per1}(a) shows outcomes on the VGG dataset. It is interesting to see that NNSR achieves the best precision performance, being marginally better than USAC, RANSAC and GMS. This result is due to the fact that the feature distinctiveness cue is rather selective with rich-textured images, e.g., images in the VGG dataset. On the down side, feature distinctiveness is sometimes ambiguous and not a robust constraint as we can see that the recall of NNSR is just mediocre. It indicates that many correct correspondences have been filtered by NNSR. For ST and LPM, they are generally inferior to the others on this dataset in terms of the F-measure. That is because ST may fail to locate the main cluster in the spectral domain if the ourlier ratio is large, resulting in quite poor recall performance. LPM achieves much better recall performance than ST, while its precision performance is surpassed by most compared ones. It arises from the loose constraint employed in LPM. Overall, USAC is the best method on this dataset. Explanation behind is that USAC is a parametric method and the parametric model of each image pair existed in this dataset can be properly fitted.
\subsubsection{Performance on the Symbench dataset}
Fig.~\ref{fig:per1}(b) presents results on the Symbench dataset. All methods suffer a clear drop in performance on this dataset when compared with that on the VGG dataset, which is attributed to light change and various rendering styles. More specifically,  we observed that the average inlier ratio of initial correspondences on this dataset is lower than $10\%$. As previously explained, the feature distinctiveness constraint strongly relies on the discriminative power of the local feature descriptor. However, the rendering style variation makes it fairly challenging to maintain descriptiveness in this case. As a result, NNSR delivers very poor precision performance. Another significant difference compared to that on the VGG dataset is USAC's performance. One can see that USAC returns the most and the second most inferior precision and recall performance, respectively. That is because USAC may find empty inlier sets in some cases when its average estimated scores decreases owing to the multiple constraints in this algorithm~\cite{Raguram2013USAC}. In general, GMS and VFC are the two most well-behaved methods after referring their F-measure rankings. A common trait of these two algorithms is that both of them are independent from the descriptor similarity.
\subsubsection{Performance on the Heinly dataset}
Fig.~\ref{fig:per1}(c) presents results on the Heinly dataset. Image pairs on this dataset only contain pure zoom or rotation, and we can observe that all methods obtain relatively decent performance on this dataset. In terms of precision, NNSR and RANSAC neatly outperform the others. Regarding recall, LPM and RANSAC are the two best ones. Note that the reason for the high recall of LPM is that  most inliers are selected with the loose constraint designed by this algorithm. For NNSR and RANSAC, the former one is attributed to the high distinctiveness of SIFT (we will see its performance variation with less distinctive descriptors in Sect.~\ref{sub:sub3}), whereas the latter one is owing to the powerful homography fitting ability of RANSAC. GMS, due to its sensitivity to large degrees of rotation~\cite{bian2017gms}, shows worse results compared to its performance on the VGG and Symbench datasets.
\subsubsection{Performance on the AdelaideRMF dataset}
Fig.~\ref{fig:per1}(d) presents results on the AdelaideRMF dataset. Two explanations should be given on this dataset. First, as only manual labeled ground-truth correspondences are available, we present the exact scores rather than curves with respect to matching tolerance for each method. Second, the keypoints on this dataset are not located by image detectors. Rather, they were labeled manually. Thus, GTM requiring local affine information and NNSR based on auto-detected keypoints are not assessed on this dataset. Since each scene in this dataset contains multiple planes, the fundamental matrix based on the epipolar geometry constraint is employed to approximate the parametric model for RANSAC and USAC.
By observing the scores in Fig.~\ref{fig:per1}(d), one can see that GMS, LPM and VFC achieve the best precision, recall and F-measure performance, respectively. All the three methods are non-parametric. This is reasonable since the AdelaideRMF contains multi-structures, and the parametric assumption for methods like RANSAC and USAC will fail in this case.  
\subsubsection{Overall performance}
By weighing up the results presented in Fig.~\ref{fig:per1} and Fig.~\ref{fig:per2}, we can draw the following conclusions. First, the performance of all correspondences selection algorithms is affected by the initial inlier ratio. For instance, the performance of all algorithms deteriorates dramatically on the Symbench dataset with less than 10\% inliers. Second, NNSR simply relying on feature's distinctiveness produces pleasurable results if images are well-textured and clean. Third, parametric approaches, i.e., RANSAC and USAC, prefer the context that the transformation between two images can be well fitted by a parametric model. While non-parametric algorithms perform better in  situations without large degrees of rigid/non-rigid transformation. Overall, VFC and RANSAC are the two best algorithms under across-dataset experiments.
\subsection{Performance on selected matches}\label{sub:sub2}
\begin{figure}[htbp] 
	\centering
	\subfigure[Brute force matching]
	{
	    \includegraphics[width=0.47\linewidth]{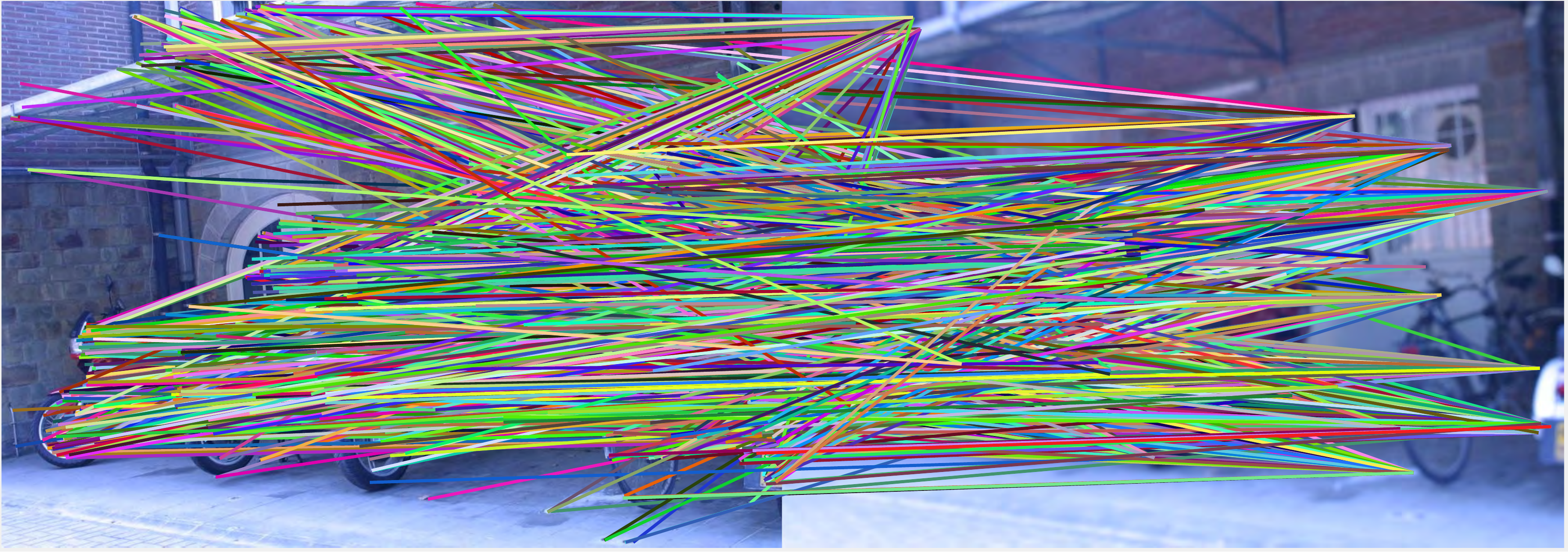}
	    \includegraphics[width=0.47\linewidth]{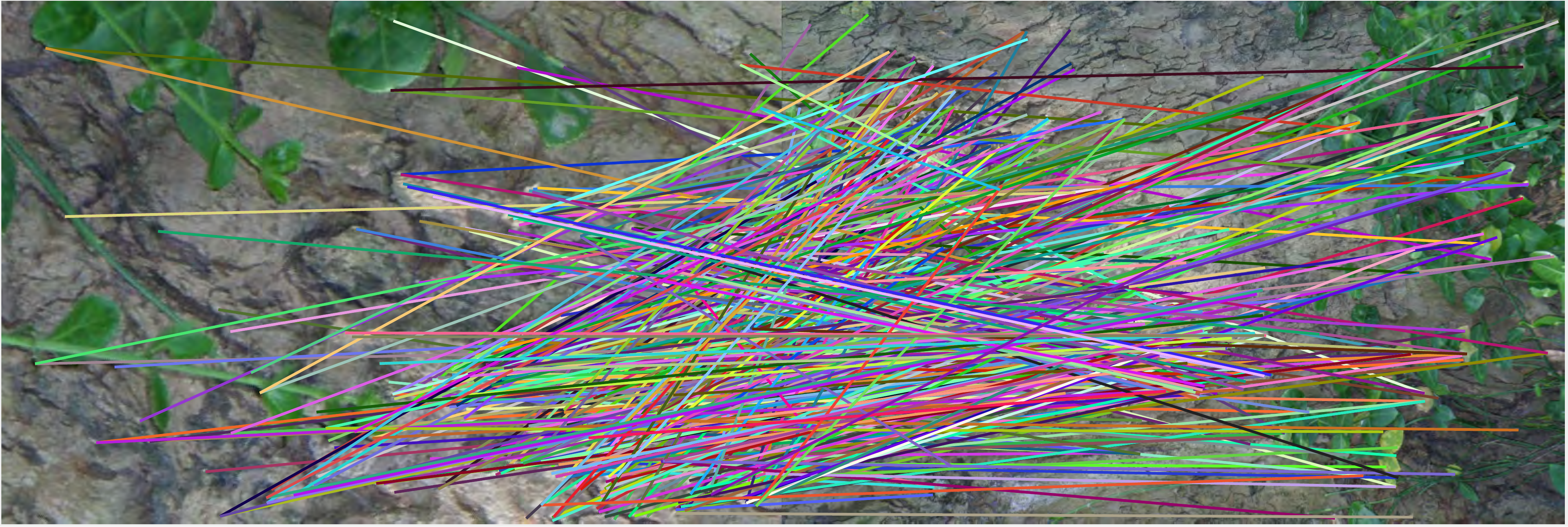} 
	}
	\subfigure[NNSR-selected matching]
	{
	    \includegraphics[width=0.47\linewidth]{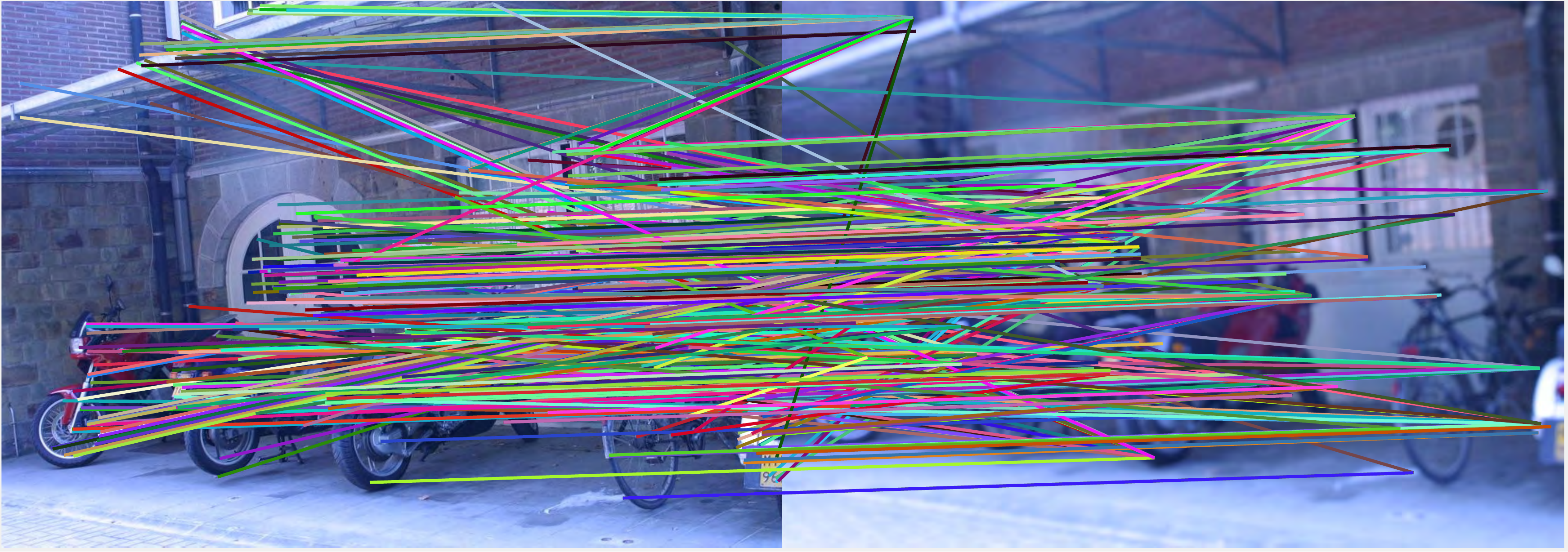}
	    \includegraphics[width=0.47\linewidth]{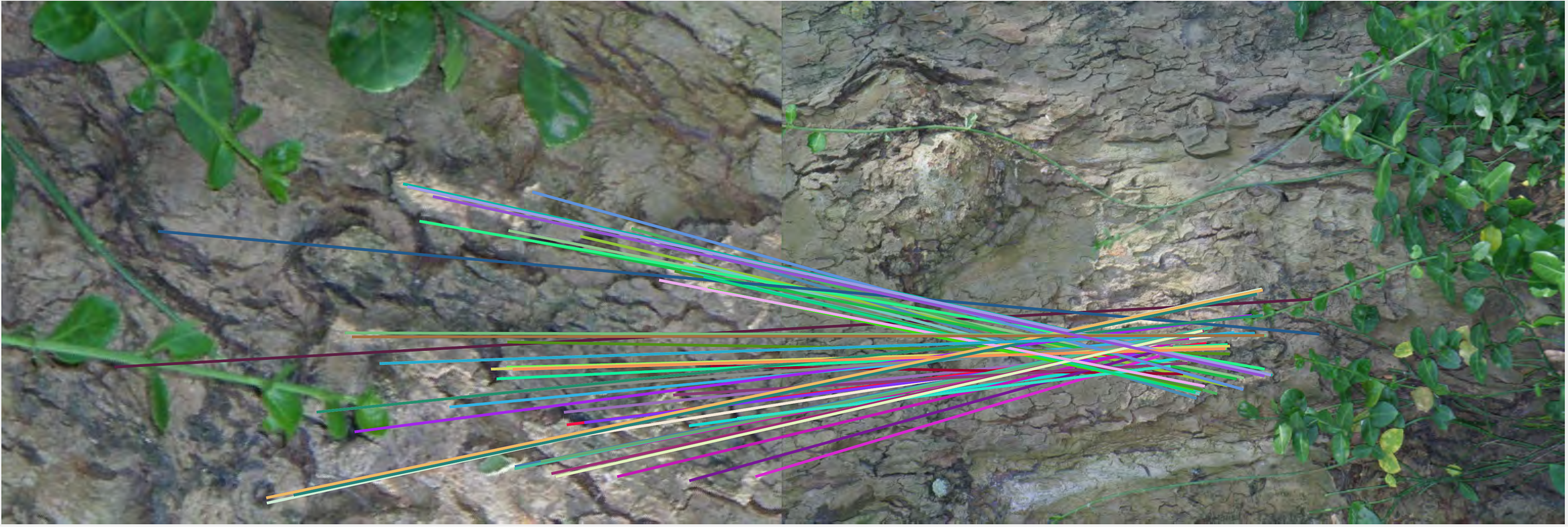} 
	}
	\caption{Examples  of correspondence sets obtained via brute-force matching (a) and NNSR selection (b). Sample image pairs are taken from the VGG dataset.}
	\label{fig:per3}
\end{figure}

\begin{figure}[htbp] 
	\centering
	\subfigure[]
	{%
	\includegraphics[width=0.325\linewidth]{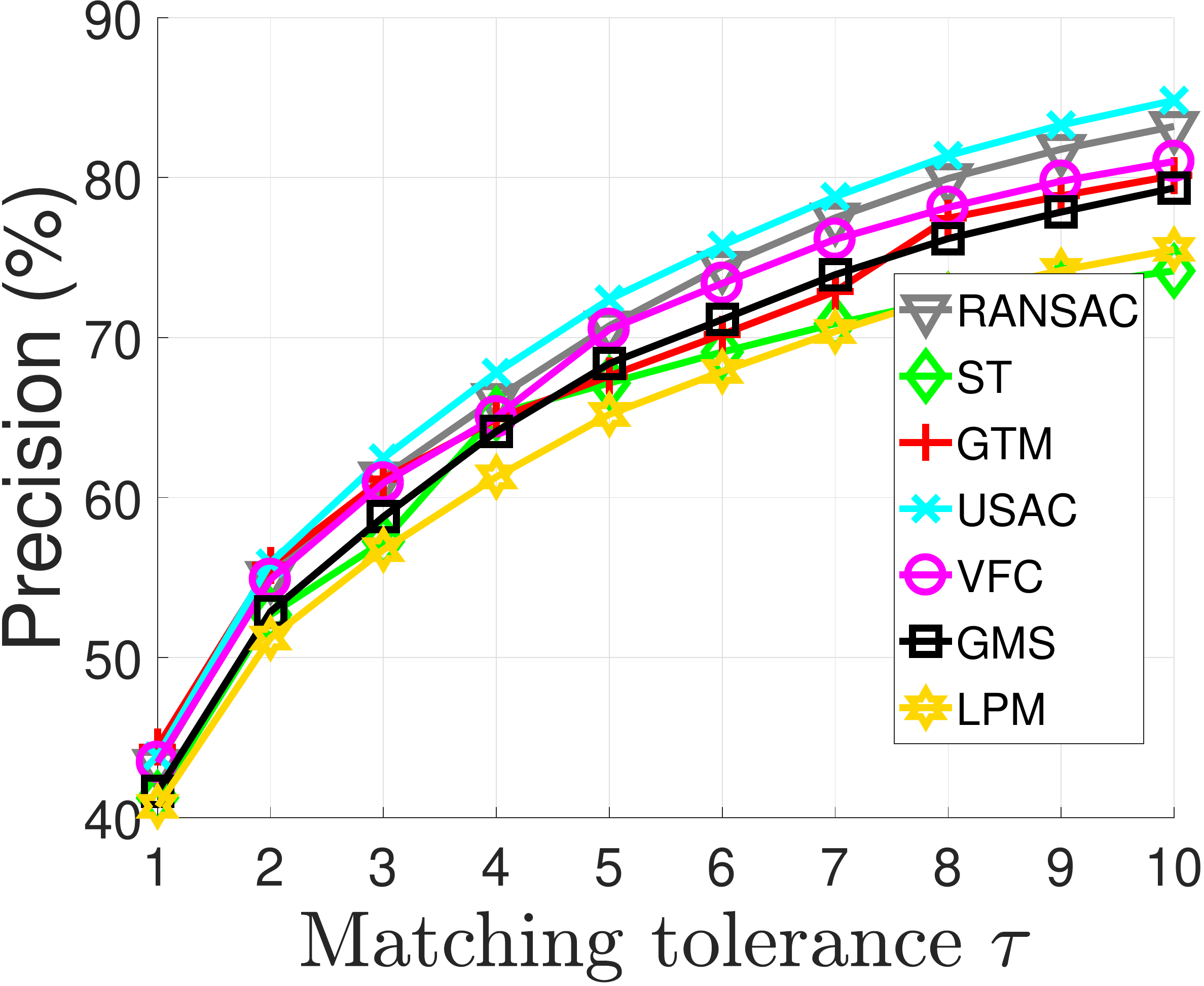}
	\includegraphics[width=0.325\linewidth]{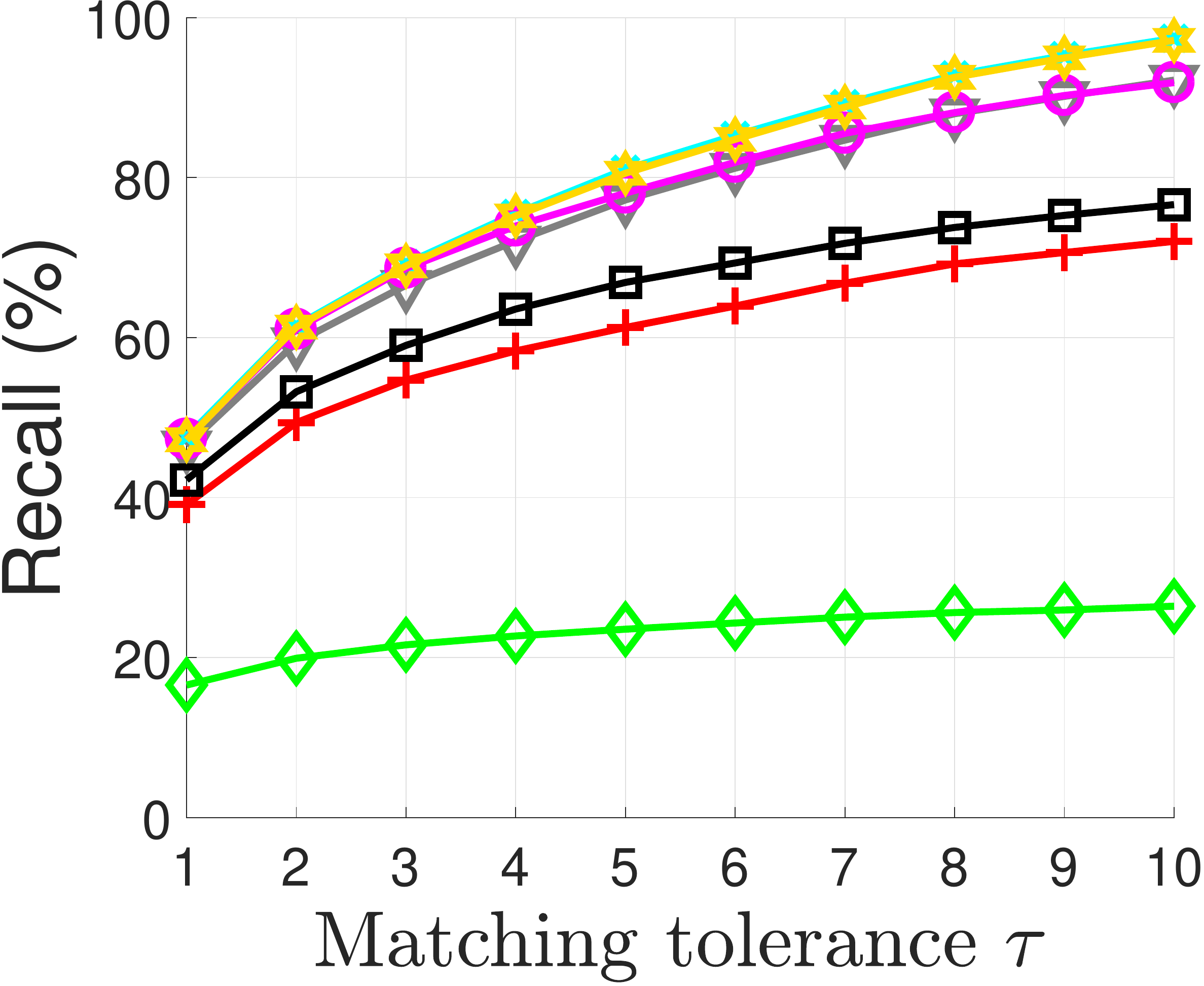} 
	\includegraphics[width=0.325\linewidth]{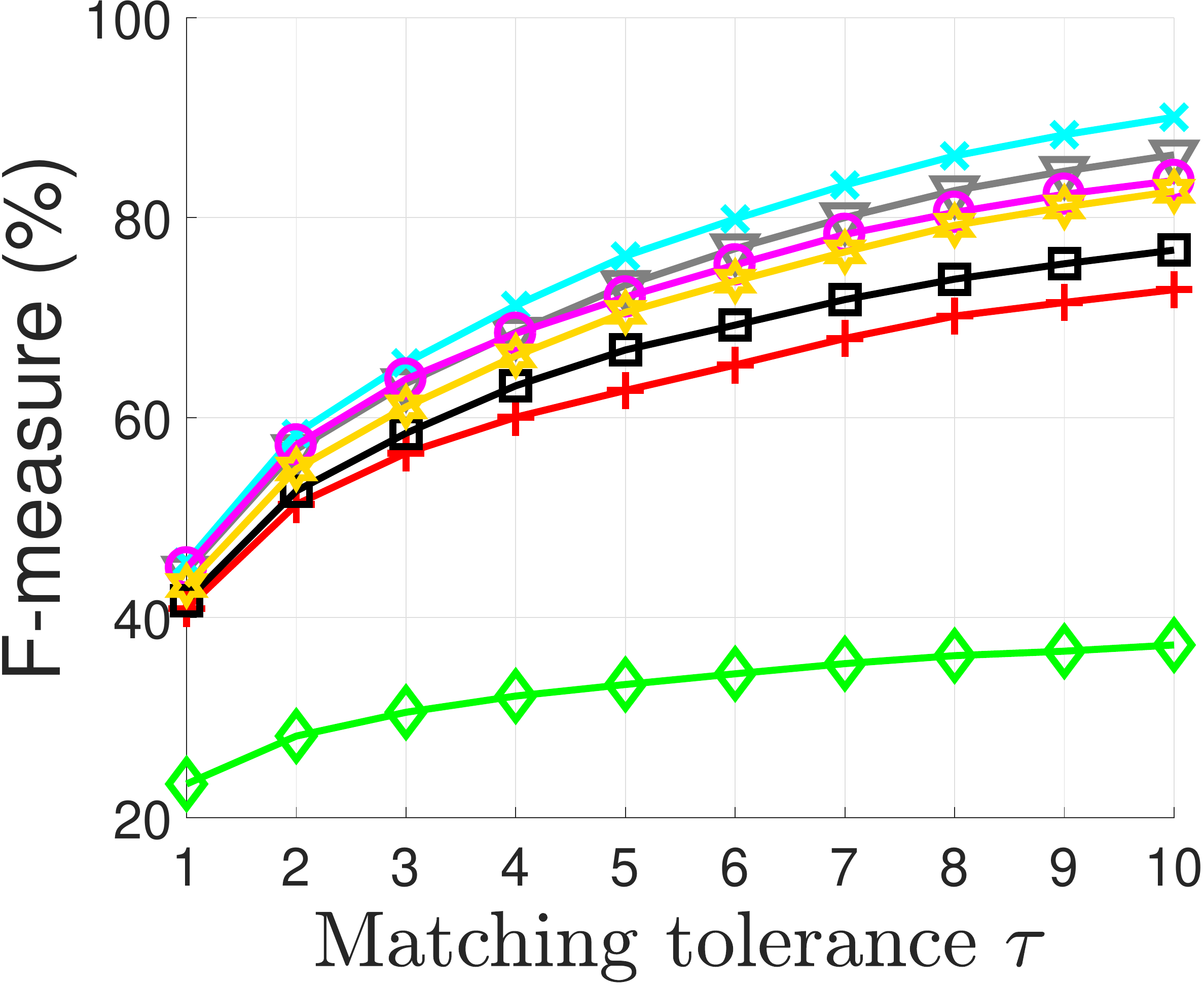}
	}
	\subfigure[]
	{
    	\scalebox{0.8}[0.8]
    	{%
    	\begin{tabular}	{lcccccccc}	
    		\hline
    		& RANSAC & ST & GTM & USAC & VFC & GMS & LPM 	\\
    		\hline
    		Symbench & 17.26 & 5.06 & 18.46 & 14.84 & 22.37 & \textbf{26.59} & 18.04	\\		
    		Heinly & 97.12 & 51.20 & 58.41 & 87.45 & 96.16 & 70.45 & \textbf{97.27} \\
    		\hline	
    	\end{tabular}\label{tab:select}	
    	}
	}
\caption{Performance of evaluated algorithms (except NNSR) on selected matches on the (a) VGG, (b)  Symbench and Heinly datasets. For aggregated view, precision, recall and F-measure curves are shown for the VGG dataset, and F-measure (\%) performance under  $\tau=5$ is shown for the Symbench and Heinly datasets. The AdelaideRMF dataset is not tested as NNSR fails to work on this dataset.}

\label{fig:per4}
\end{figure}
Many existing works~\cite{Ma2014Robust,Raguram2008A,yang2016fast,yang2017multi} first prune false correspondences via NNSR and then use parametric or non-parametric methods to for further selection. This experiment then checks this scenario. Remarkably, since NNSR fails to work on the AdelaideRMF dataset, this dataset is not considered in this test. Fig.~\ref{fig:per3} shows the difference between correspondences before and after applying NNSR, and results using NNSR-selected correspondences for selection are shown in Fig.~\ref{fig:per4}.

On the VGG dataset shown in Fig.~\ref{fig:per4}(a), one can see that the performance of all methods has been improved using NNSR-selected matches compared to brute-force matches in Fig.~\ref{fig:per1}(a). Particularly, USAC manages to be the best method regarding precision, recall and F-measure. Also, gaps between most curves excluding that of ST are relatively small. On the Symbench and Heinly datasets, GMS and LPM respectively achieve the best overall performance, where LPM even produces an extremely high F-measure score, i.e., 97.27\%, on the Heinly dataset. We can infer that LPM adapts well to initial correspondence sets with high inlier ratio.
\subsection{Performance under different detectors and descriptors}\label{sub:sub3}
\begin{figure}[htbp]
    \centering
    \subfigure[]
    {%
    	\begin{minipage}[t]{0.47\linewidth}
    		\includegraphics[width=1.0\linewidth]{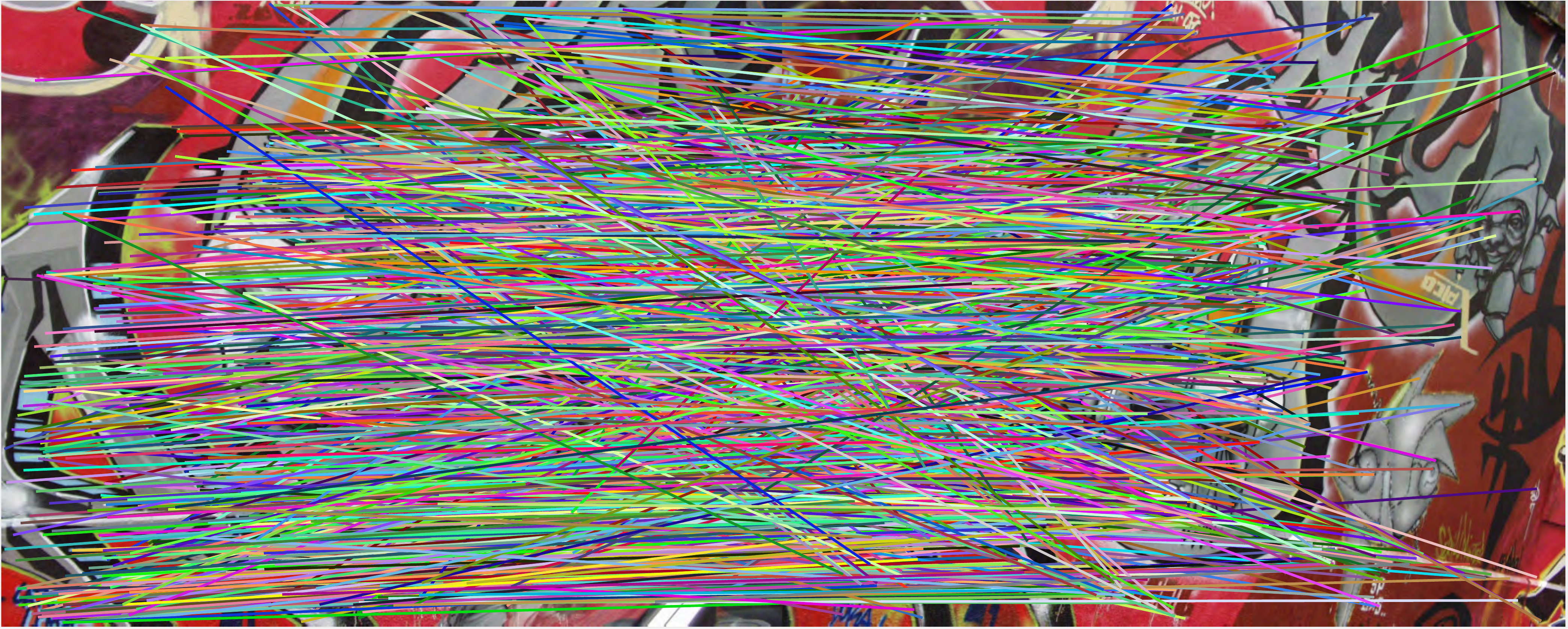} 
    	\end{minipage}
	}
	\subfigure[]
    {%
    	\begin{minipage}[t]{0.47\linewidth}
    		\includegraphics[width=1.0\linewidth]{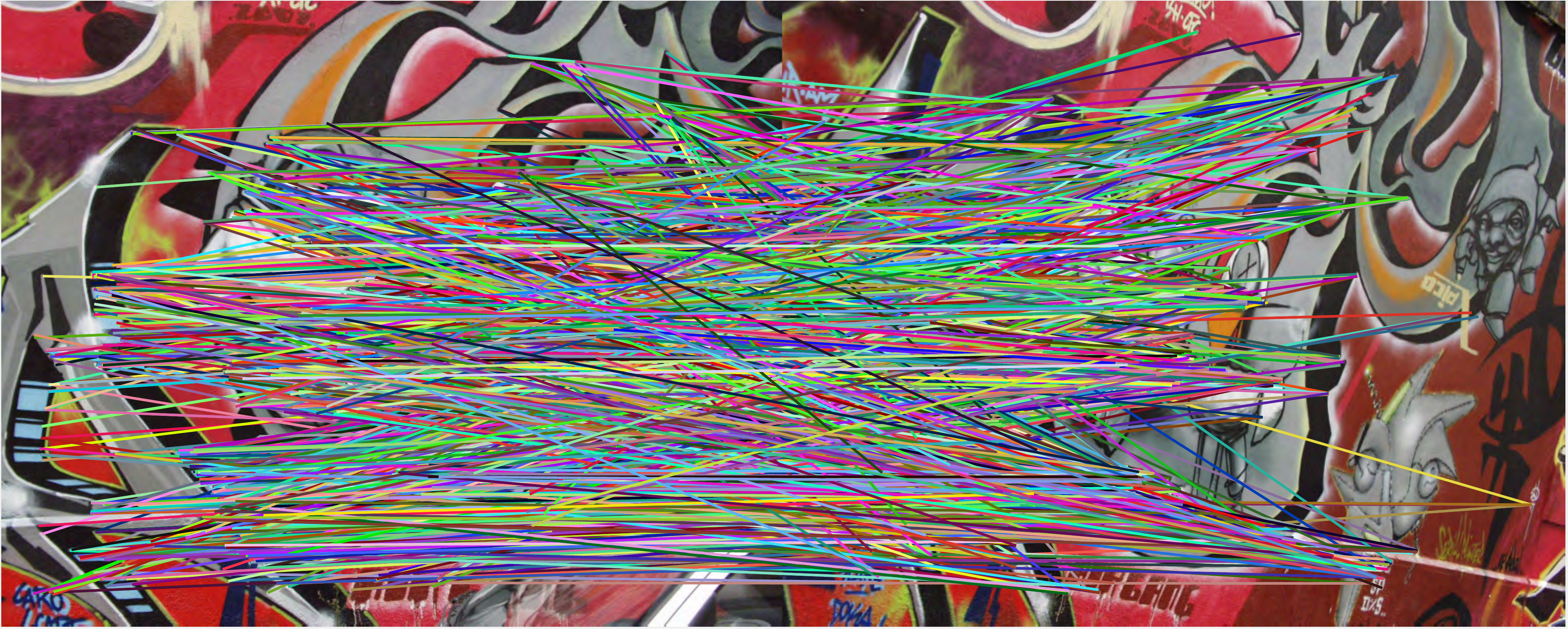} 
    	\end{minipage}
	}
	\subfigure[]
    {%
    	\begin{minipage}[t]{0.47\linewidth}
    		\includegraphics[width=1.0\linewidth]{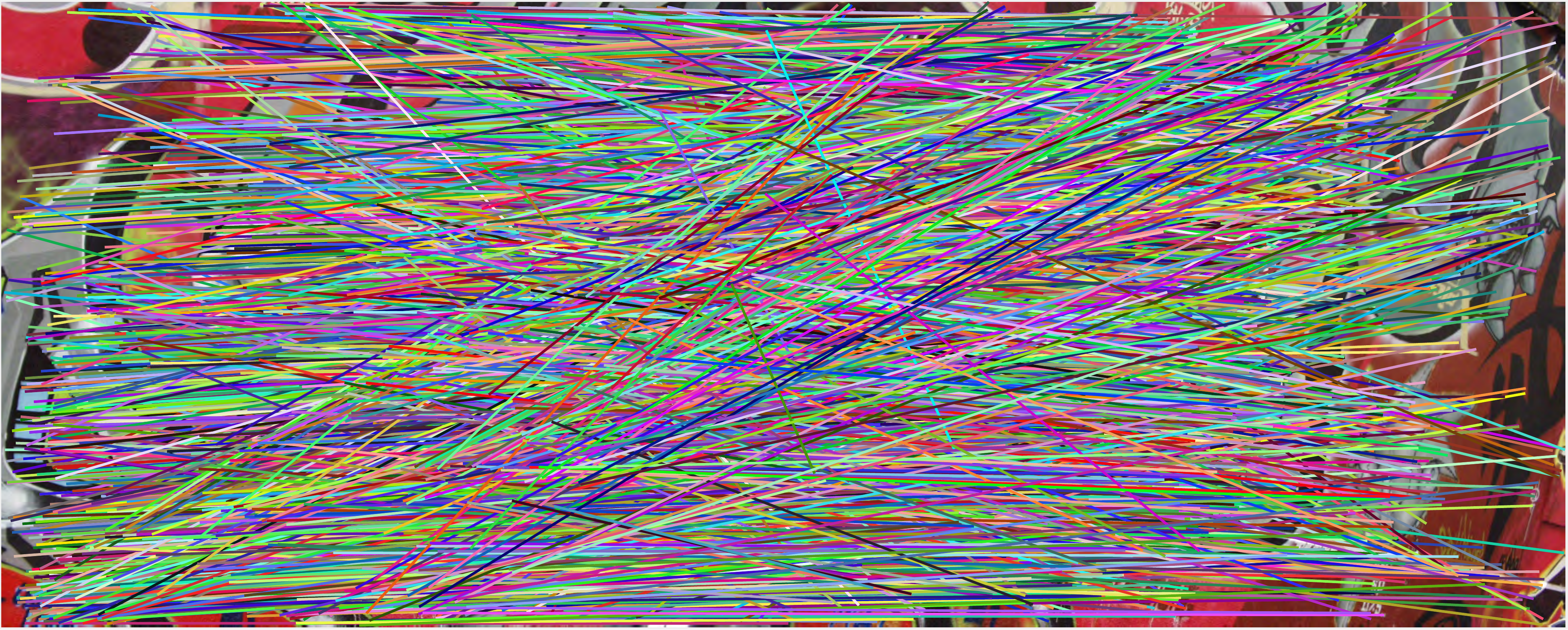} 
    	\end{minipage}
	}
	\subfigure[]
    {%
    	\begin{minipage}[t]{0.47\linewidth}
    		\includegraphics[width=1.0\linewidth]{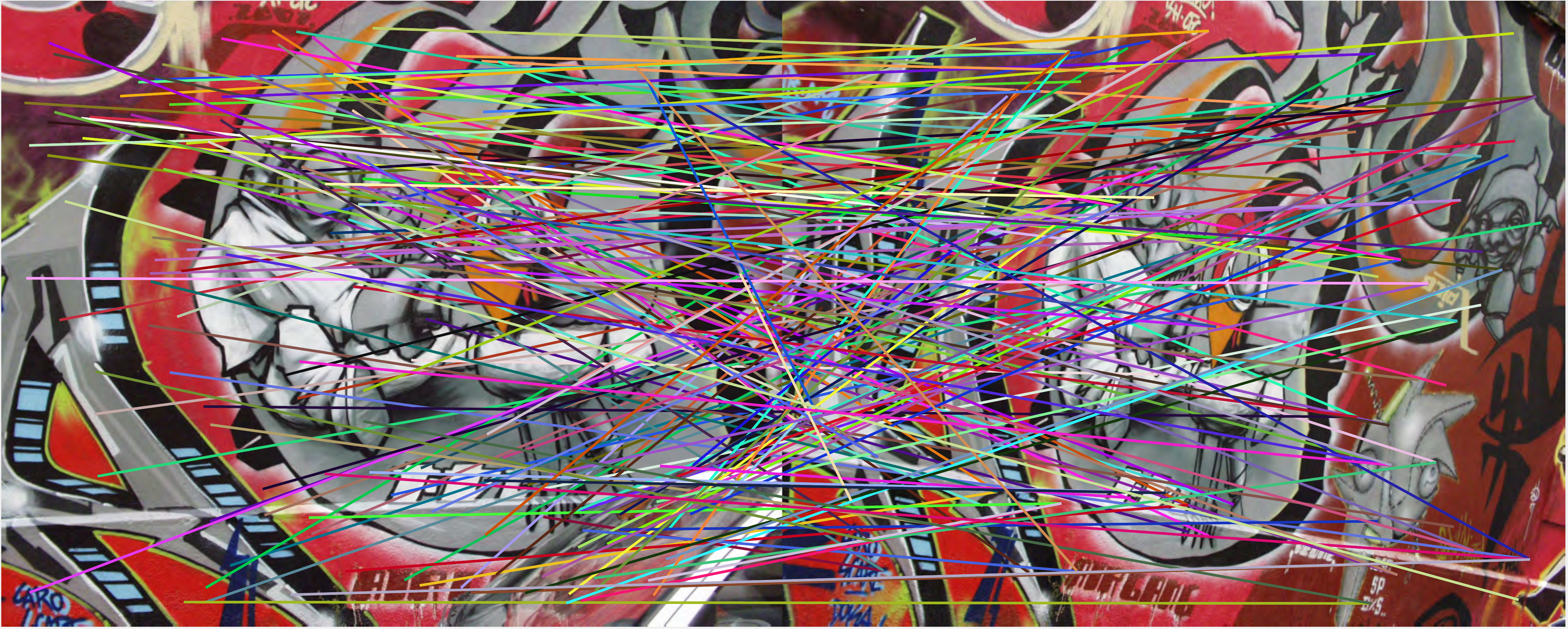} 
    	\end{minipage}
	}
	\caption{Initial correspondences using (a) SIFT + SIFT, (b) ORB + ORB, (c) ASIFT + ASIFT and (d) BLOB + FREAK on an exemplar image pair taken from the VGG dataset.}
	\label{fig:per5}
\end{figure}

\begin{figure*}[t] 
\centering
\subfigure[SIFT + SIFT]
{%
    \begin{minipage}[t]{0.47\linewidth}
        \includegraphics[width=0.3\linewidth]{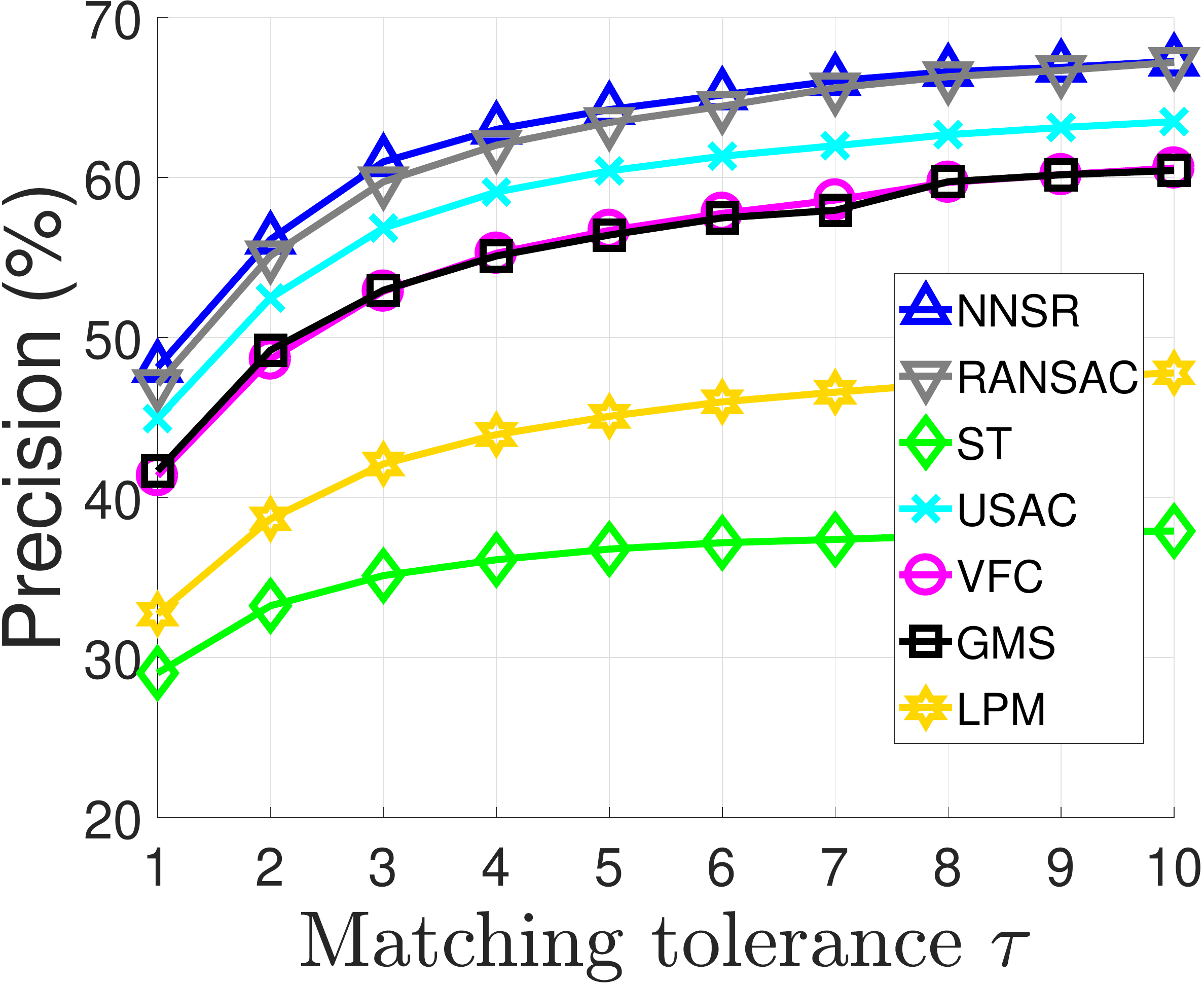} 
	    \includegraphics[width=0.3\linewidth]{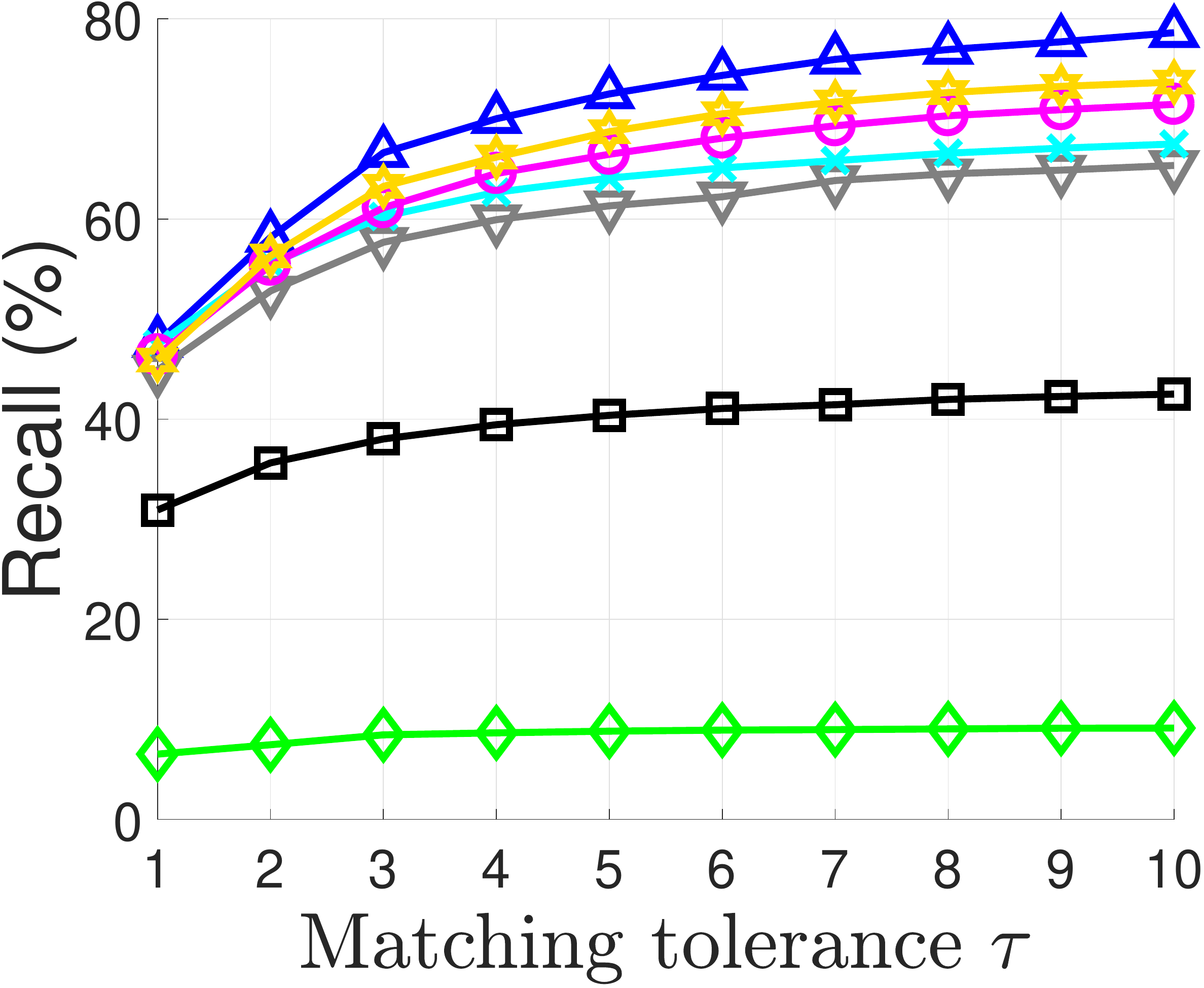} 
	    \includegraphics[width=0.3\linewidth]{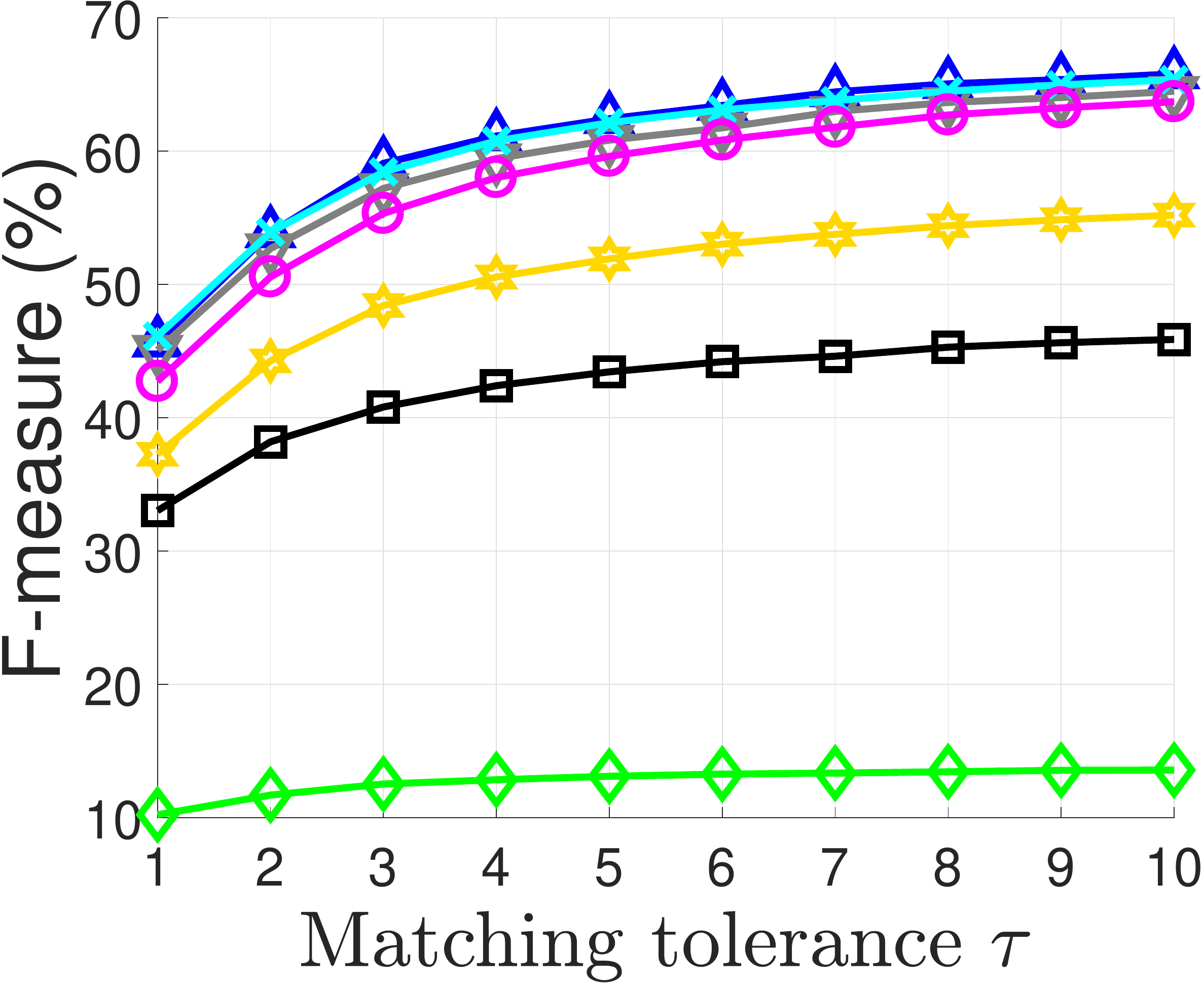}
    \end{minipage}
}
\subfigure[ ORB + ORB]
{%
    \begin{minipage}[t]{0.47\linewidth}
        \includegraphics[width=0.3\linewidth]{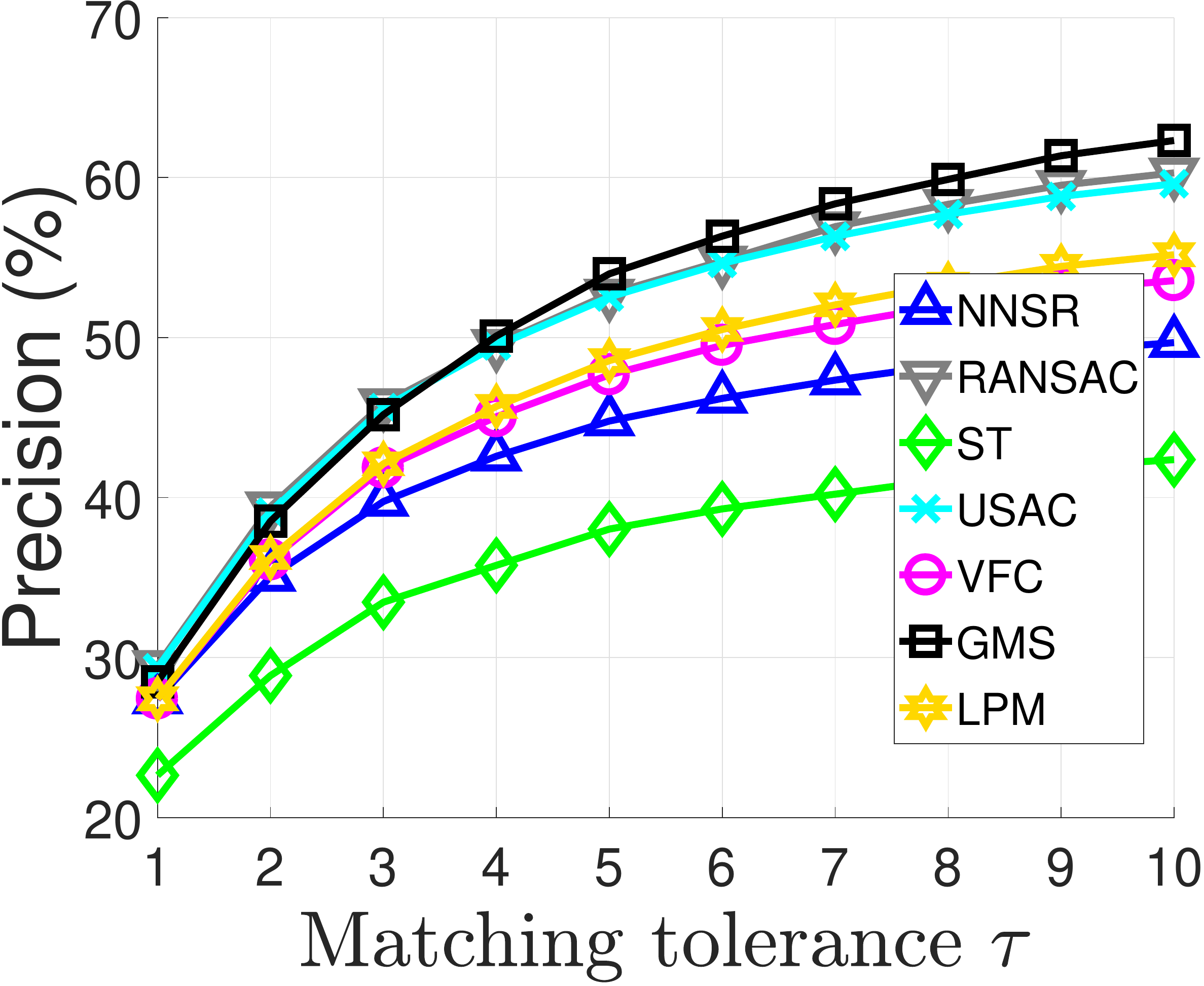} 
	    \includegraphics[width=0.3\linewidth]{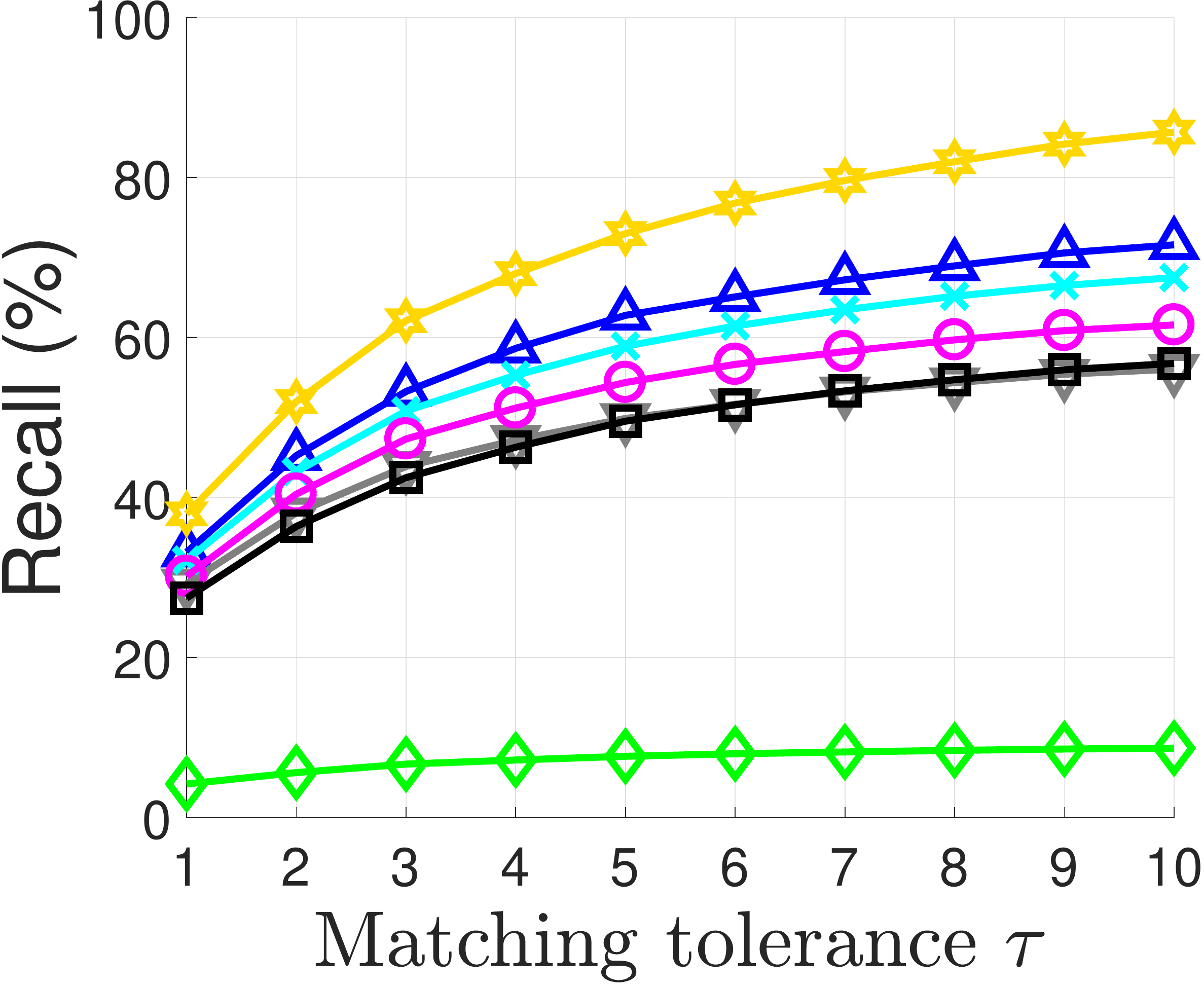} 
	    \includegraphics[width=0.3\linewidth]{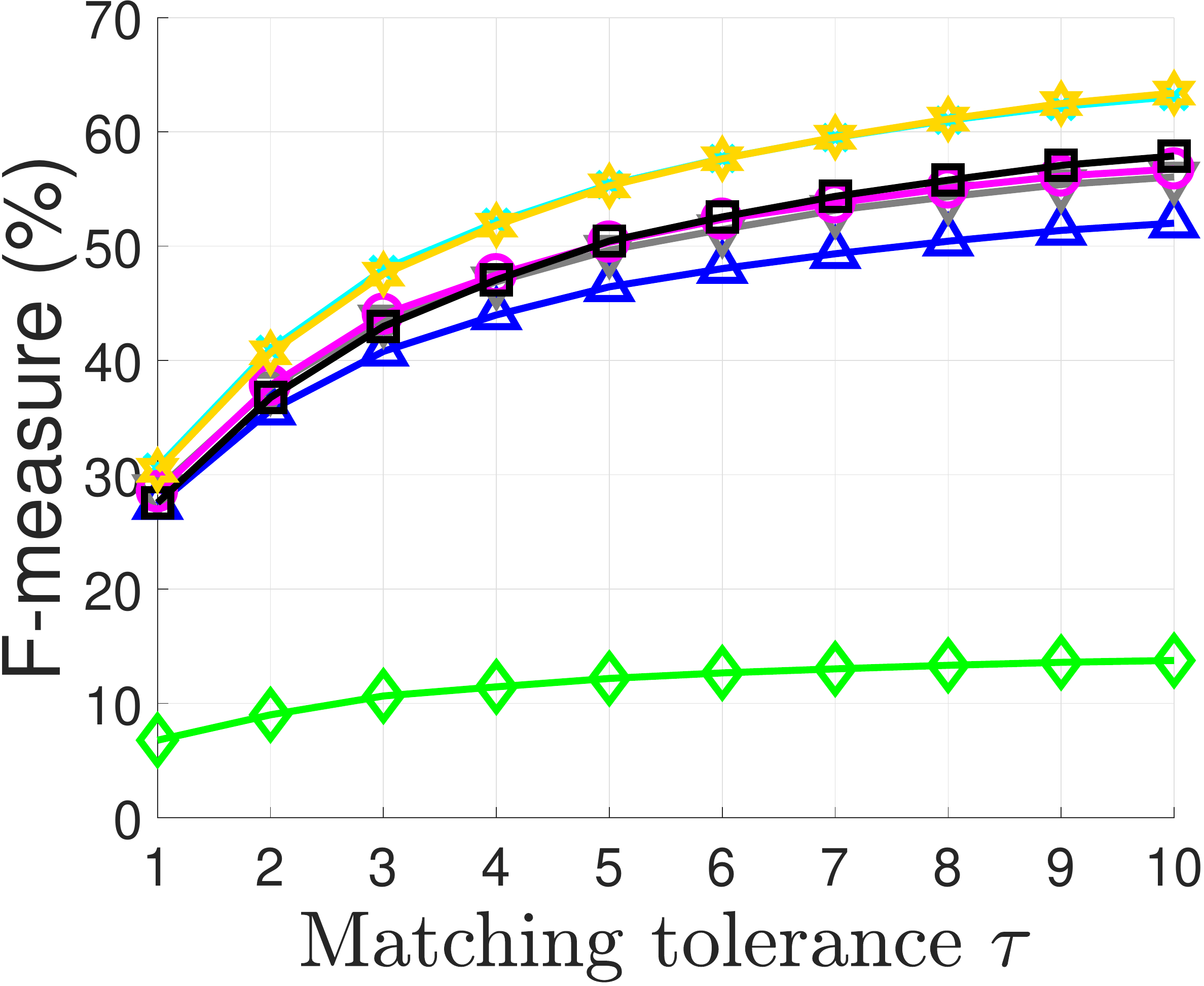}
    \end{minipage}
}
\subfigure[ ASIFT + ASIFT]
{%
    \begin{minipage}[t]{0.47\linewidth}
    	\includegraphics[width=0.3\linewidth]{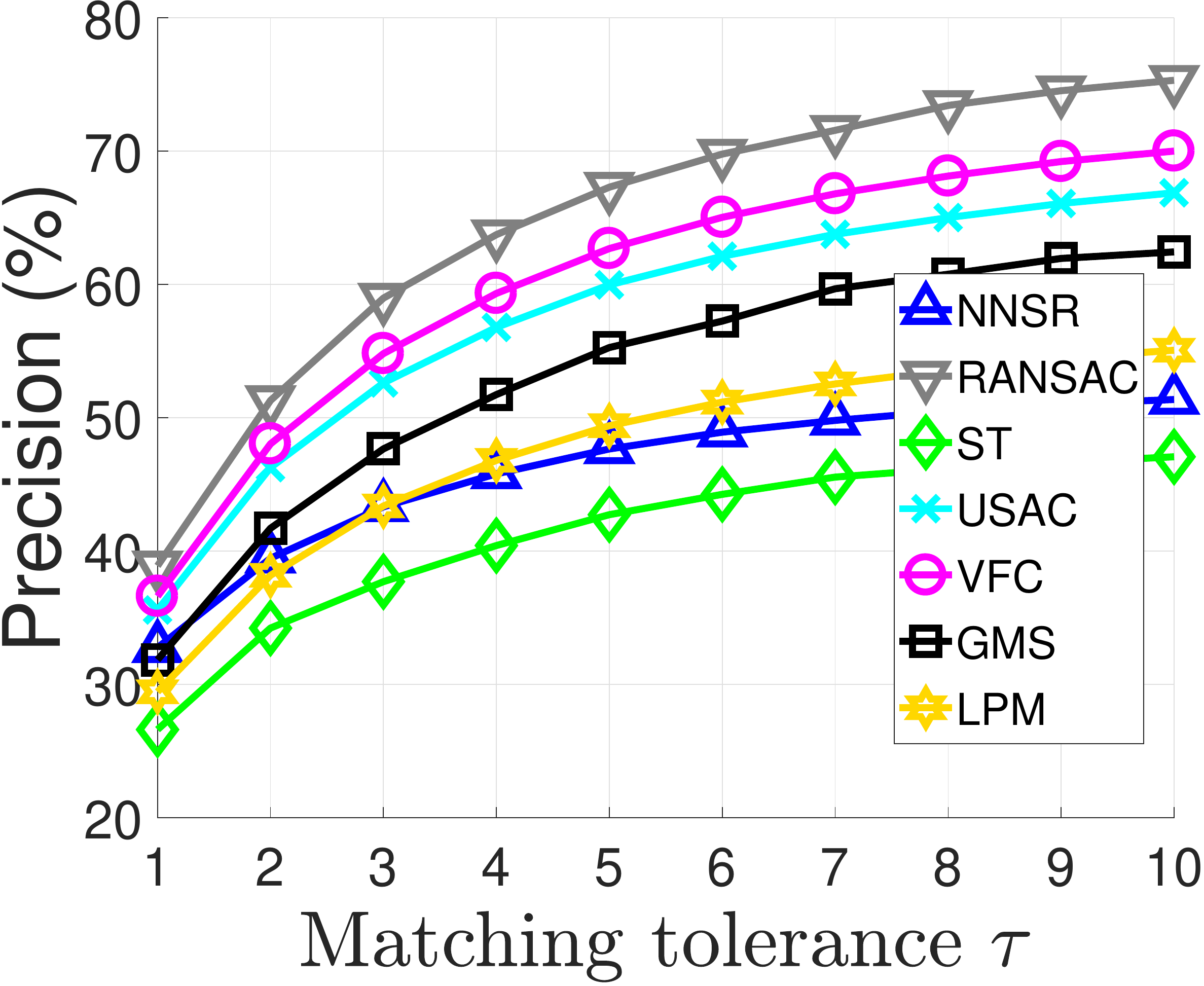} 
    	\includegraphics[width=0.3\linewidth]{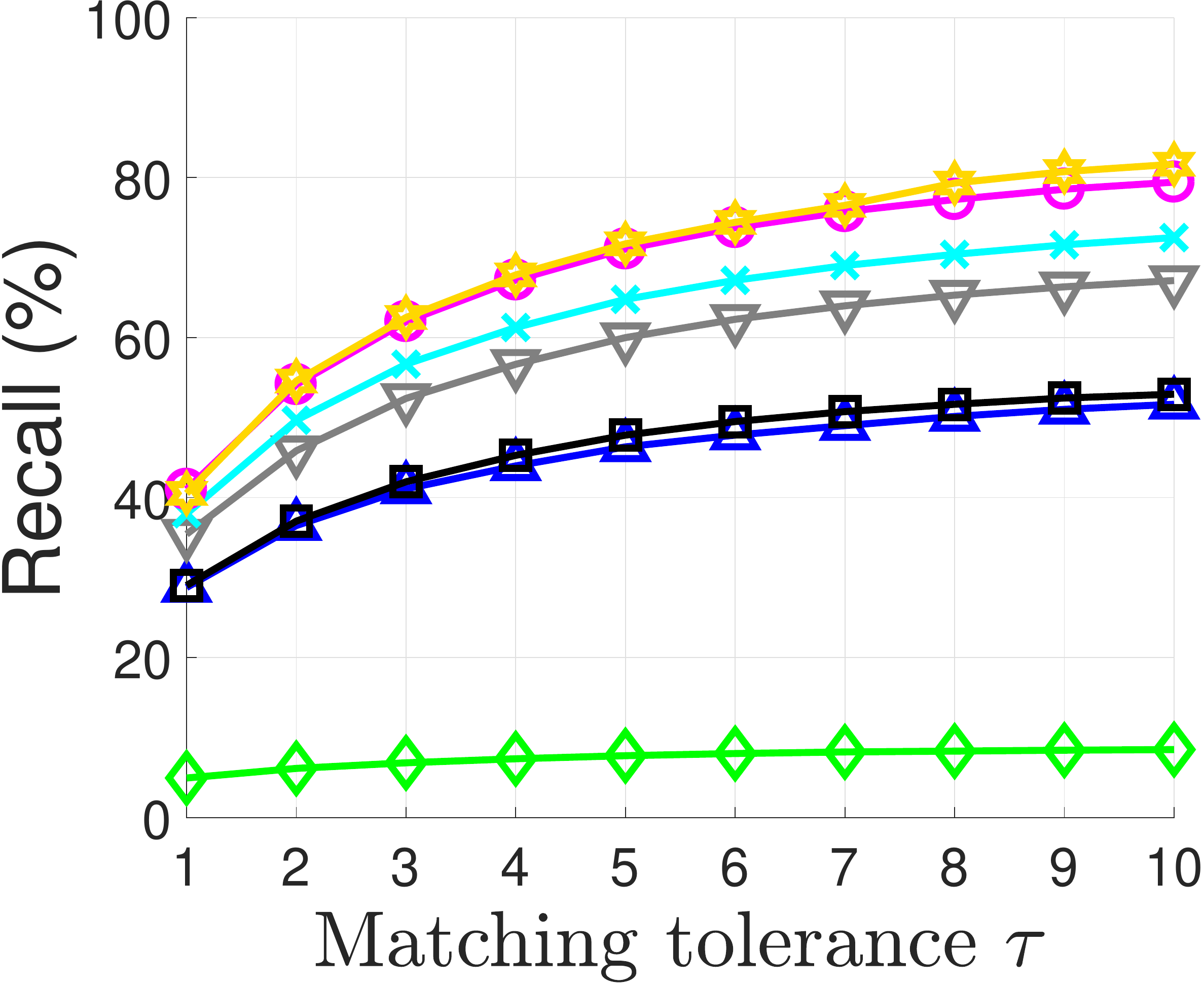}  
    	\includegraphics[width=0.3\linewidth]{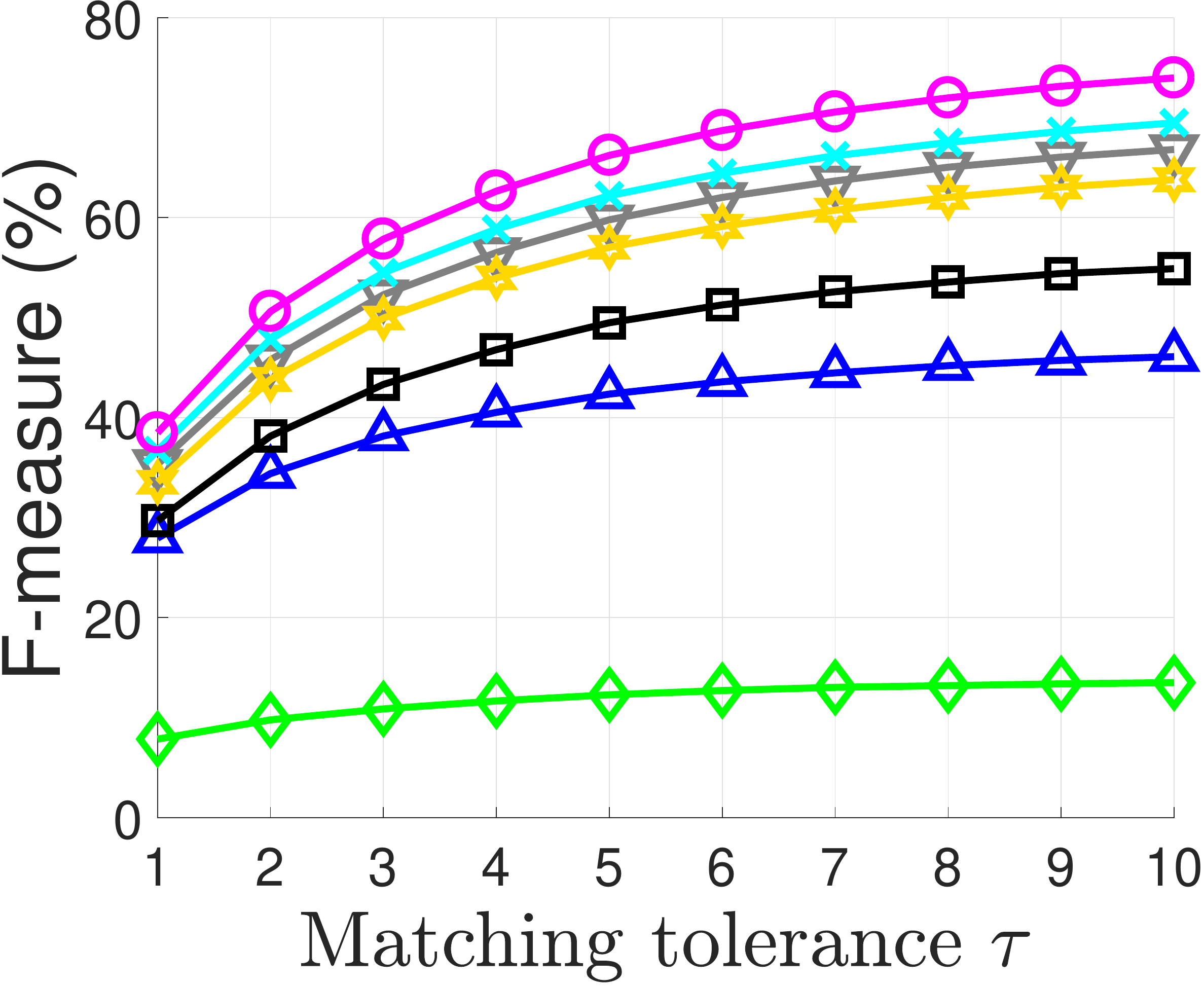}
    \end{minipage}

}
\subfigure[BLOB + FREAK]
{%
    \begin{minipage}[t]{0.47\linewidth}
    	\includegraphics[width=0.3\linewidth]{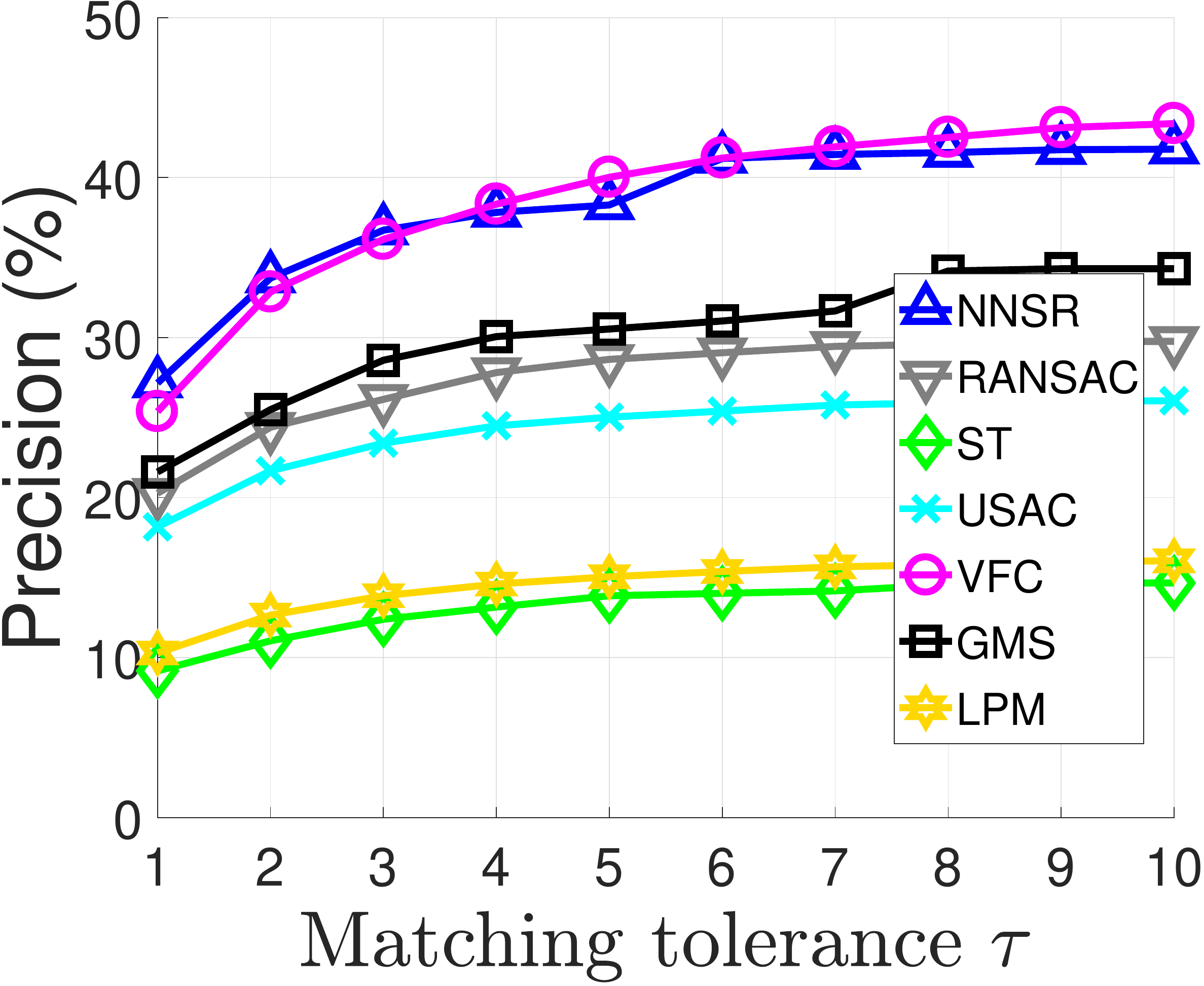} 
    	\includegraphics[width=0.3\linewidth]{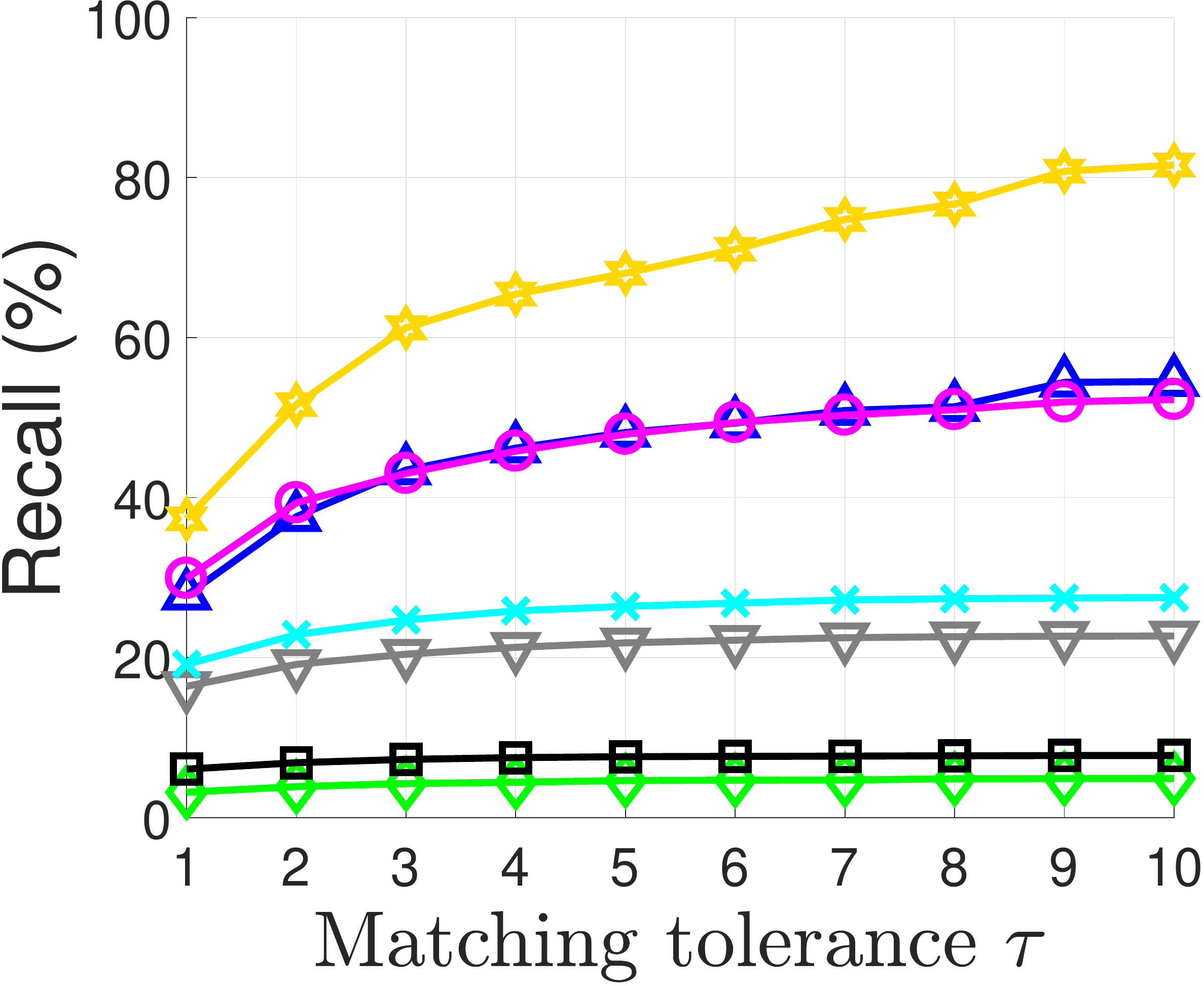} 
    	\includegraphics[width=0.3\linewidth]{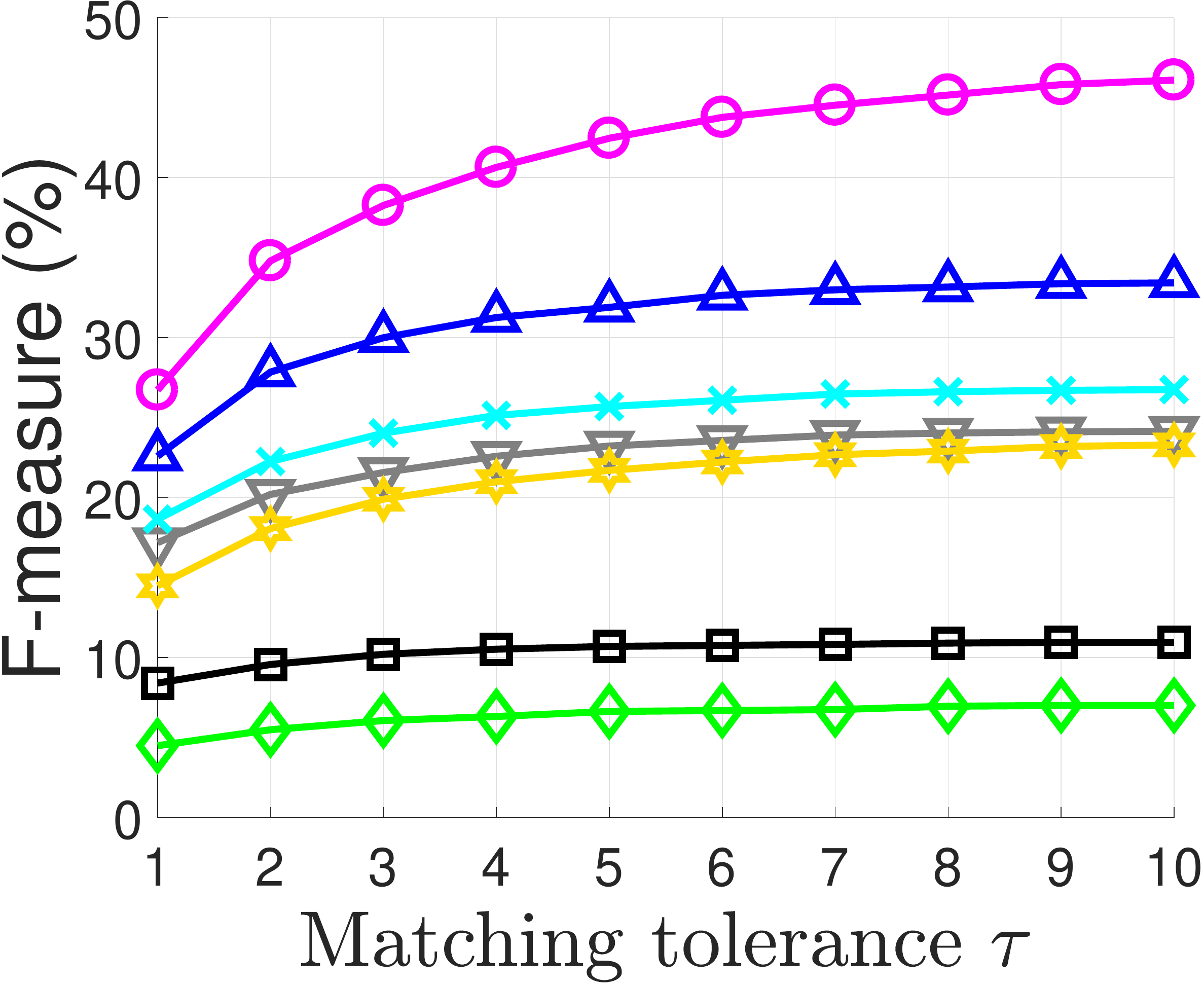}
    \end{minipage}
}
	\caption{Performance of evaluated algorithms on the VGG dataset using four different detector-descriptor combinations, i.e., (a) SIFT + SIFT, (b) ORB + ORB, (c) ASIFT + ASIFT, and (d) BLOB + FREAK.}
	\label{fig:per6}
\end{figure*}
\begin{table}[t]
	\scriptsize
	\renewcommand{\arraystretch}{1}
	\caption{F-measure $(\%)$ performance under different groups of detector and descriptor on the Symbench and Heinly datasets with $\tau=5$.}
	\label{tab:fea}
	\centering
	\begin{tabular}	{p{0.9cm}<{\centering}|p{0.8cm}<{\centering}p{0.6cm}<{\centering}p{0.6cm}<{\centering}p{0.5cm}<{\centering}p{0.5cm}<{\centering}p{0.4cm}<{\centering}p{0.4cm}<{\centering}p{0.4cm}<{\centering}}
	\hline
	& & NNSR & RANSAC & ST & USAC & VFC & GMS & LPM 	\\
	\hline
	SIFT + & Symbench & 9.34 & 3.02 & 1.22 & 3.27 & 11.56 & \textbf{11.64} & 9.36	\\		
	SIFT & Heinly & 94.13 & 95.43 & 33.82 & \textbf{98.75} & 83.08 & 40.29 & 89.84 \\
	\hline
	ORB + & Symbench & 4.70 & 5.27 & 2.13 & 3.33 & 3.00 & \textbf{11.62} & 6.31	\\		
	ORB & Heinly & 57.62 & 58.98 & 17.57 & 56.45 & 56.30 & 50.24 & \textbf{60.30} \\
	\hline	
	ASIFT + & Symbench & 7.00 & 7.15 & 3.29 & 4.54 & 14.42 & \textbf{17.48} & 12.54	\\		
	ASIFT & Heinly & 69.31 & \textbf{92.31} & 27.47 & 78.72 & 78.75 & 44.21 & 88.21 \\
	\hline
	BLOB + & Symbench & 4.62 & 2.35 & 0.95 & 1.97 & \textbf{6.25} & 0.50 & 2.19	\\		
	FREAK & Heinly & 68.63 & \textbf{76.25} & 20.32 & 74.45 & 68.71 & 4.30 & 66.64 \\
	\hline					
	\end{tabular}
\end{table}
In addition to Hessian-affine + SIFT, we also consider four other popular detector-descriptor combinations, i.e., SIFT + SIFT~\cite{lowe2004distinctive}, ORB + ORB~\cite{rublee2011orb}, ASIFT + ASIFT~\cite{Morel2009ASIFT}, and BLOB~\cite{lindeberg1998feature} + FREAK~\cite{alahi2012freak}. Fig.~\ref{fig:per5} shows the initial correspondences with these combinations on a sample image pair. Note that GTM is excluded in this test as it requires local affine information and these detectors do not provide this information. Also, the AdelaideRMF dataset is not considered due to human-labeled keypoints. The results are reported in Fig.~\ref{fig:per6} and Table~\ref{tab:fea}.

A common characteristic of these results is that the best correspondence selection algorithm generally varies with combinations of detector and descriptor. While we can still find some consistencies, e.g., the VFC method achieves pleasurable performance on the VGG dataset in spite of the descriptor-detector combinations. The performance of some methods fluctuates dramatically. For example, NNSR ranks the first with SIFT + SIFT while performs poorly using ASIFT + ASIFT on the VGG dataset. On the Symbench and Heinly datasets, GMS and RANSAC are two prominent methods under different kinds of detector-descriptor combinations. 
\subsection{Robustness}\label{sub:sub4}
\begin{figure}[H] 
	\centering	
	\includegraphics[width=0.9\linewidth]{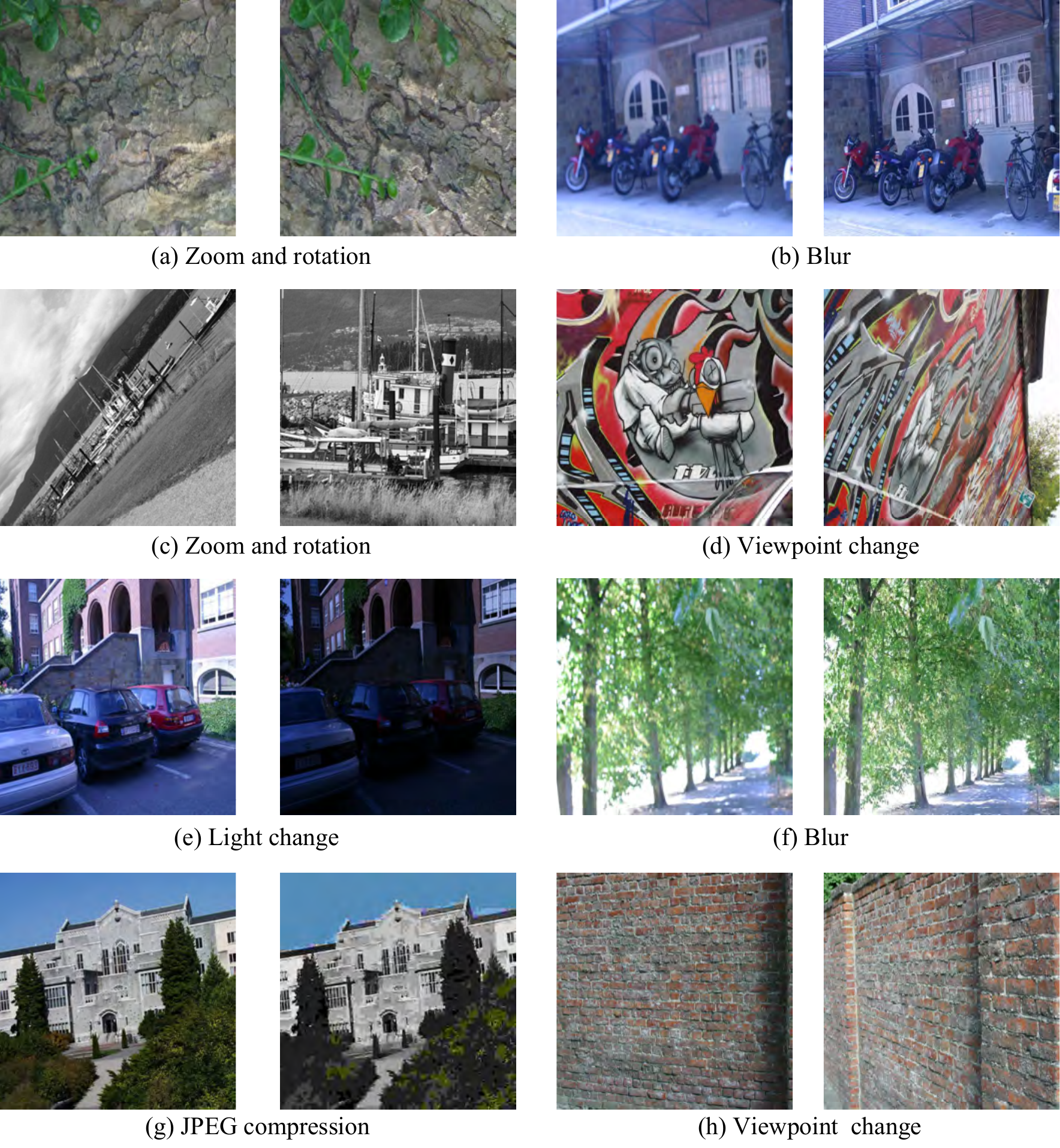}
	\caption{Sample image pairs from the 8 sub-categories of the VGG dataset including (a) zoom and rotation, (b) blur, (c) zoom and rotation, (d) viewpoint change, (e) light change, (f) blur, (g) JPEG compression and (h) viewpoint change.}
	\label{fig:per7}
\end{figure}
In this section, we independently evaluate the robustness of these algorithms to a specific nuisance, e.g., zoom, rotation, blur, viewpoint change, light change and JPEG compression on the VGG dataset. Some exemplar images with different nuisances are exhibited in Fig.~\ref{fig:per7}. Results are shown in Table~\ref{tab:rob}.

Under zoom and rotation (case1 and case3), USAC and RANSAC, i.e., two parametric methods, behave the best (F-measure is referred) mainly attributed to that zoom and rotation are faint impact on homography fitting. Under blur (case2 and case6), GMS and NNSR outperform others. GMS is independent from feature similarity constraint, thus making it rational. For NNSR, it is still explicable as SIFT is very robust to blur. Regarding viewpoint change (case4 and case8), USAC and VFC are the best methods. Note that VFC generally delivers good performance under all kinds of nuisances, being benefited from the consensus search in the non-parametric field. USAC also achieves the best performance under light change (case5) and JPEG compression (case7), being the one that is robust to the broadest categories of nuisances.
\begin{table*}[t]
	\scriptsize
	\renewcommand{\arraystretch}{1}
	\caption{Robustness results of evaluated algorithms against different nuisances with $\tau=5$. The best result is expressed in bold face.}
	\label{tab:rob}	
	\centering 
	\begin{tabular}	{p{2.5cm}<{\centering}p{2cm}<{\centering}p{1cm}<{\centering}p{1cm}<{\centering}p{1cm}<{\centering}p{1cm}<{\centering}p{1cm}<{\centering}p{1cm}<{\centering}p{1cm}<{\centering}p{1cm}<{\centering}p{1cm}<{\centering}}
		\hline
		& & NNSR & RANSAC & ST & GTM & UASC & VFC & GMS & LPM 	\\
		\hline
		Case1 & Precision & \textbf{81.16} & 76.11 & 17.98 & 43.22 & 77.38 & 67.19 & 63.61 & 42.50	\\
		(zoom and rotation) & Recall & 77.68 & 92.86 & 4.51 & 79.60 & \textbf{99.05} & 86.11 & 11.45 & 83.54	\\
		& F-measure & 77.35 & 82.42 & 6.56 & 53.69 & \textbf{84.48} & 74.27 & 18.46 & 54.75	\\
		\hline
		Case2 & Precision & \textbf{74.57} & 36.87 & 44.23 & 67.00 & 49.66 & 29.44 & 41.71 & 27.73 \\
		(blur) & Recall & \textbf{79.39} & 41.85 & 8.23 & 56.74 & 60.00 & 51.41 & 50.45 & 54.75 \\
		& F-measure & \textbf{71.87} & 38.71 & 13.46 & 61.12 & 54.30 & 35.27 & 45.54 & 35.86 \\
		\hline
		Case3 & Precision & 61.53 & \textbf{70.54} & 15.97 & 44.92 & 67.41 & 49.38 & 58.57 & 44.59\\
		(zoom and rotation) & Recall & 57.91 & 83.28 & 1.97 & 52.16 & 79.95 & \textbf{99.22} & 57.21 & 76.43 \\
		& F-measure & 53.74 & \textbf{74.81} & 3.50 & 44.83 & 73.12 & 61.91 & 56.35 & 55.10 \\
		\hline
		Case4 & Precision & 51.77 & 55.58 & 37.21 & 50.94 & \textbf{63.01} & 57.86 & 57.05 & 45.08\\
		(viewpoint change) & Recall & 61.63 & 66.38 & 3.52 & 68.55 & 79.73 & \textbf{97.08} & 75.52 & 83.97 \\
		& F-measure & 51.75 & 58.56 & 6.41 & 55.69 & 70.23 & \textbf{71.23} & 64.55 & 56.56 \\
		\hline
		Case5 & Precision & 76.28 & 81.44 & 61.90 & 68.90 & \textbf{83.76} & 71.99 & 64.89 & 57.65\\
		(light change) & Recall & 63.75 & 86.97 & 6.76 & 80.35 & \textbf{100} & \textbf{100} & 87.95 & 84.46 \\
		& F-measure & 68.00 & 82.34 & 11.61 & 73.94 & \textbf{91.11} & 82.49 & 74.37 & 67.90 \\
		\hline
		Case6 & Precision & 31.90 & 45.33 & 24.95 & 33.45 & 32.23 & 31.18 & \textbf{57.10} & 26.72 \\
		(blur) & Recall & \textbf{69.13} & 27.06 & 2.57 & 39.10 & 40.00 & 40.00 & 47.00 & 66.81 \\
		& F-measure & 31.49 & 28.86 & 4.34 & 34.29 & 35.68 & 35.02 & \textbf{50.80} & 35.82 \\
		\hline
		Case7 & Precision & 89.47 & 87.07 & 89.46 & 80.66 & \textbf{89.59} & 89.48 & 79.87 & 75.87 \\
		(JPEG compression) & Recall & 61.17 & 97.41 & 28.59 & 94.42 & \textbf{100} & \textbf{100} & 96.70 & 93.38 \\
		& F-measure & 72.42 & 91.81 & 43.07 & 86.88 & \textbf{94.43} & 94.26 & 87.25 & 83.43 \\
		\hline
		Case8 & Precision & 67.27 & 74.42 & 52.34 & 72.05 & 73.03 & 72.40 & \textbf{80.86} & 62.67 \\
		(viewpoint change) & Recall & 61.08 & 79.03 & 4.02 & 80.12 & 80.00 & 79.51 & 73.10 & \textbf{81.76} \\
		& F-measure & 58.39 & 76.23 & 7.36 & 73.42 & \textbf{76.33} & 75.74 & 76.08 & 69.64 \\
		\hline
	\end{tabular} 
\end{table*} 

\subsection{Efficiency}\label{sub:sub5}
\begin{figure}[H] 
	\centering
	\subfigure[VGG]
	{%
    	\begin{minipage}[t]{0.47\linewidth}
    		\includegraphics[width=1\linewidth]{figure/vgg-f.pdf} 
        \end{minipage}
	}
	\subfigure[Symbench]
	{%
    	\begin{minipage}[t]{0.47\linewidth}
    		\includegraphics[width=1\linewidth]{figure/sym-f.pdf} 
        \end{minipage}
	}
	\subfigure[Heinly]
	{%
    	\begin{minipage}[t]{0.47\linewidth}
    		\includegraphics[width=1\linewidth]{figure/hei-f.pdf} 
        \end{minipage}
	}
	\subfigure[AdelaideRMF]
	{%
    	\begin{minipage}[t]{0.47\linewidth}
    		\includegraphics[width=1\linewidth]{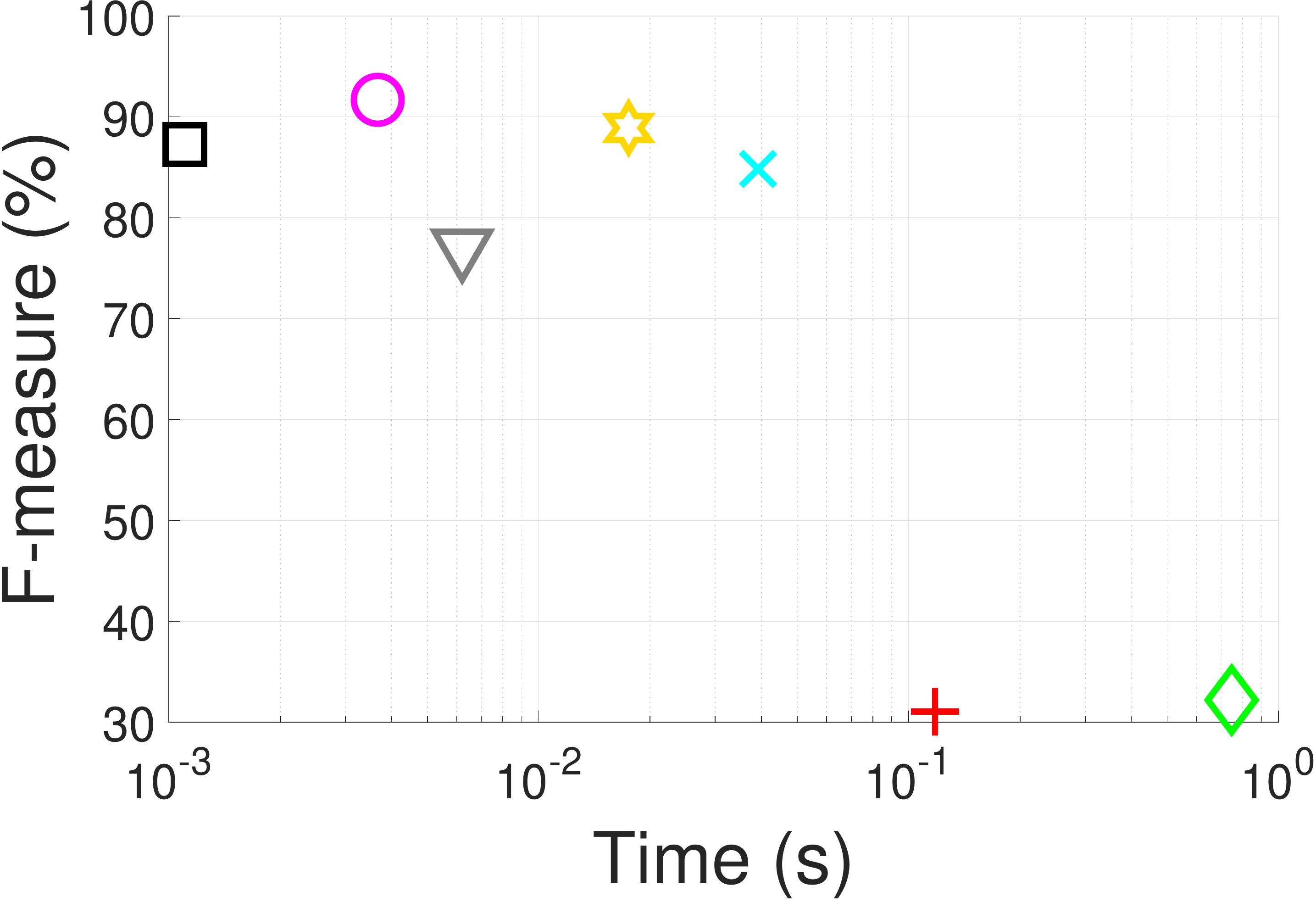} 
        \end{minipage}
	}
	\caption{Efficiency \textit{v.s.} F-measure plots on the (a) VGG, (b) Symbench, (c) Heinly and (d) AdelaideRMF datasets. The efficiency-axis is shown logarithmically for clarity.}
	\label{fig:per8}
\end{figure}
\begin{figure}[htbp] 
    \centering
    \subfigure[OPENCV]
    {%
        \begin{minipage}[t]{0.47\linewidth}
    	    \includegraphics[width=1\linewidth]{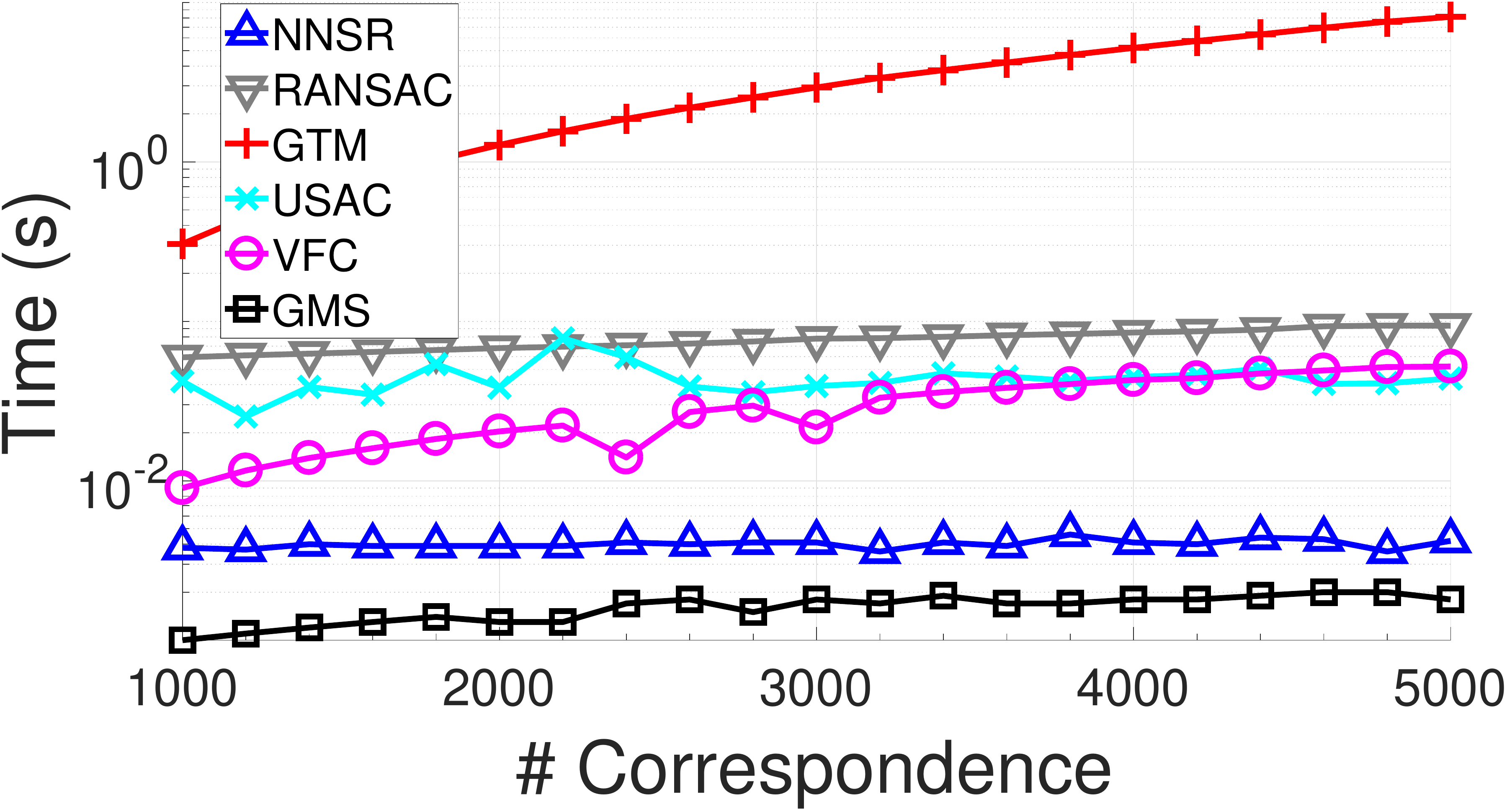}
    	\end{minipage}
    }
    \subfigure[MATLAB]
    {%
        \begin{minipage}[t]{0.47\linewidth}
    	    \includegraphics[width=1\linewidth]{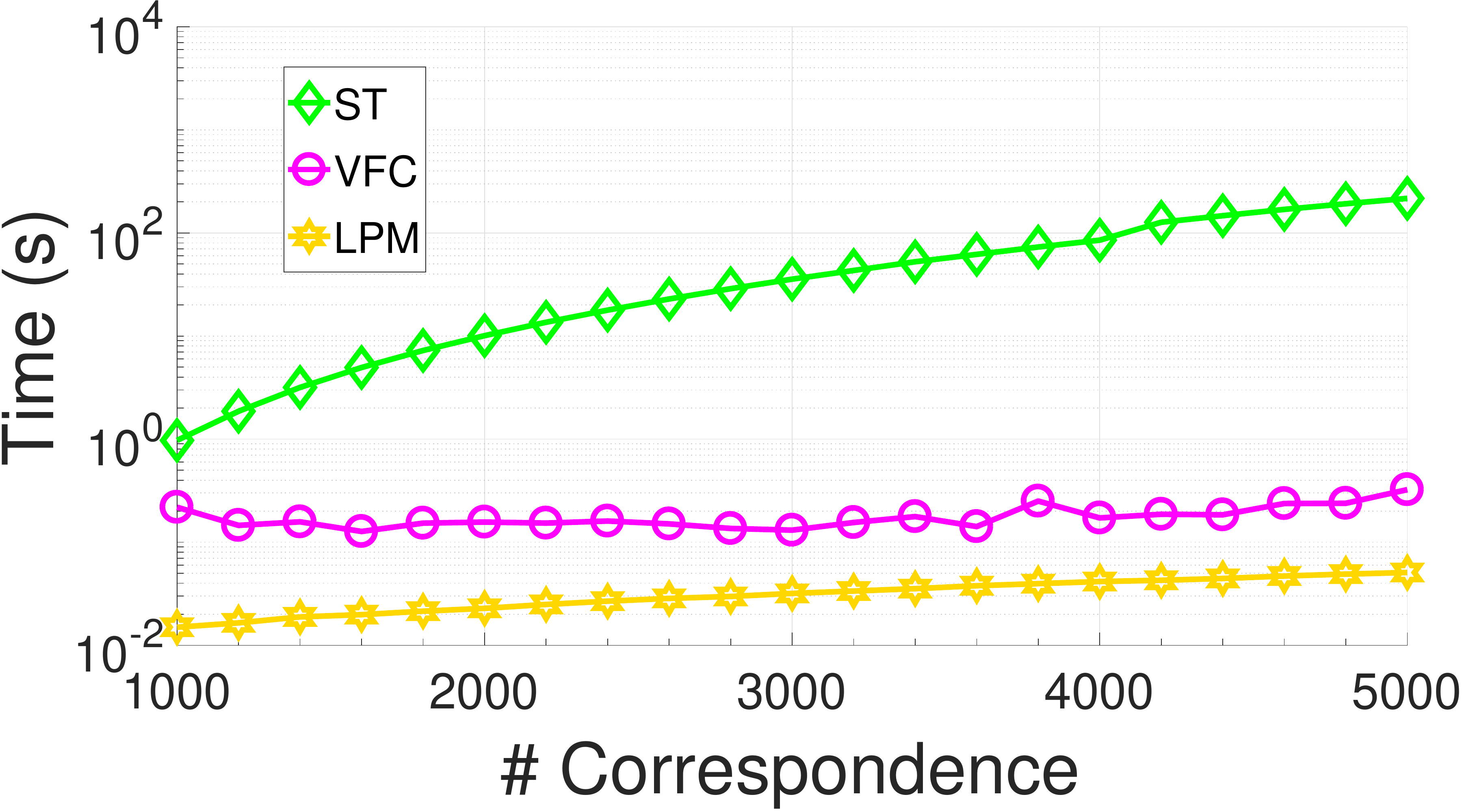}
    	\end{minipage}
    }
	\caption{Speed comparison of evaluated algorithms with respect to different numbers of initial correspondences. (a) and (b) present the results of methods implemented in OpenCV and MATLAB, respectively. To give a better comparison, VFC is implemented in both platforms. The time-axis is shown logarithmically for clarity.}
	\label{fig:per9}
\end{figure}
To provide an overview of the evaluated methods by taking both selection performance and efficiency into consideration, we present the efficiency \textit{v.s.} F-measure plots on the four experimental datasets in Fig.~\ref{fig:per8}. Owing to fast execution speed and overall decent performance, GMS strikes a good balance between selection performance and efficiency.

In order to further test an algorithms's efficiency regarding different numbers of initial correspondences, i.e., the number of initial correspondences may vary in different applications or with different feature detectors, we vary the amount of initial correspondences from $1000$ to $5000$ and record the average speed of the eight methods. This experiment has been repeated for 10 rounds and average statistics are retained. Because codes of these algorithms are implemented either in OpenCV (C++) or MATLAB, we assess methods within the same platform independently. In addition, the VFC method is evaluated on both platforms and can be a reference for comparing across-platform methods. Results are reported in Fig.~\ref{fig:per9}. 

For methods implemented in OpenCV, the efficiency of GMS is beyond all others. That is because GMS involves a grid framework for fast scoring. NNSR ranks the second, as only sort operation is needed to rank correspondences. RANSAC is slightly slower than USAC, and the core time consumption of both methods is dedicated to hypothesis generation-verification. GTM, with the computational complexity of $O(n^2)$ ($n$ being the number of input correspondences), is significantly slower than the other five methods. The margin is rather significant as the number of correspondences increases. For methods implemented in MATLAB, LPM is very efficient as it relies on a simple yet efficient strategy by preserving local neighborhood structure. ST is the most inefficient method, being slower than others by tens of magnitude with dense correspondences. It is due to the fact that the time consumption for computing eigenvalues increases exponentially with the size of the affinity matrix.
\subsection{Visual results}\label{sub:sub6}
\begin{figure}[t] 
	\vspace{-0.1cm}
	\centering
	\includegraphics[width=1.0\linewidth]{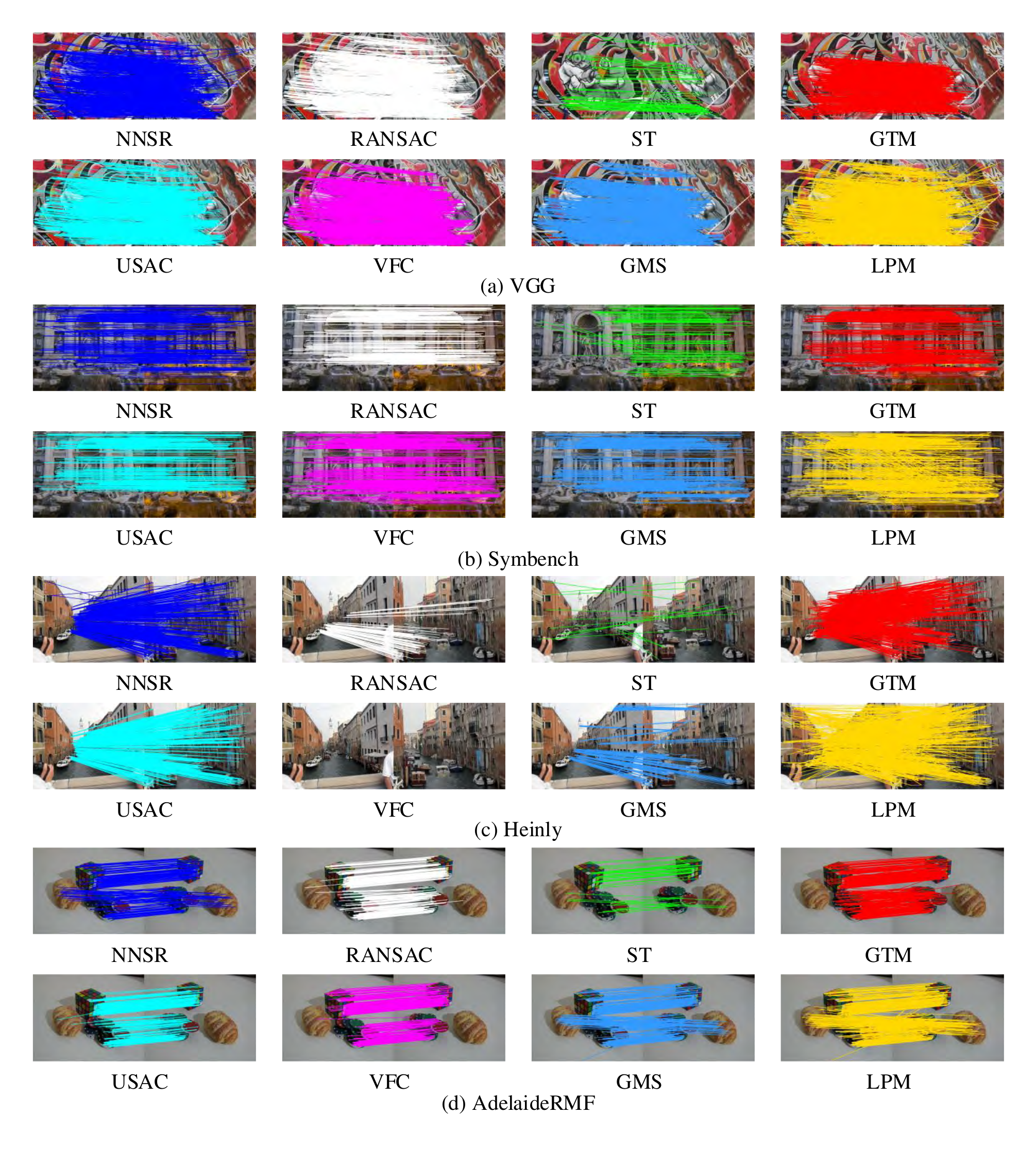} 
	\caption{Visual results of evaluated algorithms on examplar image pairs respectively taken from the  (a) VGG, (b) Symbench, (c) Heinly and (d) AdelaideRMF datasets. For the best view, lines with different colors represent results of different algorithms.}\label{fig:per10}
\end{figure}
To obtain a qualitative sense of outputs of evaluated algorithms, we present several visual results of these algorithms on the four experimental datasets in Fig.~\ref{fig:per10}.

Two main observations can be made from the figure. First, distributions of selected correspondences by different algorithms are generally different from each other. For instance, few correspondences are found by GTM on the \textit{bread} in Fig.~\ref{fig:per10}(d). However, NNSR and LPM get plenty of correspondences on it. Second, the quantity of selected correspondences also varies with different methods. In particular, LPM manages to return dense correspondences on most datasets, while ST seeks out much less than others. 
\section{Summary and discussion}\label{sec:sum}
\begin{table*}[t]\scriptsize
	\renewcommand{\arraystretch}{1}
	\caption{Summarization of superior and inferior correspondence selection algorithms in different scenarios based on the evaluation results. Note that this conclusion is drawn upon the F-measure, i.e., the aggregate performance regarding both precision and recall. Keypoint detector and descriptor are  abbreviated to Det and Des, respectively.}
	\label{tab:sum}
	\centering
	\begin{tabular}	{p{2.5cm}<{\centering}|p{3cm}<{\centering}|p{4cm}<{\centering}|p{3cm}<{\centering}}
		\hline
		\multicolumn{2}{c|}{\bf Scenarios} & \bf Superior methods & \bf Inferior methods 	\\
		\hline
		\multirow{4}{*}{Datasets} & VGG & USAC, RANSAC, VFC & ST \\
		\cline{2-4}
		 & Symbench & GMS & ST, USAC\\
		\cline{2-4}
		 &Heinly & RANSAC, NNSR, LPM & ST, GTM, GMS\\
		\cline{2-4}
		 & AdelaideRMF & VFC, LPM & ST, RANSAC\\
		\hline
		\multirow{3}{*}{NNSR pre-selection} &VGG & USAC, RANSAC, LPM & ST 	\\
		\cline{2-4}
		& Symbench & GMS, VFC & ST\\
		\cline{2-4}
		& Heinly & LPM, RANSAC, VFC & ST, GMS\\
		\hline
		\multirow{4}{*}{Det/Des combinations} & SIFT+SIFT & USAC, NNSR, GMS & ST, RANSAC \\
		\cline{2-4}
		& ORB+ORB &LPM, GMS, USAC & ST, VFC\\
		\cline{2-4}
		& ASIFT+ASIFT & VFC, RANSAC, GMS & ST, USAC, NNSR\\
		\cline{2-4}
		& BLOB+FREAK & VFC, NNSR, RANSAC & ST, GMS, USAC\\
		\hline
	    \multirow{4}{*}{Robustness} & Zoom and rotation & USAC, RANSAC & ST, GTM, GMS\\
	    \cline{2-4}
	    &Blur & NNSR, GMS & ST, RANSAC\\
	    \cline{2-4}
	    & Viewpoint change & USAC, VFC & ST, NNSR\\
	    \cline{2-4}
	    & Light change & USAC, VFC & ST\\
 	    \cline{2-4}
 	    & JPEG compression & USAC, VFC & ST, NNSR\\
		\hline
		\multicolumn{2}{c|}{Efficiency} & GMS, NNSR & ST, GTM 	\\
		\hline
	\end{tabular} 
\end{table*}

To give a quick guidance for developers regarding proper algorithms in a specific case, we list the superior and inferior correspondence selection in Table~\ref{tab:sum}. Also, peculiarities inherited to each evaluated algorithm are presented as follows:
\begin{itemize}
	\item {\bf NNSR} is arguably the most straightforward strategy to select correspondences. Its key strength is that repeatable patterns can be removed reliably in certain circumstances, provided that its employed feature detectors can locate the keypoints accurately and descriptors possess strong discriminative power, e.g., SIFT. Also, the high execution speed makes it suitable for real-time or near real-time systems. However, the limitation of NNSR is obvious because of the simple descriptor similarity constraint. It is vulnerable when image quality is low (e.g., facing with light change, blur, exposure, and style-transfer) and  texture information is limited.
	\item {\bf RANSAC} and {\bf USAC}, i.e., two evaluated parametric approaches, can fit the parametric models including the homography and fundamental matrices between two images effectively, with the premise that the image pair has homography or epipolar geometry constraint. Thus, they are prior options in such circumstances. Nevertheless, such assumption also brings drawbacks, e.g., when non-rigid objects are captured in images with large scale of parallax or the pure rotation between two camera positions, resulting in the failure of RANSAC and USAC. Further, the reliable models may not be generated by limited iterations with high outlier ratios, which will give rise to expensive time cost. For RANSAC, the minimal-sample models sometimes fall into the local optimization. USAC optimizes over RANSAC, though, it does not guarantee convergence and may produce an empty inlier set due to strict constraints.
	\item {\bf ST} and {\bf GTM} are methods relying on the affinity matrix computed from initial matches. We can find that these two methods are relatively time-consuming, especially for the ST method. The performance of GTM is much better than ST, mainly because GTM employs local affine information to judge the compatibility of two correspondences. While ST is based on rigid constraint. ST, when inputted with high-quality correspondences, is able to achieve high precision performance (as verified in Sect.~\ref{sub:sub2}). These two methods are optional for off-line applications desiring high precision and with high-quality input.
	\item {\bf LPM} rejects outliers by the local structure consistency. The constraint item in LPM is relatively loose, resulting in high recall yet relatively low precision. LPM prefers scenarios where the geometric structure information is well preserved between the same local pattern in the image pairs, e.g., small degrees of rigid transformations. Similar to NNSR, it relies strongly on the discriminative power of the feature descriptor. In other words, retrieving the local consistency can be problematic if the local region contains too few inliers. We therefore suggest to choose LPM in the context that has well preserved geometric structures and requires dense correspondences.  
	\item {\bf VFC}, as revealed by our experiment, is the most robust method under all tested scenarios. This is attributed to the fact that VFC is independent from the feature similarity and parametric models. Specifically, it performs inlier selection in a vector field. VFC generalizes well under different application contexts and can cope with various kinds of nuisances, especially for viewpoint change, light change and JPEG compression.
	\item {\bf GMS}, similar to VFC, is also independent from the feature similarity and parametric models. However, it assumes that the motion between two images is smooth. Accordingly, it behaves unsatisfactory for image pairs undergoing large degrees of rotation. While if the motion smoothness assumption holds, its performance is superior even for correspondence set with very limited number of inlier, e.g., correspondences generated from the Symbench dataset. Another attractive merit of GMS is the ultra fast execution speed even under  several thousands of initial correspondences, making it a prior selection for real-time applications.
\end{itemize}
\section{Conclusions}\label{sec:con}
This paper has comprehensively evaluated eight state-of\textbf{-th}e-art image correspondence selection algorithms, covering both parametric and non-parametric families. The experiments addressed  several critical issues regarding correspondence selection, e.g., different application scenarios (datasets), inputs from different combinations of feature detector and descriptor, robustness under various challenging conditions including  zoom, rotation, blur, viewpoint change, JPEG compression, light change, different rendering styles and multi-structures, and efficiency. Advantages and limitations, in light of experimental outcomes, are summarized so as to guide developers to choose a proper algorithm given a specific scenario. 

Remarkably, the performance of most existing algorithms changes dramatically in different scenarios and most methods fail to achieve satisfactory results when the inlier ratio of the initial correspondence set is low. We therefore believe the research should towards the development of correspondence selection algorithms with well generality and be robust to a low inlier rate.
\section*{Acknowledgment}
We are deeply grateful to the authors of the evaluated algorithms and datasets for making their contributions publicly available.  This work is supported by the National High Technology Research and Development Program of China (863 Program) under Grant 2015AA015904.

\bibliographystyle{IEEEtran}
\bibliography{mybibfile}
\end{document}